\documentclass[10pt]{article}

\usepackage[preprint]{bakelab}

\usepackage{wrapfig}
\usepackage{url}

\usepackage{longtable}
\usepackage{array}
\usepackage{multirow}
\usepackage{titletoc}
\tcbuselibrary{most}
\usepackage{minted}

\newcommand{\bench}{\textsc{VAB}}
\newcommand{\model}{\textsc{Kallisti-35B-A3B}}
\newcommand{\iconClaude}{\raisebox{-0.18\height}{\includegraphics[height=0.9em]{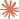}}\hspace{0.18em}}
\newcommand{\iconDoubao}{\raisebox{-0.18\height}{\includegraphics[height=0.9em]{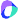}}\hspace{0.18em}}
\newcommand{\iconGemini}{\raisebox{-0.18\height}{\includegraphics[height=0.9em]{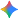}}\hspace{0.18em}}
\newcommand{\iconGPT}{\raisebox{-0.18\height}{\includegraphics[height=0.9em]{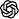}}\hspace{0.18em}}
\newcommand{\iconOseries}{\raisebox{-0.18\height}{\includegraphics[height=0.9em]{assets/logos/openai.pdf}}\hspace{0.18em}}
\newcommand{\iconGrok}{\raisebox{-0.18\height}{\includegraphics[height=0.9em]{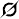}}\hspace{0.18em}}
\newcommand{\iconGLM}{\raisebox{-0.18\height}{\includegraphics[height=0.9em]{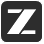}}\hspace{0.18em}}
\newcommand{\iconKimi}{\raisebox{-0.18\height}{\includegraphics[height=0.9em]{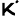}}\hspace{0.18em}}
\newcommand{\iconKallisti}{\raisebox{-0.18\height}{\includegraphics[height=0.9em]{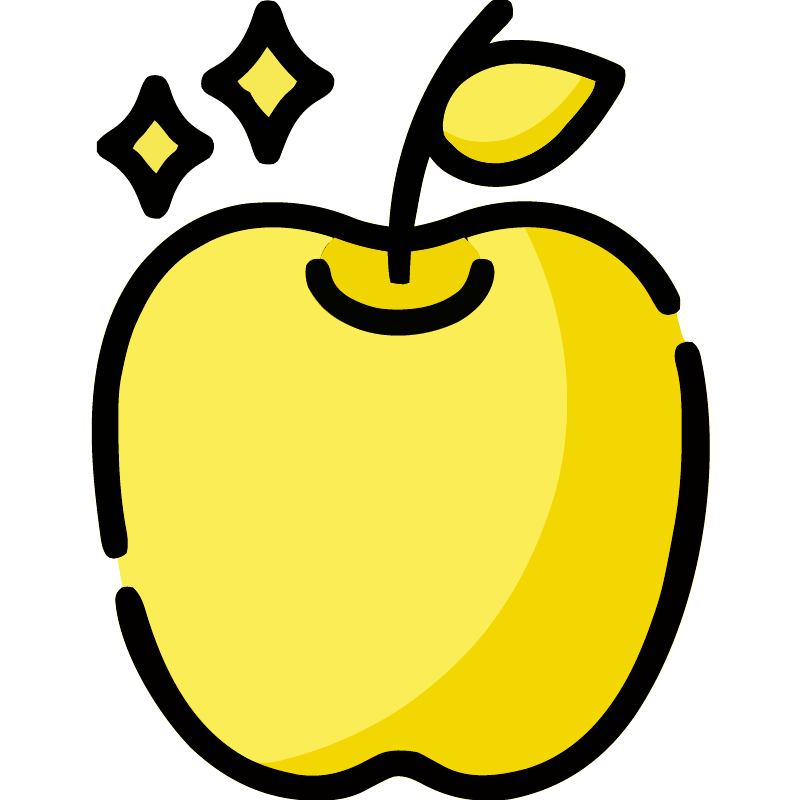}}\hspace{0.18em}}
\newcommand{\iconQwenVL}{\raisebox{-0.18\height}{\includegraphics[height=0.9em]{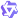}}\hspace{0.18em}}
\newcommand{\iconQwenThreeFive}{\raisebox{-0.18\height}{\includegraphics[height=0.9em]{assets/logos/qwen-color.pdf}}\hspace{0.18em}}
\newcommand{\bestcell}[1]{\textcolor{BakeAccent}{\textbf{#1}}}

\newcommand{\github}{\raisebox{-1.5pt}{\includegraphics[height=1em]{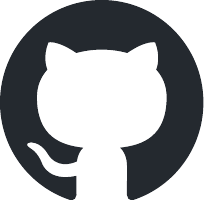}}}
\newcommand{\huggingface}{\raisebox{-1.5pt}{\includegraphics[height=1em]{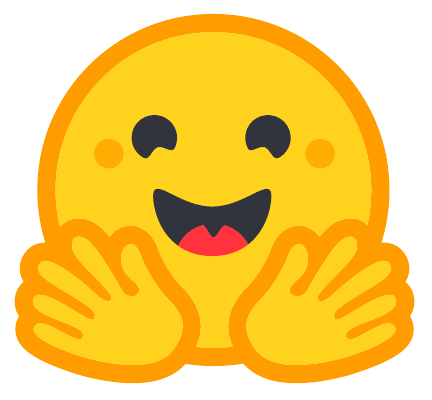}}}
\usepackage{fontawesome5}
\newcommand{\website}{\raisebox{-0.5pt}{\textcolor{BakeDark}{\faGlobe}}}
\newcommand{\corres}{\textsuperscript{\,\textcolor{BakeAccent}{\scalebox{0.7}{\faEnvelope[regular]}}}}
\makeatletter\let\@correspondence\empty\let\@paperurl\empty\makeatother

\usepackage[normalem]{ulem}
\newcommand{\rev}[1]{#1}
\newcommand{\del}[1]{}
\newcommand{\chg}[2]{#2}

\newsavebox{\tabHsAccBox}
\newlength{\tabCommonWidth}

\tcbset{
    promptstyle/.style={
        enhanced,
        breakable,
        colback=BakeTint,
        colframe=BakeAccent,
        colbacktitle=BakeAccent,
        coltitle=white,
        rounded corners,
        sharp corners=north,
        boxrule=0.6pt,
        drop shadow=BakeBorder,
        attach boxed title to top left={
            xshift=-2mm,
            yshift=-2mm
        },
        boxed title style={
            rounded corners,
            size=small,
            colback=BakeAccent,
            colframe=BakeAccent
        },
        fonttitle=\bfseries\sffamily
    },
    rubricstyle/.style={
        enhanced,
        breakable,
        colback=BakeTint,
        colframe=BakeAccent,
        colbacktitle=BakeAccent,
        coltitle=white,
        rounded corners,
        sharp corners=north,
        boxrule=0.6pt,
        drop shadow=BakeBorder,
        attach boxed title to top left={
            xshift=-2mm,
            yshift=-2mm
        },
        boxed title style={
            rounded corners,
            size=small,
            colback=BakeAccent,
            colframe=BakeAccent
        },
        fonttitle=\bfseries\sffamily
    }
}

\newtcolorbox{promptbox}[1][]{
    promptstyle,
    title=Prompt,
    #1
}

\newtcolorbox{rubricbox}[1][]{
    rubricstyle,
    title=Rubric,
    #1
}

\title{Visual Aesthetic Benchmark: Can Frontier Models Judge Beauty?}

\author{%
Yichen Feng$^{1,2,\dagger}$\equalcontrib \quad
Yuetai Li$^{2}$\equalcontrib \quad
Chunjiang Liu$^{1,6,\dagger}$\equalcontrib \\
Yuanyuan Chen$^{1}$ \quad
Fengqing Jiang$^{2}$ \quad
Yue Huang$^{5}$ \quad
Hang Hua$^{7}$ \quad
Zhengqing Yuan$^{5}$ \quad
Kaiyuan Zheng$^{2}$ \\
Luyao Niu$^{2}$ \quad
Bhaskar Ramasubramanian$^{2,8}$ \quad
Basel Alomair$^{9}$ \quad
Xiangliang Zhang$^{5}$ \quad
Misha Sra$^{3}$ \\
Zichen Chen$^{1,3,4}$\corres \quad
Radha Poovendran$^{2}$ \quad
Zhangchen Xu$^{1,2}$%
}

\affiliations{%
$^{1}$Bake AI \quad
$^{2}$University of Washington \quad
$^{3}$University of California, Santa Barbara \\
$^{4}$Stanford University \quad
$^{5}$University of Notre Dame \quad
$^{6}$Carnegie Mellon University \\
$^{7}$MIT-IBM Watson AI Lab \quad
$^{8}$Western Washington University \quad
$^{9}$King Abdulaziz City for Science and Technology \\[3pt]
{\footnotesize\itshape \textcolor{BakeAccent}{$^{*}$}Equal contribution. \quad $^{\dagger}$Work done during internship at Bake AI. \quad \textcolor{BakeAccent}{\scalebox{0.85}{\faEnvelope[regular]}}\,Corresponding author.}%
}

\paperdate{April 2026}

\begin{document}

\maketitle

\begin{abstract}
Multimodal large language models (MLLMs) are now routinely deployed for visual understanding, generation, and curation.
A substantial fraction of these applications require an explicit aesthetic judgment.
Most existing solutions reduce this judgment to predicting a scalar score for a single image.
We first ask whether such scores faithfully capture comparative preference: in a controlled study with eight expert annotators, score-derived rankings align poorly with the same annotators' direct comparisons, while direct ranking yields substantially higher inter-annotator agreement on best- and worst-image labels.
Motivated by this finding, we introduce the \textit{Visual Aesthetic Benchmark} (\bench{}), which casts aesthetic evaluation as comparative selection over candidate sets with matched subject matter.
\bench{} contains 400 tasks and 1{,}195 images across fine art, photography, and illustration, with labels derived from the consensus of 10 independent expert judges per task.
Evaluating 20 frontier MLLMs and six dedicated visual-quality reward models, we find that the strongest system identifies both the best and the worst image correctly across three random permutations of the candidate order in only 26.5\% of tasks, far below the 68.9\% achieved by human experts.
Fine-tuning a 35B-parameter model on 2{,}000 expert examples brings its accuracy close to that of a 397B-parameter open-weight model, suggesting that the comparative signal in \bench{} is transferable.
Together, these results expose a clear and measurable gap between current multimodal models and expert aesthetic judgment, and \bench{} provides the first set-based, expert-grounded testbed on which that gap can be tracked and closed.

\vspace{6pt}
{\noindent\footnotesize\sffamily
\href{https://github.com/BakeLab/Visual-Aesthetic-Benchmark}{%
  \github\,\textcolor{BakeAccent}{Code}\kern0.45em\textcolor{BakeGray}{\texttt{BakeLab/Visual-Aesthetic-Benchmark}}}%
\hspace{1.1em}\textcolor{BakeRule}{\rule[-0.2ex]{0.4pt}{1.8ex}}\hspace{1.1em}%
\href{https://huggingface.co/datasets/BakeLab/Visual-Aesthetic-Benchmark}{%
  \huggingface\,\textcolor{BakeAccent}{Dataset}\kern0.45em\textcolor{BakeGray}{\texttt{BakeLab/Visual-Aesthetic-Benchmark}}}%
\hspace{1.1em}\textcolor{BakeRule}{\rule[-0.2ex]{0.4pt}{1.8ex}}\hspace{1.1em}%
\href{https://vab.bakelab.ai}{%
  \website\,\textcolor{BakeAccent}{Website}\kern0.45em\textcolor{BakeGray}{\texttt{vab.bakelab.ai}}}%
\par}
\end{abstract}

\begin{figure}[h]
  \centering
  \includegraphics[width=\textwidth]{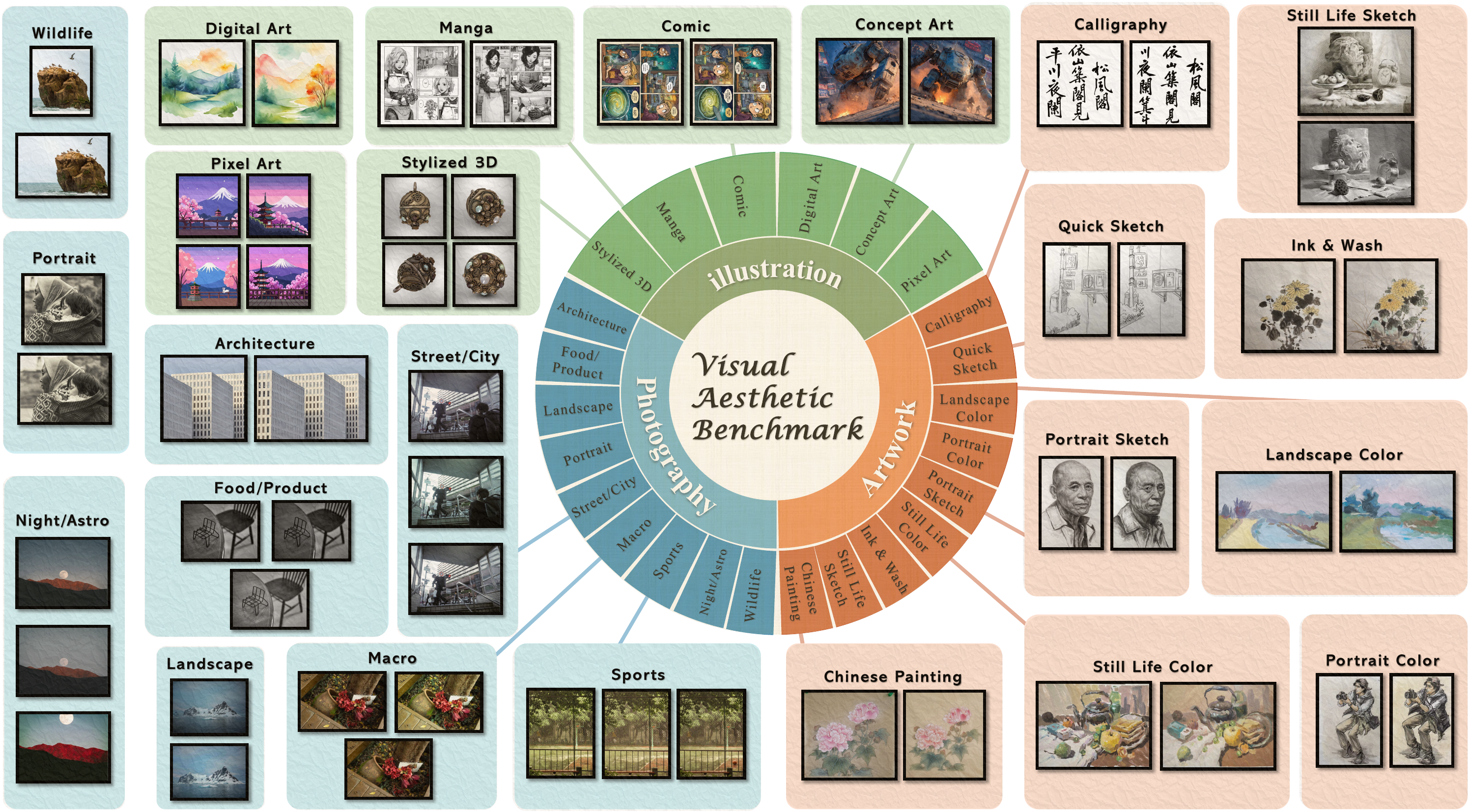}
  \caption{Overview of the \bench{} benchmark. \bench{} spans three visual domains --- fine art, photography, and illustration --- covering 24 topics across 400 evaluation tasks and 1,195 images. Each task presents a set of candidate images sharing a common subject.}
  \label{fig:teaser}
\end{figure}

\section{Introduction}

Multimodal large language models (MLLMs) have become critical to a widening range of visual tasks such as image generation and curation \citep{liu2023visual,team2023gemini,wang2024qwen2}. 
Many of these applications require aesthetic judgment, asking models to compare visual works and identify differences in execution quality across composition, color harmony, technical execution, and cultural convention.
This makes aesthetic judgment a demanding test of higher-order multimodal reasoning that is often omitted by existing visual benchmarks \citep{yu2023mm,yue2024mmmu}.

Existing image-aesthetics datasets and methods typically reduce aesthetic judgment to a mapping from \rev{an }image to a scalar score \citep{murray2012ava,kong2016photo,he2022rethinking,lee2019image,hosu2019effective,yi2023towards,he2023thinking}\rev{, an approach that we argue is poorly suited to the underlying decision being studied}.
\chg{This paradigm has three limitations. First, p}{P}reference order is inferred from independent scores rather than measured through direct comparison\chg{.\hspace{0.25em}Second, c}{; c}rowd-averaged labels mix expertise, taste, and annotation noise, \chg{making aesthetic signal difficult to isolate.}{so the aesthetic signal is difficult to isolate;}
\chg{Third, b}{and b}ecause images \rev{in such datasets }often depict different subjects, \chg{models can exploit content preference rather than discriminating execution quality.}{a model can shortcut its way to high accuracy by exploiting content preferences instead of discriminating execution quality.}

\chg{In this paper, we first test}{We first ask} whether scoring faithfully captures comparative preference \chg{by asking eight expert annotators to evaluate}{. Eight expert annotators evaluate} the same images under both independent absolute scoring and direct comparative ranking\rev{, and the two protocols disagree}.
\chg{We find that the comparative paradigm}{Comparative ranking} yields 42 percentage points higher inter-annotator agreement on best-image selection than score-derived rankings \citep{talebi2018nima,ma2017lamp,wu2023q,zhou2024uniqa}.
\chg{This indicates that comparative ranking produces fundamentally different preference signals and captures preference more faithfully than the pointwise score-based setups used by most existing solutions.}{The two protocols therefore elicit different preference signals from the same annotators, and direct comparison captures the underlying judgment more faithfully than the pointwise scoring used in most existing pipelines.}

\chg{To bridge this gap}{Building on this finding}, we introduce the \textbf{Visual Aesthetic Benchmark (\bench{})}, which formulates aesthetic judgment as a comparative selection problem over candidate sets with matched subject matter.
\bench{} \chg{spans}{contains} 400 tasks and 1{,}195 images across fine art, photography, and illustration, covering 24 topics (Figure~\ref{fig:teaser}). Each task presents two to six images sharing a common subject, and the model selects the best image\del{,} and\rev{,} under the stricter setting\rev{,} also the worst image.
\chg{\bench{} is built around three design principles. First, matched subject matter isolates differences in execution rather than content. Second, ground-truth labels are derived from 10 expert judges using a dual-threshold consensus filter. Third, each task is evaluated under three independent random permutations of the candidate order. We report}{Three design choices distinguish \bench{} from prior aesthetic benchmarks. Matched subject matter within each task isolates differences in execution from differences in content; each ground-truth label is the consensus of 10 independent expert judges, filtered through a dual-threshold criterion to remove ambiguous cases; and every task is evaluated under three independent random permutations of the candidate order. We report}
\texttt{ap@1}, the average accuracy across the three evaluations, and \texttt{pass\textasciicircum{}3}, which \chg{counts a task as correct only if the model is}{credits a task only when the model is} correct in all three.

\chg{We evaluate}{We benchmark} 20 frontier MLLMs and six dedicated visual\chg{ quality}{-quality} reward models on \bench{}.
\chg{The best model, Claude Sonnet 4.6, achieves 26.5\% \texttt{pass\textasciicircum{}3}, 42.4 points below human experts (68.9\%).}{The strongest system, Claude Sonnet 4.6, reaches 26.5\% on \texttt{TB-1} \texttt{pass\textasciicircum{}3}, 42.4 points below the human baseline of 68.9\%.}
Models are also highly sensitive to candidate order\chg{ing}{}, retaining only 34--65\% of their \texttt{ap@1} accuracy when required to be correct across all \rev{three }\chg{candidate permutations.}{permutations,}
\chg{Performance further degrades sharply with candidate set size, with models exhibiting a 7.1$\times$ drop from two-image tasks to those involving more than four images, whereas humans drop by only 2.0$\times$.}{and they degrade much faster than humans as the candidate set grows: a $7.1\times$ drop from two-image to four-or-more-image tasks, against only $2.0\times$ for human experts.}
\chg{To summarize, this paper makes the following contributions:}{The contributions of this paper are as follows.}
\begin{itemize}
    \item \chg{We conduct a controlled human study showing that direct comparative ranking is substantially more reliable than scoring for subjective aesthetic annotation.}{A controlled human study with eight expert annotators establishes that direct comparative ranking is substantially more reliable than independent scoring for subjective aesthetic annotation, and motivates the design of the benchmark.}
    \item \chg{We introduce \bench{}, the first set-based, expert-grounded, multi-domain aesthetic benchmark with permutation-robust evaluation, comprising 400 tasks and 1,195 images across fine art, photography, and illustration.}{\bench{} is, to our knowledge, the first set-based, expert-grounded, multi-domain aesthetic benchmark with permutation-robust evaluation, covering 400 tasks and 1{,}195 images across fine art, photography, and illustration.}
    \item \chg{We provide a comprehensive evaluation of 20 multimodal large language models and 6 reward models, revealing a persistent 42.4-point gap to human experts and severe positional instability.}{A systematic evaluation of 20 frontier MLLMs and six dedicated reward models exposes a persistent 42.4-point gap to human experts on \texttt{TB-1} \texttt{pass\textasciicircum{}3} together with severe positional instability across permutations of the candidate order.}
    \item \chg{We show that expert aesthetic judgments transfer effectively to smaller models through lightweight adaptation. Fine-tuning on 2,000 expert examples enables a 35B-parameter model to approach the performance of a 397B-parameter model.}{Fine-tuning on 2{,}000 expert examples brings a 35B-parameter model close to a 397B-parameter open-weight model, indicating that expert comparative judgments transfer effectively through lightweight adaptation.}
\end{itemize}

\section{Human Annotation Study: Ranking vs.\ Scoring}
\label{sec:human_study}

Existing aesthetic benchmarks rely on \textit{Scoring} to collect ground truth labels \citep{murray2012ava,kong2016photo,he2022rethinking,cao2025artimuse}.
We conduct a controlled human study to explore the following research question regarding the annotation process: 

\textbf{Which protocol yields a more scientifically sound signal for data annotation?}

\begin{itemize}[leftmargin=1.5em]
    \item \textbf{Scoring:} expert annotates each image independently on an absolute aesthetic scale.
    \item \textbf{Ranking:} expert compares images and annotates their relative order.
\end{itemize}

\paragraph{Setup and Metrics.}
\chg{We set eight expert annotators to evaluate}{Eight expert annotators evaluate} the same images under two protocols: (1) \emph{scoring} assigns each image an independent score $s_{i,a}\in[0,10]$\chg{,}{;} (2) \emph{ranking} orders images directly within groups of size $n\in\{2,3,4,5\}$.
We construct 119 \emph{homogeneous-content} tasks, in which all candidates share the same underlying content, and 107 \emph{heterogeneous-content} tasks formed by regrouping images across source groups while preserving topic consistency.
We compare the two protocols along three dimensions:
\begin{itemize}[leftmargin=1.5em]
    \item \textbf{Fidelity} (Kendall's $\tau$ \citep{kendall1938new}): we compute $\tau = (C-D)/\tbinom{n}{2}$, where $C$ and $D$ are the numbers of concordant and discordant pairs between an annotator's score-derived ranking $\sigma_a^{\text{score}}$ and the same annotator's direct comparison.
    A low $\tau$ means the two protocols elicit different preference orderings from the same person.
    We additionally report the top-1 self-consistency (SC) rate: the fraction of annotator--task pairs for which $\sigma_a^{\text{score}}$ and the direct comparison select the same top image.
    \item \textbf{Reproducibility} ($\mathrm{acc}_{\text{best/worst}}$): $\mathrm{acc}_{\text{best}}$ is the proportion of tasks where a single image receives $\geq 5$ of 8 best-image votes; $\mathrm{acc}_{\text{worst}}$ and $\mathrm{acc}_{\text{best\&worst}}$ are defined analogously.
    Higher accuracy means the protocol produces ground truth labels that different experts can independently reproduce.
    \item \textbf{Distinguishability} (top-1 entropy \citep{shannon1948mathematical}): we measure $H = -\sum_i p_i \log_2 p_i$, where $p_i$ is the fraction of annotators selecting image $i$ as best.
    Lower entropy means annotators concentrate on a single top choice rather than spreading votes.
\end{itemize}
Full protocol details, notations, formal definitions, and statistical tests are in Appendix~\ref{sec:hs:appendix}.

\paragraph{\del{Takeaway 1: }Scoring does not faithfully preserve preference order.}
Table~\ref{tab:hs:tau} shows that score-derived rankings \chg{cannot align with the direct comparison given by the same annotator.}{often diverge from the direct comparisons given by the same annotator.}
The mean Kendall's $\tau$ is 0.188 for homogeneous-content and 0.351 for heterogeneous-content tasks; the top-1 self-consistency rates are 45.5\% and 52.0\%, respectively.
\chg{This implies that}{In other words,} nearly half of the numeric score annotations \chg{do not reliably preserve the order that the annotator would express in direct comparison.}{fail to recover the order that the same annotator expresses under direct comparison.}

\begin{table}[t]
\centering
\scriptsize
\caption{We report within-annotator agreement between expert's score-derived rankings $\sigma_a^{\text{score}}$ and direct comparison from 8 annotators, measured by Kendall's $\tau$ ($\uparrow$) and top-1 self-consistency rate (SC rate, $\uparrow$), defined as the fraction of annotator--task pairs where the score-derived top-1 matches the direct comparison top-1. The results show that nearly 50\% of the numeric score annotations do not reliably preserve the order that the annotator would express in direct comparison. }
\label{tab:hs:tau}
\begin{subtable}[t]{0.48\linewidth}
\centering
\caption{Results on homogeneous-content tasks.}
\label{tab:hs:tau_same}
\begin{tabular}{c c c c c}
\toprule
$n$ & $N$ tasks & Mean $\tau$ & Std.\ $\tau$ & SC rate \\
\midrule
2 & 28 & 0.098 & 0.511 & 54.9\% \\
3 & 38 & 0.274 & 0.328 & 49.7\% \\
4 & 35 & 0.126 & 0.243 & 34.3\% \\
5 & 18 & 0.267 & 0.243 & 43.6\% \\
\midrule
\textbf{All} & \textbf{119} & \textbf{0.188} & \textbf{0.354} & \textbf{45.5\%} \\
\bottomrule
\end{tabular}
\end{subtable}
\hfill
\begin{subtable}[t]{0.48\linewidth}
\centering
\caption{Results on heterogeneous-content tasks.}
\label{tab:hs:tau_diff}
\begin{tabular}{c c c c c}
\toprule
$n$ & $N$ tasks & Mean $\tau$ & Std.\ $\tau$ & SC rate \\
\midrule
2 & 19 & 0.342 & 0.427 & 67.1\% \\
3 & 37 & 0.286 & 0.300 & 48.0\% \\
4 & 34 & 0.441 & 0.172 & 51.1\% \\
5 & 17 & 0.324 & 0.256 & 45.6\% \\
\midrule
\textbf{All} & \textbf{107} & \textbf{0.351} & \textbf{0.291} & \textbf{52.0\%} \\
\bottomrule
\end{tabular}
\end{subtable}
\end{table}

\paragraph{\del{Takeaway 2: }Ranking yields more reproducible labels.}
Table~\ref{tab:hs:acc} shows that ranking yields substantially higher inter-annotator agreement on best and worst labels.
On homogeneous-content tasks, ranking improves $\mathrm{acc}_{\text{best}}$ by 42.0 points, $\mathrm{acc}_{\text{worst}}$ by 37.0 points, and $\mathrm{acc}_{\text{best\&worst}}$ by 51.3 points (all $p<0.001$, McNemar test \citep{mcnemar1947note})\chg{.\ }{; }\chg{T}{t}he same pattern holds for heterogeneous-content tasks.
\chg{In practical terms, ranking is more likely to yield best/worst labels that independent experts can reproduce.}{Independent experts can therefore reproduce best/worst labels far more reliably under ranking than under scoring.}

\begin{table}[t]
\centering
\scriptsize
\caption{We report inter-annotator consistency on best and worst decisions for 8 annotators, where a majority requires agreement from at least 5 of 8. Each row shows task-level accuracy $\mathrm{acc}$ ($\uparrow$), improvement $\Delta$, and the exact two-sided McNemar $p$-value ($\downarrow$) \citep{mcnemar1947note,fagerland2013mcnemar}. The Significance level is marked as $^{**}p<0.001$. Results show that ranking is more likely to yield annotation labels that independent experts can reproduce. }
\label{tab:hs:acc}
\setlength{\tabcolsep}{4pt}
\sbox{\tabHsAccBox}{%
\begin{tabular}{c cccc cccc}
\toprule
& \multicolumn{4}{c}{Homogeneous-content ($N=119$)}
& \multicolumn{4}{c}{Heterogeneous-content ($N=107$)} \\
\cmidrule(lr){2-5}\cmidrule(lr){6-9}
Metric & Scoring & Ranking & $\Delta$ & $p$ & Scoring & Ranking & $\Delta$ & $p$ \\
\midrule
$\mathrm{acc}_{\text{best}}$
  & 52.9\% & 95.0\% & $+$42.0\% & $<$0.001$^{**}$
  & 65.4\% & 94.4\% & $+$29.0\% & $<$0.001$^{**}$ \\
$\mathrm{acc}_{\text{worst}}$
  & 55.5\% & 92.4\% & $+$37.0\% & $<$0.001$^{**}$
  & 69.2\% & 93.5\% & $+$24.3\% & $<$0.001$^{**}$ \\
$\mathrm{acc}_{\text{best\&worst}}$
  & 39.5\% & 90.8\% & $+$51.3\% & $<$0.001$^{**}$
  & 49.5\% & 90.7\% & $+$41.1\% & $<$0.001$^{**}$ \\
\bottomrule
\end{tabular}%
}%
\global\setlength{\tabCommonWidth}{\wd\tabHsAccBox}%
\usebox{\tabHsAccBox}
\end{table}

\paragraph{\del{Takeaway 3: }Ranking makes the top choice stand out more clearly.}
Table~\ref{tab:hs:entropy} measures how concentrated their top-choice preference is within each task.
It shows that ranking produces lower entropy over top-1 selections \citep{shannon1948mathematical} (0.729 vs.\ 1.213) and a much higher rate of unanimous choices (10.1\% vs.\ 0.8\%; Wilcoxon \del{significance level }$p=1.7\times10^{-15}$ \citep{wilcoxon1945individual}).
\chg{More generally, ranking captures the preferred image stand out more clearly instead of spreading votes across several plausible winners.}{More generally, ranking concentrates votes on a single preferred image rather than spreading them across several plausible winners.}

\begin{table}[t]
\centering
\scriptsize
\caption{We report top-choice concentration in the homogeneous-content setting (119 tasks, 8 annotators), measured by entropy $H$ \citep{shannon1948mathematical} and the rate of unanimous tasks ($H=0$). The 95\% confidence intervals use a normal approximation for mean $H$ and a Wilson interval \citep{wilson1927probable} for the unanimous-task rate. A Wilcoxon signed-rank test \citep{wilcoxon1945individual} yields that the difference is statistically significant ($p=1.7\times10^{-15}$).}
\label{tab:hs:entropy}
\begin{tabular*}{\tabCommonWidth}{@{\extracolsep{\fill}} c c c}
\toprule
Metric & Scoring & Ranking \\
\midrule
 Mean $H$ ($\downarrow$) [95\% CI] & $1.213\ [1.137,\ 1.289]$ & $0.729\ [0.665,\ 0.793]$ \\
 Unanimous rate ($\uparrow$) & $0.8\%\ [0.1\%,\ 4.6\%]$ & $10.1\%\ [5.9\%,\ 16.8\%]$ \\
\bottomrule
\end{tabular*}
\vspace{-1.0em}
\end{table}

\paragraph{Interpretation.}
Our findings align with conclusions from two complementary disciplines.
\textbf{Aesthetically}, modern aesthetics favors a relational view of beauty \citep{goldman1995aesthetic,levinson2005oxford}: people judge preference more reliably when alternatives are juxtaposed than when items are assessed in isolation. \textbf{Cognitively}, comparative judgment is a more primitive operation than absolute evaluation \citep{thurstone1927law}. Absolute scoring requires maintaining an internal reference scale and converting qualitative impressions into numbers, imposing heavier cognitive load and greater instability \citep{stewart2005absolute,parducci1965category}.

We therefore adopt ranking as the annotation protocol for \bench{}. Extended discussion is provided in Appendix~\ref{subsec:hs:interpretation:app}.

\section{The \bench{} Benchmark}
\label{sec:benchmark}

\chg{Motivated by the findings in}{Building on the findings of} Section~\ref{sec:human_study}, we construct the \textbf{Visual Aesthetic Benchmark (\bench{})} \chg{to evaluate}{to test} whether frontier multimodal models can make expert-aligned aesthetic judgments under a ranking-based protocol.
\chg{Unlike}{In contrast with} prior evaluations that aggregate scalar ratings from crowd annotators \citep{murray2012ava,kong2016photo,he2022rethinking}, \bench{} \chg{formulates}{casts} aesthetic evaluation as \del{a }comparative selection \del{problem }over candidate sets with matched subject matter\chg{.}{:}
\chg{Each task asks the model to select the best image, and in the stricter setting also the worst image.}{each task asks the model to identify the best image and, in the stricter setting, also the worst.}
\chg{The remainder of this section summarizes the benchmark layout, data construction process, and ground-truth formation procedure.}{The remainder of this section describes the task layout, the data construction pipeline, and the ground-truth formation procedure.}

\subsection{Overview}

\begin{wrapfigure}{r}{0.5\textwidth}
  \centering
  \vspace{-0.8em}
  \includegraphics[width=\linewidth]{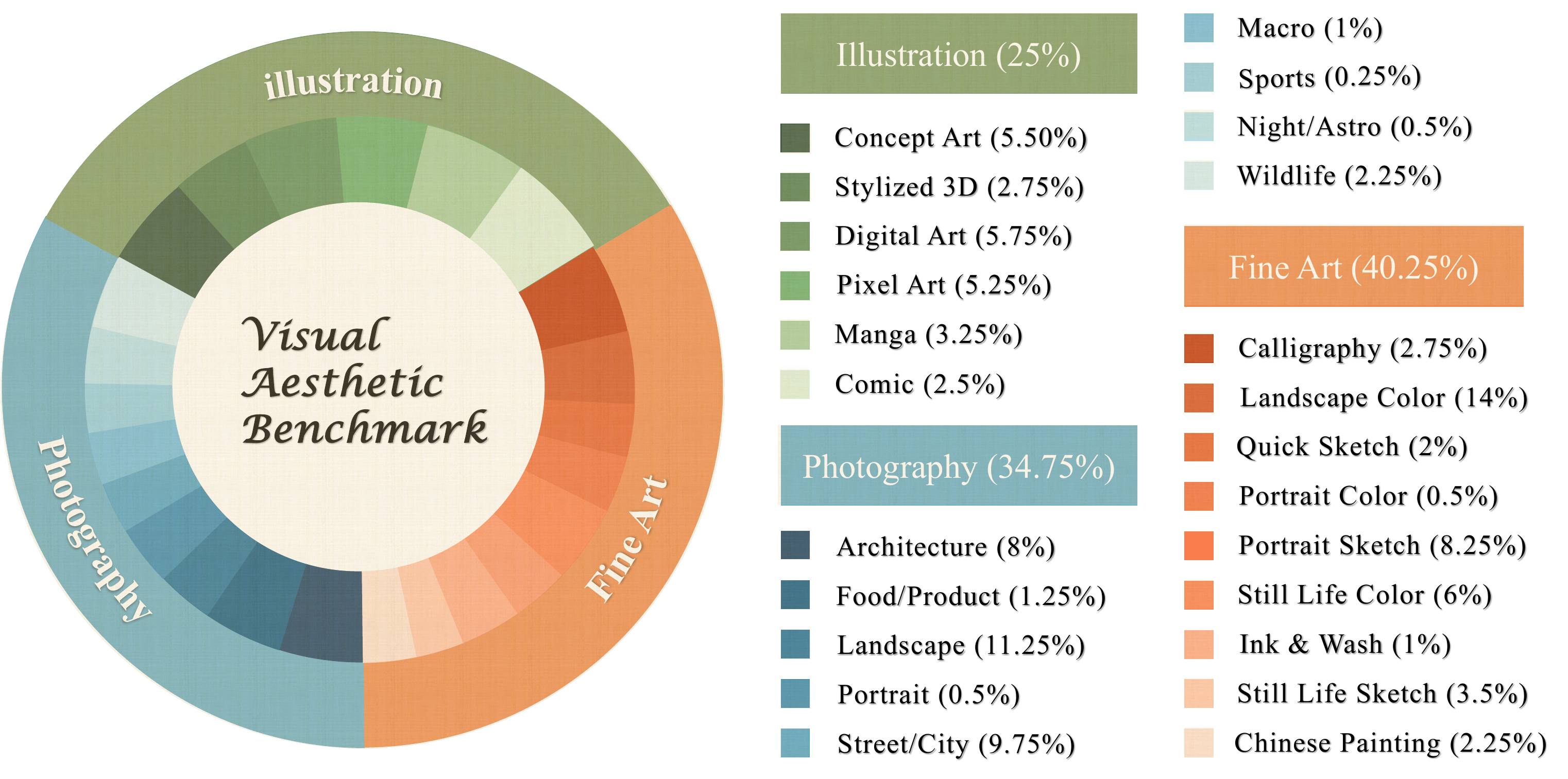}
  \caption{Detailed distribution of \bench{}.}
  \label{fig:distribution}
  \vspace{-1.0em}
\end{wrapfigure}

\textbf{Task formulation.} Each benchmark task presents a candidate set of images with matched subject matter, so performance depends on differences in execution rather than differences in depicted content. \bench{} contains 400 released tasks and 1,195 images across fine art, photography, and illustration, covering 24 topics with candidate set sizes from 2 to 6 images. We consider two settings: in \texttt{Top-1}, the model selects the best image in the set; in \texttt{TB-1}, it identifies both the best and the worst image. Figure~\ref{fig:teaser} provides a visual overview, Figure~\ref{fig:distribution} summarizes the released domain and topic distribution, and Appendix~\ref{app:benchmark_stats} reports detailed benchmark statistics.

\textbf{Evaluation protocol.} Labels are derived from direct comparative judgments by independent experts rather than post hoc aggregation of absolute scores. Each task is evaluated three times, each under an independent random permutation of the candidate order. We report \texttt{ap@1}, the average accuracy across the three evaluations, and \texttt{pass\textasciicircum{}3}, which counts a task as correct only if the model is correct in all three. Representative topic-level examples are collected in Appendix~\ref{app:annotation_examples}.

\subsection{Data Construction}

\begin{figure}[t]
  \centering
  \includegraphics[width=\linewidth]{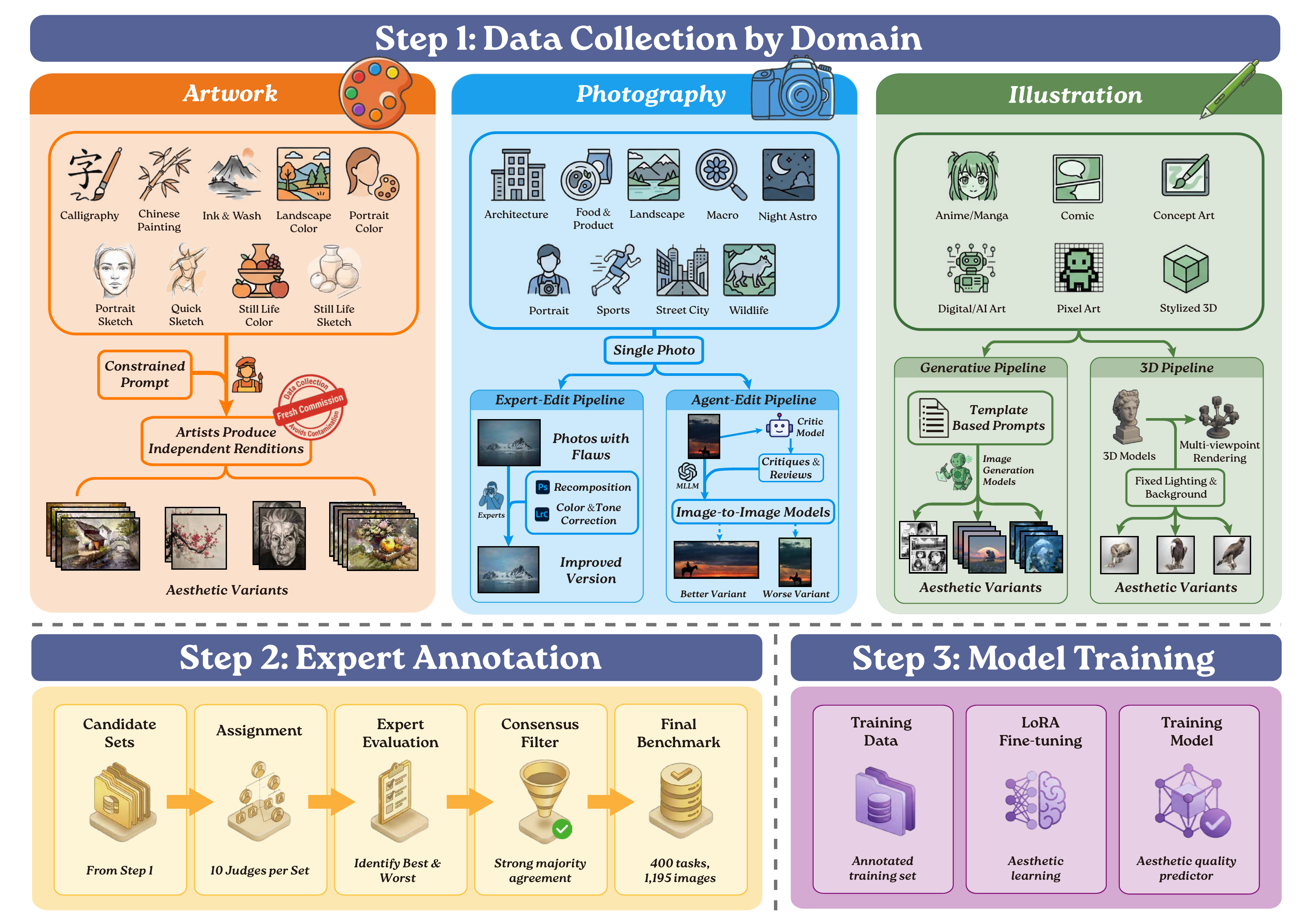}
  \caption{Overview of the \bench{} data construction pipeline. Images are organized into candidate sets with matched subject matter and varying aesthetic execution, then passed through expert review and consensus filtering before entering the benchmark.}
  \label{fig:pipeline}
  \vspace{-1.0em}
\end{figure}

\chg{Since \bench{} is annotated through direct ranking over candidate sets, data construction is organized around matched image groups rather than standalone images.}{Because \bench{} is annotated through direct ranking over candidate sets, data construction is organized around matched image groups rather than standalone images:}
\chg{Across all three domains, we construct sets whose members share subject matter but vary in execution quality.}{in every domain, the members of a set share subject matter but vary in execution quality.}
\chg{Figure~\ref{fig:pipeline} summarizes the pipeline, Appendix~\ref{app:benchmark_raw_stats} reports the full raw-pool statistics, and Appendix~\ref{app:benchmark_construction} provides domain-level construction details.}{Figure~\ref{fig:pipeline} summarizes the pipeline; Appendix~\ref{app:benchmark_raw_stats} reports the full raw-pool statistics, and Appendix~\ref{app:benchmark_construction} contains the domain-level construction details.}

\textbf{Fine Art.} We commissioned 426 painting sets comprising 1,142 images. Each set is built from constrained artist prompts that preserve subject matter while allowing variation in composition, value structure, color handling, and finish. All works were commissioned specifically for \bench{}.

\textbf{Photography.} We constructed 670 photography sets comprising 1{,}809 images. Each set originates from a single source photograph, many provided directly by professional photographers, and is expanded through two pipelines: an expert-edit pipeline based on professional retouching tools, which can produce either original/improved pairs or larger multi-expert sets, and a scalable agent-edit pipeline that generates improved and degraded variants. 
The agent-edit prompt sequence and runtime pipeline are recorded in Appendix~\ref{app:prompts:photo}.

\textbf{Illustration.} We constructed 271 illustration sets comprising 908 images. These sets are formed either by prompt-based generation with text-to-image models or by rendering CC0 3D assets under controlled viewpoint and presentation changes, producing multiple images with matched subject matter but varying execution. Prompt templates are in Appendix~\ref{app:prompts:illustration}.

All collected sets are deduplicated and reviewed for semantic consistency before annotation.

\subsection{Expert Annotation and Ground Truth}
\label{subsec:gt}

Following Section~\ref{sec:human_study}, \bench{} uses ranking-based expert consensus for ground truth.

\textbf{Review setup.} Creation and evaluation are strictly separated. In the raw pre-filter pool, each collected candidate set is reviewed by 10 independent expert judges drawn from a pool of more than 100 evaluators, yielding over 13,000 assessments and more than 300 hours of review. Judges remain blind to artist identity throughout annotation. Rubrics and interface examples are in Appendix~\ref{app:rubric}. Additional topic-level examples are in Appendix~\ref{app:annotation_examples}.

\begin{figure*}[t]
  \centering
  \includegraphics[width=\textwidth]{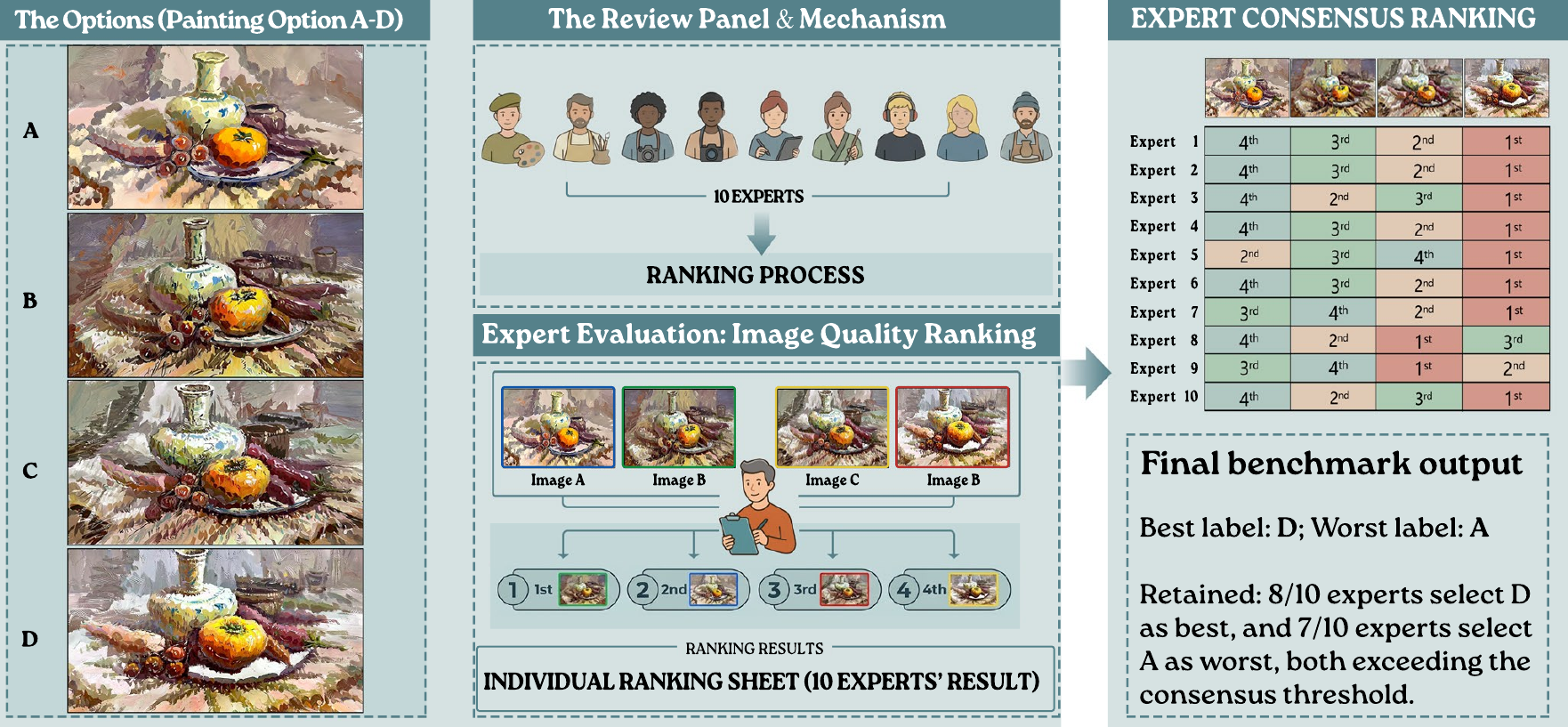}
  \caption{Example of a single \bench{} annotation task, illustrating the candidate set presented to expert judges and the resulting consensus decision. The consensus threshold is determined through statistical confidence; see Appendix~\ref{app:gt_math} for details.}
  \label{fig:annotation_example}
  \vspace{-1.5em}
\end{figure*}

\textbf{Annotation protocol.} Each task presents $k$ candidate images side by side. Judges first assess each image against the domain-specific rubric and then make a comparative selection over the full set. For $k = 2$, a judge selects the best image. For $k \ge 3$, a judge selects both the best and the worst image, with distinct choices required. Every task receives exactly 10 judgments. A representative annotation example is shown in Figure~\ref{fig:annotation_example}.

\textbf{Consensus filter.} A task is retained only if expert votes are sufficiently concentrated. For $k = 2$, the best-image votes must meet the threshold; for $k \ge 3$, the best-image and worst-image votes must both meet it, so satisfying only one is insufficient. The thresholds are chosen by statistical confidence under a uniform-random null, and the exact derivation and threshold table are provided in Appendix~\ref{app:gt_math}.

\section{Experiments}
\label{sec:experiments}

\subsection{Setup}

\textbf{Evaluation scope.} 
We evaluate 20 \chg{MLLMss}{MLLMs} and six reward models on \bench\rev{{}}\rev{.} \chg{Unless otherwise noted.}{Unless otherwise noted,} \chg{E}{e}ach model is evaluated under its default inference configuration.

We compare all systems against two baselines. The \emph{Human Expert} baseline is obtained from a fresh round of annotation in which 10 expert judges evaluate each released task under the benchmark interface, and we report their average accuracy against the released labels. The \emph{Random Guess} baseline is computed analytically for each task size $k$.

\paragraph{MLLMs.} Our MLLM evaluation covers 20 closed-source and open-weight models spanning the Claude family \citep{anthropic2026claudehaiku45,anthropic2026claudeopus45,anthropic2026claudeopus46,anthropic2026claudesonnet45,anthropic2026claudesonnet46}, the Gemini family \citep{google2026gemini3flash,google2026gemini3pro,google2026gemini31pro}, GPT and o-series models \citep{openai2025gpt41,openai2025gpt5,openai2025gpt51,openai2025gpt51codex,openai2025gpt52,openai2025o4mini}, and additional systems from Doubao, Grok, GLM, Kimi, and Qwen \citep{bytedance2026seed2,xai2025grok41fast,zai2026glm46v,team2026kimi,bai2023qwen,qwen2026qwen35}. For each task, the model selects either the best image (\texttt{Top-1}) or both the best and worst images (\texttt{TB-1}). We evaluate each task three times under independent random permutations of candidate order and report \texttt{ap@1} and \texttt{pass\textasciicircum{}3} as defined in Section~\ref{sec:benchmark}. Prompts are provided in Appendix~\ref{app:prompts:eval}.

\paragraph{Reward Models.} We additionally evaluate six reward models: ArtiMuse \citep{cao2025artimuse}, PEAS-aes and PEAS-ava \citep{yun2024scaling}, Q-Align \citep{wu2023q}, Q-Instruct \citep{wu2024q}, and RealQA-ava \citep{li2025next}. Unlike MLLMs, these models assign scalar scores to images independently rather than selecting directly from a set. We therefore follow each model's official scoring interface, score each image three times, average the scores, and then rank images within each task. The main text reports \texttt{Top-1} and \texttt{TB-1}; model-specific scoring interfaces and full result tables, including \texttt{Bot-1} and topic-level breakdowns, are deferred to Appendix~\ref{app:prompts:rm}, Appendix~\ref{app:rm_full_results}, and Appendix~\ref{app:rm_subdomain_results}.

\subsection{Results}

\chg{Table~\ref{tab:mllm:results} and Table~\ref{tab:rm:results} summarize the headline results.}{Tables~\ref{tab:mllm:results} and \ref{tab:rm:results} summarize the main results;} Appendices~\ref{app:full_results}--\ref{app:rm_subdomain_results} provide the full tables, topic-level breakdowns, and auxiliary metrics.

\subsubsection{Results for Multimodal Large Language Models}
\label{subsec:results:mllm}

\begin{table*}[t]
\centering
\caption{Main \bench{} results for 20 frontier MLLMs. We report \texttt{Top-1} and \texttt{TB-1} accuracy under both \texttt{ap@1} and \texttt{pass\textasciicircum{}3} (\%). Best result in each column is shown in \bestcell{bold}.}
\label{tab:mllm:results}
\resizebox{\textwidth}{!}{%
\footnotesize
\setlength{\tabcolsep}{2.5pt}
\renewcommand{\arraystretch}{1.05}
\begin{tabular}{@{}c *{4}{c} c *{4}{c} c *{4}{c} c *{4}{c}@{}}
\toprule
& \multicolumn{4}{c}{Overall} && \multicolumn{4}{c}{Fine Art} && \multicolumn{4}{c}{Illustration} && \multicolumn{4}{c}{Photography} \\
\cmidrule(r){2-5} \cmidrule(lr){7-10} \cmidrule(lr){12-15} \cmidrule(l){17-20}
& \multicolumn{2}{c}{\texttt{Top-1}} & \multicolumn{2}{c}{\texttt{TB-1}} && \multicolumn{2}{c}{\texttt{Top-1}} & \multicolumn{2}{c}{\texttt{TB-1}} && \multicolumn{2}{c}{\texttt{Top-1}} & \multicolumn{2}{c}{\texttt{TB-1}} && \multicolumn{2}{c}{\texttt{Top-1}} & \multicolumn{2}{c}{\texttt{TB-1}} \\
\cmidrule(r){2-3} \cmidrule(r){4-5} \cmidrule(lr){7-8} \cmidrule(lr){9-10} \cmidrule(lr){12-13} \cmidrule(lr){14-15} \cmidrule(l){17-18} \cmidrule(l){19-20}
Model & ap@1 & p\textasciicircum{}3 & ap@1 & p\textasciicircum{}3 && ap@1 & p\textasciicircum{}3 & ap@1 & p\textasciicircum{}3 && ap@1 & p\textasciicircum{}3 & ap@1 & p\textasciicircum{}3 && ap@1 & p\textasciicircum{}3 & ap@1 & p\textasciicircum{}3 \\
\midrule
\rowcolor{gray!10} Random Guess & 37.2 & 6.6 & 27.5 & 5.3 && 41.0 & 8.9 & 36.0 & 8.5 && 31.5 & 4.4 & 19.2 & 3.3 && 36.8 & 5.6 & 23.7 & 3.1 \\
\rowcolor{gray!10} Human Expert & \multicolumn{2}{c}{77.7} & \multicolumn{2}{c}{68.9} && \multicolumn{2}{c}{80.4} & \multicolumn{2}{c}{74.7} && \multicolumn{2}{c}{70.3} & \multicolumn{2}{c}{54.4} && \multicolumn{2}{c}{79.9} & \multicolumn{2}{c}{72.4} \\
\midrule
\multicolumn{20}{c}{\textit{Closed-Source Models}} \\
\midrule
\iconClaude\textsc{Claude-Haiku-4.5} & 43.3 & 18.8 & 32.8 & 12.5 && 44.7 & 19.9 & 40.4 & 19.3 && 32.7 & 13.0 & 18.7 & 5.0 && 49.4 & 21.6 & 34.3 & 10.1 \\
\iconClaude\textsc{Claude-Sonnet-4.5} & 47.2 & 24.2 & 35.7 & 14.5 && 44.5 & 20.5 & 40.2 & 19.3 && 37.3 & 17.0 & 20.7 & 8.0 && 57.6 & 33.8 & 41.2 & 13.7 \\
\iconClaude\textsc{Claude-Sonnet-4.6} & 54.8 & \bestcell{40.2} & 40.6 & \bestcell{26.5} && 48.4 & 37.3 & 44.7 & \bestcell{34.2} && 45.3 & 30.0 & \bestcell{29.3} & \bestcell{19.0} && \bestcell{68.8} & \bestcell{51.1} & 43.9 & 23.0 \\
\iconClaude\textsc{Claude-Opus-4.5} & 51.2 & 34.2 & 39.3 & 20.2 && 45.3 & 28.0 & 40.4 & 24.2 && 40.0 & 27.0 & 25.3 & 14.0 && 65.9 & 46.8 & 48.2 & 20.1 \\
\iconClaude\textsc{Claude-Opus-4.6} & 53.0 & 35.5 & 38.8 & 20.0 && 48.7 & 27.3 & 43.3 & 23.6 && 42.3 & 31.0 & 25.0 & 17.0 && 65.7 & 48.2 & 43.6 & 18.0 \\
\iconDoubao\textsc{Doubao-Seed-2.0-Pro} & 51.3 & 33.2 & \bestcell{42.8} & 23.5 && \bestcell{55.9} & \bestcell{39.8} & \bestcell{49.9} & 32.9 && 39.3 & 18.0 & 25.0 & 6.0 && 54.7 & 36.7 & 47.5 & 25.2 \\
\iconGemini\textsc{Gemini-3-Flash} & 52.8 & 28.0 & 40.7 & 15.8 && 51.8 & 23.0 & 46.8 & 20.5 && 46.0 & 22.0 & 26.0 & 8.0 && 58.8 & 38.1 & 44.1 & 15.8 \\
\iconGemini\textsc{Gemini-3-Pro} & \bestcell{55.4} & 35.0 & 40.0 & 22.0 && 52.0 & 32.9 & 46.6 & 29.8 && 47.3 & 28.0 & 28.7 & 14.0 && 65.2 & 42.4 & 40.5 & 18.7 \\
\iconGemini\textsc{Gemini-3.1-Pro} & 54.8 & 35.0 & 41.8 & 22.2 && 51.8 & 31.1 & 45.1 & 26.1 && \bestcell{49.7} & \bestcell{32.0} & 28.3 & 13.0 && 62.1 & 41.7 & 47.5 & 24.5 \\
\iconGPT\textsc{GPT-4.1} & 49.3 & 29.5 & 39.8 & 21.2 && 47.2 & 28.0 & 43.1 & 25.5 && 43.0 & 24.0 & 27.7 & 15.0 && 56.4 & 35.3 & 44.8 & 20.9 \\
\iconGPT\textsc{GPT-5} & 51.0 & 32.2 & 40.0 & 21.8 && 49.9 & 30.4 & 42.7 & 25.5 && 36.7 & 19.0 & 20.3 & 9.0 && 62.6 & 43.9 & 51.1 & 26.6 \\
\iconGPT\textsc{GPT-5.1} & 52.3 & 29.5 & 40.2 & 20.0 && 49.9 & 30.4 & 44.5 & 29.8 && 45.7 & 25.0 & 26.7 & 11.0 && 60.0 & 31.7 & 44.8 & 15.1 \\
\iconGPT\textsc{GPT-5.1-Codex} & 50.2 & 31.8 & 39.2 & 22.0 && 47.4 & 29.8 & 42.0 & 27.3 && 41.3 & 22.0 & 24.3 & 15.0 && 60.0 & 41.0 & 46.8 & 20.9 \\
\iconGPT\textsc{GPT-5.2} & 46.3 & 24.0 & 37.4 & 15.5 && 40.8 & 18.0 & 35.4 & 14.9 && 33.7 & 13.0 & 21.3 & 8.0 && 61.9 & 38.8 & 51.3 & 21.6 \\
\iconGrok\textsc{Grok-4.1-Fast} & 44.6 & 15.5 & 35.2 & 12.5 && 44.5 & 18.0 & 39.5 & 17.4 && 36.3 & 15.0 & 25.0 & 11.0 && 50.6 & 12.9 & 37.4 & 7.9 \\
\iconOseries\textsc{o4-mini} & 48.2 & 25.0 & 39.8 & 21.5 && 44.7 & 23.6 & 41.0 & 23.0 && 38.0 & 16.0 & 20.7 & 7.0 && 59.5 & 33.1 & \bestcell{52.3} & \bestcell{30.2} \\
\midrule
\multicolumn{20}{c}{\textit{Open-Weight Models}} \\
\midrule
\iconGLM\textsc{GLM-4.6V} & 43.0 & 17.5 & 34.1 & 11.5 && 42.7 & 18.0 & 39.3 & 18.6 && 43.3 & 19.0 & 27.0 & 8.0 && 43.2 & 15.8 & 33.1 & 5.8 \\
\iconKimi\textsc{Kimi-K2.5} & 49.4 & 26.8 & 36.8 & 15.0 && 43.5 & 19.9 & 38.5 & 18.0 && 43.0 & 21.0 & 29.0 & 14.0 && 60.9 & 38.8 & 40.5 & 12.2 \\
\iconQwenVL\textsc{Qwen3-VL-235B-A22B} & 44.9 & 20.5 & 35.2 & 14.0 && 42.4 & 22.4 & 36.9 & 19.3 && 38.3 & 12.0 & 25.7 & 8.0 && 52.5 & 24.5 & 40.0 & 12.2 \\
\iconQwenThreeFive\textsc{Qwen3.5-397B-A17B} & 50.1 & 29.8 & 38.9 & 17.2 && 42.7 & 21.1 & 37.3 & 17.4 && 44.3 & 24.0 & 26.0 & 8.0 && 62.8 & 43.9 & 50.1 & 23.7 \\
\bottomrule
\end{tabular}%
}
\vspace{-1.0em}
\end{table*}

\textbf{Gap to experts.} \chg{Current frontier models remain substantially below expert performance.}{Current frontier models remain far below expert performance.}
\chg{Under the strictest setting, \texttt{TB-1} \texttt{pass\textasciicircum{}3}, the strongest model, Claude Sonnet 4.6, achieves 26.5\%, compared with 68.9\% for human experts.}{On the strictest metric, \texttt{TB-1} \texttt{pass\textasciicircum{}3}, Claude Sonnet 4.6 attains 26.5\%, against 68.9\% for human experts;}
\chg{Even under \texttt{TB-1} \texttt{ap@1}, the best model, Doubao Seed 2.0 Pro at 42.8\%, remains 26.1 points below the expert baseline.}{even under the more lenient \texttt{TB-1} \texttt{ap@1}, the strongest model, Doubao Seed 2.0 Pro, reaches 42.8\%, still 26.1 points short of the expert baseline.}

\textbf{\chg{Model variation.}{Progress across families.}} \chg{Progress is uneven across model families.}{Improvements are uneven across model families.}
\chg{Claude Sonnet 4.6 improves over Claude Sonnet 4.5 from 14.5\% to 26.5\% on \texttt{TB-1} \texttt{pass\textasciicircum{}3}, while the GPT-5 series declines from 21.8\% for GPT-5 to 20.0\% for GPT-5.1 and 15.5\% for GPT-5.2.}{Claude Sonnet 4.6 nearly doubles its predecessor on \texttt{TB-1} \texttt{pass\textasciicircum{}3} (14.5\% $\rightarrow$ 26.5\%), whereas the GPT-5 series regresses from 21.8\% (GPT-5) to 20.0\% (GPT-5.1) and 15.5\% (GPT-5.2).}
\chg{A closed/open gap also remains under the stricter metric: Qwen3.5-397B-A17B reaches 17.2\% on \texttt{TB-1} \texttt{pass\textasciicircum{}3}, versus 26.5\% for Claude Sonnet 4.6.}{A clear closed/open gap persists under the stricter metric: the strongest open-weight system, Qwen3.5-397B-A17B, reaches 17.2\% on \texttt{TB-1} \texttt{pass\textasciicircum{}3}, compared with 26.5\% for Claude Sonnet 4.6.}

\textbf{\chg{Domain and stability.}{Domain difficulty and stability.}} \chg{Performance varies strongly across domains and across repeated evaluations.}{Accuracy varies strongly across domains and across repeated evaluations.}
\chg{Illustration is the hardest domain, where Claude Sonnet 4.6 reaches 19.0\% on \texttt{TB-1} \texttt{pass\textasciicircum{}3}, while photography is comparatively easier, with o4-mini reaching 30.2\%.}{Illustration is the most difficult domain, where Claude Sonnet 4.6 attains only 19.0\% on \texttt{TB-1} \texttt{pass\textasciicircum{}3}, while photography is comparatively more tractable, with o4-mini reaching 30.2\%.}
\chg{The large \texttt{ap@1}-to-\texttt{pass\textasciicircum{}3} drop, for example from 34.1\% to 11.5\% for GLM-4.6V on \texttt{TB-1}, indicates that candidate-order sensitivity and response instability remain major failure modes.}{The large \texttt{ap@1}-to-\texttt{pass\textasciicircum{}3} gap---for example, GLM-4.6V drops from 34.1\% to 11.5\% on \texttt{TB-1}---indicates that sensitivity to candidate order and response instability remain major failure modes.}

\subsubsection{Results for Reward Models}
\label{subsec:results:rm}

\begin{table}[t]
\centering
\footnotesize
\setlength{\tabcolsep}{2pt}
\renewcommand{\arraystretch}{1.05}
\caption{Main reward-model results on \bench{}. We report overall and domain-level \texttt{Top-1} and \texttt{TB-1} accuracy (\%). Best non-baseline result in each column is shown in \bestcell{bold}.}
\label{tab:rm:results}
\begin{tabular}{@{}c *{2}{c} c *{2}{c} c *{2}{c} c *{2}{c}@{}}
\toprule
& \multicolumn{2}{c}{Overall} && \multicolumn{2}{c}{Fine Art} && \multicolumn{2}{c}{Illustration} && \multicolumn{2}{c}{Photography} \\
\cmidrule(r){2-3} \cmidrule(lr){5-6} \cmidrule(lr){8-9} \cmidrule(l){11-12}
Model & Top-1 & TB-1 && Top-1 & TB-1 && Top-1 & TB-1 && Top-1 & TB-1 \\
\midrule
\rowcolor{gray!10} Random Guess & 37.2 & 27.5 && 41.0 & 36.0 && 31.5 & 19.2 && 36.8 & 23.7 \\
\rowcolor{gray!10} Human Expert & 77.7 & 68.9 && 80.4 & 74.7 && 70.3 & 54.4 && 79.9 & 72.4 \\
\midrule
\textsc{ArtiMuse}~\citep{cao2025artimuse} & 44.2 & 34.8 && 50.9 & \bestcell{46.0} && \bestcell{37.0} & 20.0 && 41.7 & 32.4 \\
\textsc{PEAS-aes}~\citep{yun2024scaling} & 48.5 & 37.8 && 49.7 & 44.1 && 34.0 & \bestcell{22.0} && 57.6 & 41.7 \\
\textsc{PEAS-ava}~\citep{yun2024scaling} & 49.0 & 37.0 && 50.9 & 44.1 && 25.0 & 13.0 && 64.0 & 46.0 \\
\textsc{Q-Align}~\citep{wu2023q} & \bestcell{52.2} & \bestcell{38.2} && \bestcell{52.2} & 44.7 && 31.0 & 16.0 && \bestcell{67.6} & \bestcell{46.8} \\
\textsc{Q-Instruct}~\citep{wu2024q} & 38.5 & 31.2 && 37.9 & 34.2 && 32.0 & 21.0 && 43.9 & 35.2 \\
\textsc{RealQA-ava}~\citep{li2025next} & 44.2 & 30.2 && 40.4 & 36.6 && 33.0 & 15.0 && 56.8 & 33.8 \\
\bottomrule
\end{tabular}
\vspace{-1.0em}
\end{table}

\textbf{Overall performance.} \chg{Reward models remain below human performance, though some are competitive with MLLMs on \texttt{Top-1}.}{Reward models also remain well below human experts, although some are competitive with MLLMs on \texttt{Top-1}.}
\chg{Q-Align achieves the strongest overall result at 52.2\% on \texttt{Top-1} and 38.2\% on \texttt{TB-1}, versus 77.7\% and 68.9\% for human experts.}{Q-Align is the strongest overall, reaching 52.2\% on \texttt{Top-1} and 38.2\% on \texttt{TB-1}, against 77.7\% and 68.9\% for the human baseline.}

\textbf{\chg{Domain trend and failure asymmetry.}{Domain trends and asymmetric failure modes.}} Domain trends broadly mirror the MLLM results.
\chg{Appendix~\ref{app:rm_full_results} reveals a different failure asymmetry:}{Appendix~\ref{app:rm_full_results} additionally reveals an asymmetry between best and worst predictions:}
Q-Instruct reaches 51.5\% on overall \texttt{Bot-1} and 77.0\% on photography \texttt{Bot-1}, despite \chg{much}{considerably} weaker \texttt{Top-1} scores of 38.5\% overall and 43.9\% in photography.
\rev{This pattern suggests that detecting aesthetic failure is in some cases easier than identifying the strongest image in a set.}

\textbf{Comparison \chg{to}{with} MLLMs.} The best reward-model \texttt{Top-1} result \chg{(52.2\%)}{, 52.2\%,} is numerically close to the best MLLM \texttt{Top-1} \texttt{ap@1} result \chg{(55.4\%),}{, 55.4\%,} \chg{but the comparison is not like-for-like because MLLMs make a single comparative decision over a candidate set whereas reward models score images independently.}{but the comparison is not strictly like-for-like: MLLMs make a single comparative decision over a candidate set, whereas reward models score images independently and are then ranked within the set.}

\subsection{Ablations and Additional Analysis}
\label{subsec:ablation}

\textbf{Temperature ablation.} For models with an exposed temperature parameter, greedy decoding improves \texttt{Top-1} \texttt{pass\textasciicircum{}3} by 1.1 points on average and \texttt{TB-1} \texttt{pass\textasciicircum{}3} by 2.3 points. Decoding stochasticity therefore contributes noise, but it does not explain the main performance gap; full model-by-model results are reported in Appendix~\ref{app:temp_ablation}.

\textbf{Candidate set size.} Performance degrades sharply as the candidate set grows. Under \texttt{TB-1} \texttt{pass\textasciicircum{}3}, the best model falls from 47.3\% on two-image tasks to 6.7\% on four-image tasks, while human experts decline from 87.1\% to 43.6\%. Larger candidate sets therefore amplify both comparative difficulty and instability; the full breakdown is reported in Appendix~\ref{app:imgcount}.

\section{Fine-Tuning on Expert Aesthetic Judgments}
\label{sec:finetune}

\paragraph{Training Setup.} We fine-tune \textsc{Qwen3.5-35B-A3B} using Low-Rank Adaptation (LoRA) \citep{hu2022lora} with rank 64 and $\alpha=128$, freezing the vision tower while updating the multimodal projector. Supervision comes from 2,000 additional expert-annotated examples collected with the same annotation pipeline described in Section~\ref{subsec:gt}; these examples are fully disjoint from the benchmark evaluation set. We train for three epochs with a learning rate of $1 \times 10^{-4}$ and cosine scheduling. Additional hyperparameters and full fine-tuning results are provided in Appendix~\ref{app:finetune}. The fine-tuned model is denoted as \model{}.

\begin{table}[t]
\centering
\caption{Fine-tuning results on \bench{}. We compare open-weight baselines with \model{}, which is obtained by LoRA fine-tuning on 2{,}000 expert-annotated training examples. All models are evaluated under their default inference configuration. Best result in each column is shown in \bestcell{bold}.}
\label{tab:finetune:results}
\footnotesize
\setlength{\tabcolsep}{5pt}
\renewcommand{\arraystretch}{1.05}
\begin{tabular}{@{}c c@{\hspace{18pt}}c@{\hspace{11pt}} c c@{\hspace{18pt}}c@{\hspace{11pt}}@{}}
\toprule
& \multicolumn{2}{c}{\texttt{Top-1}} && \multicolumn{2}{c}{\texttt{TB-1}} \\
\cmidrule(r){2-3} \cmidrule(l){5-6}
Model & ap@1 & p\textasciicircum{}3 && ap@1 & p\textasciicircum{}3 \\
\midrule
\multicolumn{6}{c}{\textit{Open-Weight Models}} \\
\midrule
\iconGLM\textsc{GLM-4.6V} & 43.0 & 17.5 && 34.1 & 11.5 \\
\iconKimi\textsc{Kimi-K2.5} & 49.4 & 26.8 && 36.8 & 15.0 \\
\iconQwenVL\textsc{Qwen3-VL-235B-A22B} & 44.9 & 20.5 && 35.2 & 14.0 \\
\iconQwenThreeFive\textsc{Qwen3.5-397B-A17B} & 50.1 & \bestcell{29.8} && \bestcell{38.9} & 17.2 \\
\midrule
\multicolumn{6}{c}{\textit{Ours}} \\
\midrule
\iconQwenThreeFive\textsc{Qwen3.5-35B-A3B} (base) & 47.5 & 25.5 && 36.5 & 15.8 \\
\iconKallisti\model{} & \bestcell{51.3}\rlap{\,\tiny\textcolor{red}{+3.8}} & 29.5\rlap{\,\tiny\textcolor{red}{+4.0}} && 38.8\rlap{\,\tiny\textcolor{red}{+2.3}} & \bestcell{17.5}\rlap{\,\tiny\textcolor{red}{+1.7}} \\
\bottomrule
\end{tabular}
\vspace{-1.0em}
\end{table}

\paragraph{\del{Finding 1: }Improvement over the base model.} Table~\ref{tab:finetune:results} compares \model{} against its base model and a set of open-weight baselines. Fine-tuning yields consistent gains across all \rev{four }reported metrics: \texttt{Top-1} \texttt{pass\textasciicircum{}3} improves from 25.5\% to 29.5\%, \texttt{TB-1} \texttt{pass\textasciicircum{}3} from 15.8\% to 17.5\%, \texttt{Top-1} \texttt{ap@1} from 47.5\% to 51.3\%, and \texttt{TB-1} \texttt{ap@1} from 36.5\% to 38.8\%. \chg{Appendix~\ref{app:finetune_results} reports the topic-level breakdown under all four main-text metrics.}{The topic-level breakdown is reported in Appendix~\ref{app:finetune_results}.}

\paragraph{\del{Finding 2: }Scaling implication.}
\model{} approaches the substantially larger Qwen3.5-397B-A17B on \texttt{Top-1} \texttt{pass\textasciicircum{}3}, \chg{reaching 29.5\% versus 29.8\%.}{29.5\% versus 29.8\%, and lies within roughly one point on the other metrics.}
\chg{This provides preliminary evidence that expert comparative judgments collected for \bench{} can transfer to smaller open-weight models through lightweight adaptation.}{This is preliminary evidence that the comparative supervision in \bench{} carries enough aesthetic signal to compensate for an order-of-magnitude reduction in parameter count under lightweight adaptation.}

\section{Conclusion}

\chg{This paper introduced \bench{}, a benchmark that formulates aesthetic evaluation as comparative selection over candidate sets with matched subject matter, grounded in expert ranking rather than scalar scoring.}{We introduced \bench{}, a benchmark that recasts aesthetic evaluation as comparative selection over candidate sets with matched subject matter, with labels grounded in expert ranking rather than scalar scoring.}
\chg{\bench{} comprised 400 tasks and 1{,}195 images across fine art, photography, and illustration,}{\bench{} contains 400 tasks and 1{,}195 images spanning fine art, photography, and illustration.}
\chg{Our controlled human study showed that direct ranking yields 42 percentage points higher inter-annotator agreement than score-derived rankings, motivating the benchmark's annotation protocol.}{A controlled human study motivates the protocol: direct ranking yields 42 percentage points higher inter-annotator agreement than score-derived rankings.}
\chg{An evaluation of 20 frontier MLLMs and six reward models on \bench{} revealed a 42 point gap to human experts on \texttt{TB-1} \texttt{pass\textasciicircum{}3}, severe positional instability, and sharp degradation as candidate set size grows.}{Across 20 frontier MLLMs and six reward models, the strongest system reaches 26.5\% on \texttt{TB-1} \texttt{pass\textasciicircum{}3} against 68.9\% for human experts, and accuracy drops sharply as the candidate set grows; models also exhibit pronounced positional instability across permutations of the candidate order.}
\rev{Fine-tuning a 35B-parameter model on 2{,}000 expert examples brings its accuracy close to that of a 397B-parameter open-weight model, suggesting that expert comparative supervision transfers efficiently with lightweight adaptation.}
\chg{Our future work will expand domain and annotator diversity, improving positional robustness, and scaling expert comparative supervision are promising directions.}{We see several directions for future work: broadening domain and annotator diversity beyond the present three domains, improving robustness to candidate-order permutations, and scaling expert comparative supervision to larger training corpora. By making expert aesthetic judgment a measurable, comparative target, \bench{} establishes the foundation on which expert-aligned multimodal aesthetic judgment can now be systematically pursued.}

\section*{Ethics Statement}
This study involved human participants providing aesthetic judgments through an online evaluation platform. The University of Washington Human Subjects Division (HSD) reviewed the protocol and determined that the research qualifies as exempt human subjects research (Category 2 and 3; minimal-risk behavioral research) under U.S. federal regulations. Participation was voluntary and participants could stop at any time. No sensitive personal information was collected. Only aggregated benchmark results are publicly released, and no identifiable participant data are included in any publications or datasets. The exempt determination letter is on file with the authors.

All fine art works were created as commissioned works for \bench{}, with participating artists informed of and consenting to the use of their work for research.
Professional photographers whose images served as source material for the photography pipeline contributed their work under appropriate licensing terms. 
Agent-edited variants derived from these source photographs were produced within the scope of those agreements. 
Illustration sets generated through text-to-image models used structured prompts without reproducing copyrighted material, and 3D-rendered sets were constructed exclusively from CC0-licensed assets. 
We are committed to crediting and fairly compensating all human creators whose work contributed to this benchmark.

\bibliography{references}

\clearpage
\appendix

\section*{Appendix}

\startcontents[appendices]
\titlecontents{section}[0em]{}{\thecontentslabel\hspace{0.5em}}{}{\dotfill\contentspage}
\titlecontents{subsection}[1.5em]{}{\thecontentslabel\hspace{0.5em}}{}{\dotfill\contentspage}
\printcontents[appendices]{l}{1}{\setcounter{tocdepth}{2}}

\vspace{1em}

\section{Related Work}
\label{sec:relatedwork}

\paragraph{Multimodal Large Language Models.}
Recent MLLMs such as Flamingo, BLIP-2, InstructBLIP, LLaVA, Gemini, and Qwen2-VL have rapidly improved multimodal perception, instruction following, and reasoning \citep{alayrac2022flamingo,li2023blip,dai2023instructblip,liu2023visual,team2023gemini,wang2024qwen2}. Their capabilities are typically assessed on broad benchmarks such as MM-Vet and MMMU \citep{yu2023mm,yue2024mmmu}, where each example usually admits a single correct answer. Aesthetic judgment is different: it is comparative, convention-sensitive, and not reducible to factual correctness alone. Our focus is therefore whether frontier MLLMs can make stable, expert-aligned selections over candidate sets with matched subject matter rather than pointwise judgments on isolated images.

\paragraph{Image Aesthetics Assessment.}
Image aesthetics assessment (IAA) has traditionally been posed as single-image score prediction. Canonical datasets such as AVA and AADB provide large-scale photographic images with scalar ratings.
Later resources extend this paradigm through comments, theme-aware labels, and personalization, including AVA-Reviews, TAD66K, and PARA \citep{murray2012ava,kong2016photo,wang2019neural,he2022rethinking,yang2022personalized}. These resources support a broad line of regression and ranking models, including NIMA, A-Lamp, theme-aware predictors, and more recent vision-language scorers such as Q-Align and UniQA \citep{talebi2018nima,ma2017lamp,he2022rethinking,wu2023q,zhou2024uniqa}. Still, the dominant setup remains pointwise, mapping from one image to a score, with preference order inferred only indirectly from independent estimates.

More recently, MLLM-based efforts have moved toward richer aesthetic reasoning. AesBench evaluates aesthetic perception in frontier MLLMs; AesExpert and UNIAA build instruction-tuned systems for critique and assessment; and ArtiMuse and the photography-focused \emph{The Photographer's Eye} move closer to expert-style analysis \citep{huang2024aesbench,huang2024aesexpert,zhou2024uniaa,cao2025artimuse,qi2025photographer}. Nevertheless, existing resources remain largely single-image, score-centric, or domain-specific. In contrast, \bench{} is explicitly set-based: each example is a group with matched subject matter, and the task is to identify the strongest and weakest images under controlled content. This formulation targets relative aesthetic discrimination directly, derives labels from expert consensus rather than crowd averages, and spans fine art, photography, and illustration within a unified benchmark.

\section{Benchmark Details}
\label{app:benchmark_details}

\subsection{Final Benchmark Statistics}
\label{app:benchmark_stats}

Table~\ref{tab:vab:stats} reports the final benchmark statistics after expert annotation and consensus filtering.

\begin{table}[t]
\centering
\footnotesize
\setlength{\tabcolsep}{2.5pt}
\renewcommand{\arraystretch}{0.95}
\caption{Final benchmark statistics for \bench{} by domain and topic after expert consensus filtering. T/I denotes the number of tasks and images.}
\label{tab:vab:stats}
\begin{tabular}{@{}cc c c cc c c cc c@{}}
\toprule
\multicolumn{2}{c}{\textbf{Fine Art} (161/458)} & & &
\multicolumn{2}{c}{\textbf{Photography} (139/389)} & & &
\multicolumn{2}{c}{\textbf{Illustration} (100/348)} \\
\cmidrule(r){1-2} \cmidrule(lr){5-6} \cmidrule(l){9-10}
Topic & \!\!T / I & & & Topic & \!\!T / I & & & Topic & \!\!T / I \\
\midrule
Landscape Color   & 56/243 & & & Landscape        & 45/130 & & & Digital \& AI Art & 23/84 \\
Portrait Sketch   & 33/66  & & & Street \& City   & 39/106 & & & Concept Art       & 22/88 \\
Still Life Color  & 24/53  & & & Architecture     & 32/89  & & & Pixel Art         & 21/84 \\
Still Life Sketch & 14/28  & & & Wildlife         & 9/24   & & & Anime \& Manga    & 13/26 \\
Calligraphy       & 11/22  & & & Food \& Product  & 5/14   & & & Stylized 3D       & 11/44 \\
Chinese Painting  & 9/18   & & & Macro            & 4/12   & & & Comic             & 10/22 \\
Quick Sketch      & 8/16   & & & Night \& Astro   & 2/6    & & & & \\
Ink \& Wash       & 4/8    & & & Portrait         & 2/5    & & & & \\
Portrait Color    & 2/4    & & & Sports           & 1/3    & & & & \\
\bottomrule
\end{tabular}
\end{table}

Table~\ref{tab:app:candidate_dist} shows the candidate set size distribution in the final benchmark by domain, and Table~\ref{tab:app:topic_dist} provides the corresponding topic-level breakdown.

\begin{table}[t]
\centering
\caption{Candidate set size distribution in the final benchmark by domain.}
\label{tab:app:candidate_dist}
\footnotesize
\begin{tabular}{@{}cccccc@{}}
\toprule
Domain & $k=2$ & $k=3$ & $k=4$ & $k=5$ & $k=6$ \\
\midrule
Fine Art & 109 & 4 & 13 & 34 & 1 \\
Photography & 30 & 107 & 2 & 0 & 0 \\
Illustration & 26 & 0 & 74 & 0 & 0 \\
\midrule
Total & 165 & 111 & 89 & 34 & 1 \\
\bottomrule
\end{tabular}
\end{table}

\begin{table}[t]
\centering
\caption{Topic-level breakdown of the final benchmark after expert consensus filtering.}
\label{tab:app:topic_dist}
\footnotesize
\setlength{\tabcolsep}{4pt}
\renewcommand{\arraystretch}{0.95}
\begin{tabular}{@{}ccccccccc@{}}
\toprule
Domain & Topic & $k=2$ & $k=3$ & $k=4$ & $k=5$ & $k=6$ & Tasks & Avg.\ $k$ \\
\midrule
Fine Art & Calligraphy & 11 & 0 & 0 & 0 & 0 & 11 & 2.00 \\
Fine Art & Chinese Painting & 9 & 0 & 0 & 0 & 0 & 9 & 2.00 \\
Fine Art & Ink and Wash & 4 & 0 & 0 & 0 & 0 & 4 & 2.00 \\
Fine Art & Landscape Color & 6 & 4 & 12 & 33 & 1 & 56 & 4.34 \\
Fine Art & Portrait Color & 2 & 0 & 0 & 0 & 0 & 2 & 2.00 \\
Fine Art & Portrait Sketch & 33 & 0 & 0 & 0 & 0 & 33 & 2.00 \\
Fine Art & Quick Sketch & 8 & 0 & 0 & 0 & 0 & 8 & 2.00 \\
Fine Art & Still Life Color & 22 & 0 & 1 & 1 & 0 & 24 & 2.21 \\
Fine Art & Still Life Sketch & 14 & 0 & 0 & 0 & 0 & 14 & 2.00 \\
\midrule
Photography & Architecture & 7 & 25 & 0 & 0 & 0 & 32 & 2.78 \\
Photography & Food and Product & 1 & 4 & 0 & 0 & 0 & 5 & 2.80 \\
Photography & Landscape & 6 & 38 & 1 & 0 & 0 & 45 & 2.89 \\
Photography & Macro & 0 & 4 & 0 & 0 & 0 & 4 & 3.00 \\
Photography & Night and Astrophotography & 0 & 2 & 0 & 0 & 0 & 2 & 3.00 \\
Photography & Portrait & 1 & 1 & 0 & 0 & 0 & 2 & 2.50 \\
Photography & Sports & 0 & 1 & 0 & 0 & 0 & 1 & 3.00 \\
Photography & Street and City & 11 & 28 & 0 & 0 & 0 & 39 & 2.72 \\
Photography & Wildlife & 4 & 4 & 1 & 0 & 0 & 9 & 2.67 \\
\midrule
Illustration & Anime and Manga & 13 & 0 & 0 & 0 & 0 & 13 & 2.00 \\
Illustration & Comic & 9 & 0 & 1 & 0 & 0 & 10 & 2.20 \\
Illustration & Concept Art & 0 & 0 & 22 & 0 & 0 & 22 & 4.00 \\
Illustration & Digital and AI Art & 4 & 0 & 19 & 0 & 0 & 23 & 3.65 \\
Illustration & Pixel Art & 0 & 0 & 21 & 0 & 0 & 21 & 4.00 \\
Illustration & Stylized 3D & 0 & 0 & 11 & 0 & 0 & 11 & 4.00 \\
\bottomrule
\end{tabular}
\end{table}

\subsection{Raw Collection Statistics Before Expert Filtering}
\label{app:benchmark_raw_stats}

Before expert filtering, the collected pool contains 1,367 tasks and 3,859 images. Table~\ref{tab:app:domain_stats} summarizes the raw collection by domain, Table~\ref{tab:app:raw_topic_dist} reports the corresponding topic-level breakdown, and Table~\ref{tab:app:raw_photo_pipeline} further decomposes the photography subset by collection pipeline.

\begin{table}[t]
\centering
\caption{Raw collection statistics by domain before expert filtering.}
\label{tab:app:domain_stats}
\footnotesize
\setlength{\tabcolsep}{6pt}
\begin{tabular}{@{}cccc@{}}
\toprule
Domain & Tasks & Images & Avg. $k$ \\
\midrule
Fine Art & 426 & 1,142 & 2.68 \\
Photography & 670 & 1,809 & 2.70 \\
Illustration & 271 & 908 & 3.35 \\
\midrule
Total & 1,367 & 3,859 & 2.82 \\
\bottomrule
\end{tabular}
\end{table}

\begin{table}[t]
\centering
\caption{Topic-level raw collection statistics before expert filtering.}
\label{tab:app:raw_topic_dist}
\scriptsize
\setlength{\tabcolsep}{4pt}
\renewcommand{\arraystretch}{0.95}
\resizebox{\textwidth}{!}{%
\begin{tabular}{@{}cccccccccc@{}}
\toprule
Domain & Topic & Tasks & Images & $k=2$ & $k=3$ & $k=4$ & $k=5$ & $k=6$ & Avg.\ $k$ \\
\midrule
Fine Art & Landscape Color & 113 & 502 & 12 & 6 & 20 & 70 & 5 & 4.44 \\
Fine Art & Portrait Sketch & 93 & 186 & 93 & 0 & 0 & 0 & 0 & 2.00 \\
Fine Art & Still Life Color & 73 & 160 & 68 & 0 & 1 & 4 & 0 & 2.19 \\
Fine Art & Still Life Sketch & 43 & 86 & 43 & 0 & 0 & 0 & 0 & 2.00 \\
Fine Art & Calligraphy & 44 & 88 & 44 & 0 & 0 & 0 & 0 & 2.00 \\
Fine Art & Quick Sketch & 30 & 60 & 30 & 0 & 0 & 0 & 0 & 2.00 \\
Fine Art & Chinese Painting & 13 & 26 & 13 & 0 & 0 & 0 & 0 & 2.00 \\
Fine Art & Ink and Wash & 10 & 20 & 10 & 0 & 0 & 0 & 0 & 2.00 \\
Fine Art & Portrait Color & 7 & 14 & 7 & 0 & 0 & 0 & 0 & 2.00 \\
\midrule
Photography & Landscape & 280 & 761 & 81 & 197 & 2 & 0 & 0 & 2.72 \\
Photography & Street and City & 165 & 450 & 45 & 120 & 0 & 0 & 0 & 2.73 \\
Photography & Architecture & 157 & 420 & 51 & 106 & 0 & 0 & 0 & 2.68 \\
Photography & Wildlife & 24 & 62 & 11 & 12 & 1 & 0 & 0 & 2.58 \\
Photography & Portrait & 13 & 31 & 8 & 5 & 0 & 0 & 0 & 2.38 \\
Photography & Macro & 13 & 36 & 3 & 10 & 0 & 0 & 0 & 2.77 \\
Photography & Night and Astrophotography & 7 & 20 & 1 & 6 & 0 & 0 & 0 & 2.86 \\
Photography & Food and Product & 7 & 18 & 3 & 4 & 0 & 0 & 0 & 2.57 \\
Photography & Sports & 4 & 11 & 1 & 3 & 0 & 0 & 0 & 2.75 \\
\midrule
Illustration & Digital and AI Art & 51 & 184 & 10 & 0 & 41 & 0 & 0 & 3.61 \\
Illustration & Concept Art & 50 & 200 & 0 & 0 & 50 & 0 & 0 & 4.00 \\
Illustration & Stylized 3D & 50 & 200 & 0 & 0 & 50 & 0 & 0 & 4.00 \\
Illustration & Pixel Art & 40 & 160 & 0 & 0 & 40 & 0 & 0 & 4.00 \\
Illustration & Anime and Manga & 40 & 82 & 39 & 0 & 1 & 0 & 0 & 2.05 \\
Illustration & Comic & 40 & 82 & 39 & 0 & 1 & 0 & 0 & 2.05 \\
\bottomrule
\end{tabular}%
}
\end{table}

\begin{table}[t]
\centering
\caption{Raw photography collection statistics by topic and collection pipeline before expert filtering. Agent-Edit and Expert-Edit denote task counts from the automated editing pipeline and the expert retouching pipeline, respectively.}
\label{tab:app:raw_photo_pipeline}
\scriptsize
\setlength{\tabcolsep}{4pt}
\renewcommand{\arraystretch}{0.95}
\begin{tabular}{@{}ccccccccc@{}}
\toprule
Topic & Tasks & Images & $k=2$ & $k=3$ & $k=4$ & Avg.\ $k$ & Agent-Edit & Expert-Edit \\
\midrule
Landscape & 280 & 761 & 81 & 197 & 2 & 2.72 & 188 & 92 \\
Street and City & 165 & 450 & 45 & 120 & 0 & 2.73 & 115 & 50 \\
Architecture & 157 & 420 & 51 & 106 & 0 & 2.68 & 104 & 53 \\
Wildlife & 24 & 62 & 11 & 12 & 1 & 2.58 & 10 & 14 \\
Portrait & 13 & 31 & 8 & 5 & 0 & 2.38 & 5 & 8 \\
Macro & 13 & 36 & 3 & 10 & 0 & 2.77 & 10 & 3 \\
Night and Astrophotography & 7 & 20 & 1 & 6 & 0 & 2.86 & 5 & 2 \\
Food and Product & 7 & 18 & 3 & 4 & 0 & 2.57 & 4 & 3 \\
Sports & 4 & 11 & 1 & 3 & 0 & 2.75 & 3 & 1 \\
\bottomrule
\end{tabular}
\end{table}

\subsection{Data Construction Details}
\label{app:benchmark_construction}

\paragraph{Fine Art.}
We commissioned 426 painting sets comprising 1,142 images across nine fine-art topics: calligraphy, Chinese painting, ink and wash, landscape color, portrait color, portrait sketch, quick sketch, still-life color, and still-life sketch. For each topic, artists received constrained prompts that fixed the subject and high-level compositional requirements while allowing variation in execution. The resulting sets preserve semantic intent while differing in composition, value structure, color harmony, and paint handling. All works were commissioned specifically for \bench{} rather than drawn from existing collections, which reduces the risk of contamination from publicly indexed art. Before annotation, we remove near-duplicates and discard sets with semantic mismatches or trivial cues.

\paragraph{Photography.}
We constructed 670 photography sets comprising 1,809 images across nine topics: architecture, food and product, landscape, macro, night and astrophotography, portrait, sports, street and city, and wildlife. Each set originates from a single source photograph, many provided directly by professional photographers. We transform these source images into controlled comparison sets through two pipelines.

In the \emph{expert-edit pipeline}, photographers and retouchers improve each source image using tools such as Adobe Photoshop and Lightroom \citep{adobe2025photoshop,adobe2025lightroom}, applying recomposition, color and tone correction, and content-aware expansion. The original photograph and its improved variant form a pair. For a subset of source images, multiple experts independently edit the same photograph, yielding larger candidate sets with matched subject matter and varying aesthetic quality.

In the \emph{agent-edit pipeline}, each source image is first sent to a critic model, ArtiMuse \citep{cao2025artimuse}, which produces a detailed aesthetic review. The review and the original image are then passed to a general-purpose MLLM, GPT-5 \citep{openai2025gpt5}, to generate two editing prompts: one for improvement and one for degradation. These prompts, together with the original image, are fed to an image-to-image model, Gemini-3-Pro-Image (Nano Banana Pro) \citep{google2025nanobananapro}, to generate an improved and a degraded variant. The original, improved, and degraded images form a three-image candidate set. The exact prompt sequence is recorded in Appendix~\ref{app:prompts:photo}.

All photography sets from both pipelines are deduplicated and reviewed for semantic consistency before annotation.

\paragraph{Illustration.}
We constructed 271 illustration sets comprising 908 images across six topics: anime and manga, comic, concept art, digital and AI art, pixel art, and stylized 3D. The sets are derived from two distinct sources, prompt-based generation and 3D asset rendering.

In the \emph{generative pipeline}, each set originates from a structured prompt that specifies subject identity, scene context, composition, camera viewpoint, lighting, and stylistic attributes. For each set, the prompt is instantiated using a single text-to-image model, such as Midjourney v7 \citep{midjourney2025v7}, Pixellab.ai (Create M-XL) \citep{pixellab2025}, or LlamaGen \citep{sun2024autoregressive}, to generate candidate images with matched subject matter and varying aesthetic quality. The prompt templates are provided in Appendix~\ref{app:prompts:illustration}.

In the \emph{3D-rendering pipeline}, each set is constructed from CC0-licensed assets sourced from Sketchfab \citep{creativecommonscc02009,sketchfab2025}. For each asset, we render multiple images by varying camera viewpoint while keeping object identity, scene semantics, lighting, and background conditions fixed. The resulting renders form candidate sets with matched subject matter and controlled variation in composition and visual presentation.

All illustration sets from both pipelines are deduplicated and reviewed for semantic consistency before annotation.

\clearpage
\section{Annotation Details}
\label{app:rubric}

\subsection{Annotation Rubrics}

The rubric serves as a structured comparison guide rather than a separate numerical scoring form. Judges assess each image against domain-specific criteria before making their final comparative selection. The rubrics below describe the criteria used for each domain.

\begin{rubricbox}[title=Fine Art Rubric]
Given a set of 2 to 6 fine art paintings with the same style, subject matter, and medium, and only subtle differences, determine which work demonstrates greater artistic completeness and more mature aesthetic expression.

\textbf{Core Comparison Criteria}
\begin{enumerate}[leftmargin=1.5em]
    \item \textbf{Composition and Visual Order}
    \begin{itemize}[leftmargin=1.5em]
        \item Is the primary subject more clearly defined?
        \item Does the composition feel more stable or intentionally dynamic?
        \item Is the spatial arrangement natural, without unnecessary congestion or emptiness?
    \end{itemize}
    \item \textbf{Color and Tonal Relationships}
    \begin{itemize}[leftmargin=1.5em]
        \item Are the color relationships more harmonious or layered?
        \item Is there stronger control over warmth/coolness and light/dark contrast?
        \item Does the work avoid muddiness, dullness, or scattered color impressions?
    \end{itemize}
    \item \textbf{Technical Control}
    \begin{itemize}[leftmargin=1.5em]
        \item Are the brushstrokes more confident, fluid, and rhythmic?
        \item Are there signs of hesitation, redundancy, or ineffective marks?
        \item Does the technique serve the image rather than draw attention to itself?
    \end{itemize}
    \item \textbf{Degree of Completion}
    \begin{itemize}[leftmargin=1.5em]
        \item Are there any areas that appear unfinished or perfunctory?
        \item Are edges and transitions handled naturally?
        \item Does the work feel resolved, at a point where the artist could reasonably stop?
    \end{itemize}
    \item \textbf{Artistic Expression and Overall Character}
    \begin{itemize}[leftmargin=1.5em]
        \item Is the mood or atmosphere more focused and coherent?
        \item Does the work convey a stronger overall character or presence?
        \item Does it have greater artistic persuasiveness?
    \end{itemize}
    \item \textbf{Final Selection}
    \begin{itemize}[leftmargin=1.5em]
        \item Considering the set as a whole, which image is the strongest work?
        \item If the set contains three or more images, which image is the weakest work?
    \end{itemize}
\end{enumerate}

\textbf{Should Not Be Considered}
\begin{itemize}[leftmargin=1.5em]
    \item Personal preference for subject matter.
    \item Any contextual or background information beyond what is visible in the image itself.
\end{itemize}
\end{rubricbox}

\begin{rubricbox}[title=Photography Rubric]
Given a set of 2 to 6 photographs taken of similar subjects, in similar scenes, and under closely matched conditions, determine which image is more successful in terms of photographic language and visual expression.

\textbf{Core Comparison Criteria}
\begin{enumerate}[leftmargin=1.5em]
    \item \textbf{Composition and Image Structure}
    \begin{itemize}[leftmargin=1.5em]
        \item Is the subject more clearly emphasized?
        \item Is the frame more balanced or intentionally dynamic?
        \item Is the background cleaner and more supportive of the subject?
    \end{itemize}
    \item \textbf{Quality of Light}
    \begin{itemize}[leftmargin=1.5em]
        \item Is the light more natural or more expressive?
        \item Are highlights and shadows better controlled?
        \item Does the image avoid crushed blacks or blown highlights?
    \end{itemize}
    \item \textbf{Sharpness and Focus}
    \begin{itemize}[leftmargin=1.5em]
        \item Is focus more accurate where it matters?
        \item Is the relationship between sharpness and blur more appropriate?
        \item Are there any technical errors that detract from the image?
    \end{itemize}
    \item \textbf{Color or Tonal Rendering}
    \begin{itemize}[leftmargin=1.5em]
        \item Is white balance or color temperature more accurate or more intentional?
        \item Are colors or tonal values more consistent and unified?
        \item Does the image avoid unwanted color casts or muddy tones?
    \end{itemize}
    \item \textbf{Moment and Emotional Expression}
    \begin{itemize}[leftmargin=1.5em]
        \item Does the image capture a stronger or more meaningful moment?
        \item Is the emotional or narrative content more compelling?
        \item Does it carry a stronger sense of photographic presence?
    \end{itemize}
    \item \textbf{Final Selection}
    \begin{itemize}[leftmargin=1.5em]
        \item Considering the set as a whole, which photograph is the strongest image?
        \item If the set contains three or more images, which photograph is the weakest image?
    \end{itemize}
\end{enumerate}

\textbf{Should Not Be Considered}
\begin{itemize}[leftmargin=1.5em]
    \item The personal appeal or attractiveness of the subject.
    \item Assumptions about shooting difficulty.
    \item Assumptions about post-processing effort.
\end{itemize}
\end{rubricbox}

\begin{rubricbox}[title=Digital Illustration Rubric]
Given a set of 2 to 6 digital illustrations with similar purpose, style, and level of completion, determine which image demonstrates stronger visual design and higher professional illustration quality.

\textbf{Core Comparison Criteria}
\begin{enumerate}[leftmargin=1.5em]
    \item \textbf{Visual Focus and Information Clarity}
    \begin{itemize}[leftmargin=1.5em]
        \item Is the visual focal point clearer?
        \item Does the visual flow guide the viewer more effectively?
        \item Is there less visual noise or competition for attention?
    \end{itemize}
    \item \textbf{Form and Proportion}
    \begin{itemize}[leftmargin=1.5em]
        \item Are the structures of figures or objects more accurate?
        \item Are poses and silhouettes more aesthetically refined?
        \item Does the image avoid stiffness, distortion, or unnatural forms?
    \end{itemize}
    \item \textbf{Color Design}
    \begin{itemize}[leftmargin=1.5em]
        \item Is the color palette more unified and thematically coherent?
        \item Are value and lighting hierarchies clearer?
        \item Is there effective use of contrast and visual rhythm?
    \end{itemize}
    \item \textbf{Detail Control and Finish}
    \begin{itemize}[leftmargin=1.5em]
        \item Are details handled appropriately, neither excessive nor careless?
        \item Are materials and lighting consistent throughout?
        \item Are edges, masks, and gradients clean and well controlled?
    \end{itemize}
    \item \textbf{Stylistic Consistency and Purpose Fit}
    \begin{itemize}[leftmargin=1.5em]
        \item Does the illustration better align with its intended stylistic direction?
        \item Is quality more consistent within a unified style?
        \item Is the image more suitable for its potential use case?
    \end{itemize}
    \item \textbf{Final Selection}
    \begin{itemize}[leftmargin=1.5em]
        \item Considering the set as a whole, which illustration is the strongest work?
        \item If the set contains three or more images, which illustration is the weakest work?
    \end{itemize}
\end{enumerate}

\textbf{Should Not Be Considered}
\begin{itemize}[leftmargin=1.5em]
    \item Personal preference for illustration style.
    \item Software or tools used.
    \item Hypothetical or imagined client requirements.
\end{itemize}
\end{rubricbox}

\clearpage
\subsection{Annotation Interface}

\begin{figure}[H]
\centering
\includegraphics[width=\linewidth]{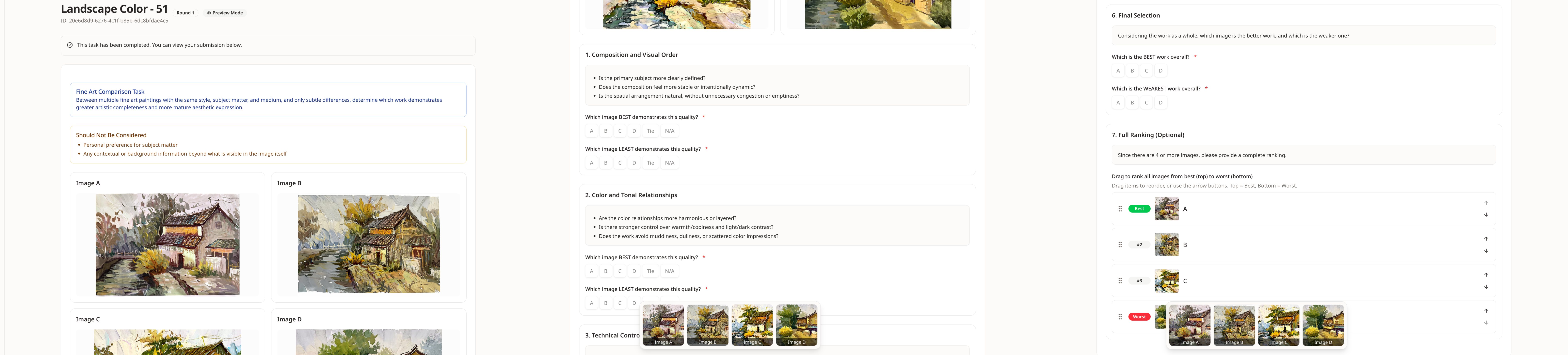}
\caption{The annotation interface presents all $k$ candidate images side by side. Judges first review each image with respect to the rubric criteria, then make a final comparative selection: selecting the best image for $k=2$ and selecting both the best and worst images for $k \ge 3$.}
\label{fig:app:annotation_interface}
\end{figure}

Figure~\ref{fig:app:annotation_interface} shows an example annotation interface used by expert judges.

\section{Ground Truth Mathematical Details}
\label{app:gt_math}

This appendix provides additional mathematical details for the consensus filter described in Section~\ref{subsec:gt}.

\subsection{Selected Thresholds}

Table~\ref{tab:gt:filter} reports the consensus thresholds used in the benchmark together with their corresponding null pass probabilities.

\begin{table}[t]
\centering
\small
\caption{Consensus thresholds and corresponding null pass probabilities by candidate set size $k$, with $n=10$ judges per task.}
\label{tab:gt:filter}
\begin{tabular}{c c c}
\toprule
$k$ & Threshold $m$ & $p_{\mathrm{accept}}(k, m)$ \\
\midrule
2 & 8 & 10.94\% \\
3 & 7 & 0.88\% \\
4 & 6 & 0.95\% \\
5 & 6 & 0.15\% \\
6 & 6 & 0.03\% \\
\bottomrule
\end{tabular}
\end{table}

\subsection{Derivation of Null Pass Probabilities}

For $k \ge 3$ candidates with $n = 10$ judges, each judge's response is modeled as a directed edge $i \to j$ (selecting image $i$ as best and image $j$ as worst, with $i \ne j$) drawn uniformly from $k(k-1)$ possibilities. Let $c_{ij}$ denote the count of judges choosing the pair $(i, j)$. The probability of any configuration $C$ under the random null is:
\begin{equation*}
  \Pr(C) = \frac{n!}{\prod_{i \ne j} c_{ij}!} \left(\frac{1}{k(k-1)}\right)^n.
\end{equation*}

The exact by-chance pass probability is computed by summing over all configurations satisfying the consensus threshold:
\begin{equation*}
  p_{\mathrm{accept}}(k, m) = \sum_{\substack{C:\, c_{ii}=0,\\ \sum_{i \ne j} c_{ij}=n}}
  \frac{n!}{\prod_{i \ne j} c_{ij}!} \left(\frac{1}{k(k-1)}\right)^n
  \mathbf{1}\!\left(\max_i B_i \ge m\right)
  \mathbf{1}\!\left(\max_j W_j \ge m\right),
\end{equation*}
where $B_i = \sum_{j \ne i} c_{ij}$ (votes for $i$ as best) and $W_j = \sum_{i \ne j} c_{ij}$ (votes for $j$ as worst).
The exact probabilities were computed by exhaustive enumeration over all valid count configurations $C=(c_{ij})$ satisfying $c_{ii}=0$ and $\sum_{i \ne j} c_{ij}=n$, summing the multinomial probability mass over the configurations that satisfy the acceptance condition. For $k \ge 3$, this acceptance condition is conjunctive: a task passes only if the best-vote threshold and the worst-vote threshold are both met, so satisfying only one of the two is insufficient.

\subsection{Monte Carlo Verification}

We verify the exact probabilities via Monte Carlo simulation with $10^7$ samples per $(k, m)$ configuration. Table~\ref{tab:app:mc_verify} compares the exact values from Table~\ref{tab:gt:filter} against Monte Carlo estimates and reports 95\% Wald confidence intervals.

\begin{table}[t]
\centering
\caption{Monte Carlo verification of null pass probabilities ($10^7$ samples).}
\label{tab:app:mc_verify}
\footnotesize
\begin{tabular}{@{}ccccc@{}}
\toprule
$k$ & $m$ & Exact & MC Estimate & 95\% CI \\
\midrule
2 & 8 & 10.94\% & 10.9496\% & [10.9303\%, 10.9690\%] \\
3 & 7 & 0.88\% & 0.8818\% & [0.8760\%, 0.8876\%] \\
4 & 6 & 0.95\% & 0.9507\% & [0.9447\%, 0.9567\%] \\
5 & 6 & 0.15\% & 0.1489\% & [0.1465\%, 0.1513\%] \\
6 & 6 & 0.03\% & 0.0307\% & [0.0296\%, 0.0318\%] \\
\bottomrule
\end{tabular}
\end{table}

\subsection{Special Case for Binary Tasks}

For binary tasks ($k = 2$), the worst image is implied once the best is selected. The consensus filter reduces to requiring $\max_i B_i \ge m$ where $B \sim \mathrm{Binomial}(n, 1/2)$:
\begin{equation*}
  p_{\mathrm{accept}}(2, m) = \Pr\!\left(B \ge m\right) + \Pr\!\left(B \le n - m\right) = 2 \sum_{b=m}^{n} \binom{n}{b} \left(\frac{1}{2}\right)^n.
\end{equation*}
For $n = 10$ and $m = 8$, this gives $p_{\mathrm{accept}}(2, 8) = 2 \times (45 + 10 + 1) / 1024 = 112/1024 \approx 10.94\%$. We use this lower binary threshold because $k=2$ tasks provide only a single comparative decision per annotator; compared with $k \ge 3$, demanding the same level of concentration would remove a disproportionate number of otherwise informative binary comparisons.

\section{Detailed Experimental Results}
\label{app:detailed_results}

This appendix supplements Section~\ref{sec:experiments} with full result tables that are too large for the main text. We retain the same evaluation focus as in the main paper: overall performance relative to expert baselines, cross-domain variation, sensitivity to candidate order, and auxiliary analyses on decoding and task size.

\subsection{Comprehensive MLLM Results}
\label{app:full_results}

Table~\ref{tab:app:mllm_full_results} reports comprehensive MLLM results, including \texttt{Bot-1}, across overall and domain-level evaluations. These results complement Table~\ref{tab:mllm:results} in the main text by making the asymmetry between identifying the strongest and weakest images explicit. They also make clear that the gap between \texttt{ap@1} and \texttt{pass\textasciicircum{}3} is pervasive rather than concentrated in a small subset of models, consistent with the interpretation in Section~\ref{subsec:results:mllm} that many MLLM predictions remain sensitive to candidate order and response instability.

The full table also makes several model-level patterns easier to inspect. Within the Claude family, Claude Sonnet 4.6 improves substantially over Claude Sonnet 4.5 on the stricter \texttt{TB-1} \texttt{pass\textasciicircum{}3} metric, whereas the GPT-5 series declines monotonically from GPT-5 to GPT-5.2. The same table also makes the closed/open gap clearer under stability-sensitive metrics: Qwen3.5-397B-A17B is much closer to the strongest closed models under \texttt{ap@1} than under \texttt{pass\textasciicircum{}3}, suggesting weaker consistency across repeated evaluations. Finally, the domain columns reinforce the main-text headline that illustration is the most difficult domain, photography is generally more tractable, and fine art lies between them.

\begin{table}[t]
\centering
\caption{Comprehensive \bench{} results for evaluated MLLMs and reference baselines. We report \texttt{Top-1}, \texttt{Bot-1}, and \texttt{TB-1} accuracy under both \texttt{ap@1} and \texttt{pass\textasciicircum{}3} (\%). Best result in each non-baseline column is shown in \textbf{bold}.}
\label{tab:app:mllm_full_results}
\resizebox{\textwidth}{!}{%
\footnotesize
\setlength{\tabcolsep}{2.2pt}
\renewcommand{\arraystretch}{1.03}
\begin{tabular}{@{}c *{6}{c} c *{6}{c} c *{6}{c} c *{6}{c}@{}}
\toprule
& \multicolumn{6}{c}{Overall} && \multicolumn{6}{c}{Fine Art} && \multicolumn{6}{c}{Illustration} && \multicolumn{6}{c}{Photography} \\
\cmidrule(r){2-7} \cmidrule(lr){9-14} \cmidrule(lr){16-21} \cmidrule(l){23-28}
& \multicolumn{2}{c}{\texttt{Top-1}} & \multicolumn{2}{c}{\texttt{Bot-1}} & \multicolumn{2}{c}{\texttt{TB-1}} && \multicolumn{2}{c}{\texttt{Top-1}} & \multicolumn{2}{c}{\texttt{Bot-1}} & \multicolumn{2}{c}{\texttt{TB-1}} && \multicolumn{2}{c}{\texttt{Top-1}} & \multicolumn{2}{c}{\texttt{Bot-1}} & \multicolumn{2}{c}{\texttt{TB-1}} && \multicolumn{2}{c}{\texttt{Top-1}} & \multicolumn{2}{c}{\texttt{Bot-1}} & \multicolumn{2}{c}{\texttt{TB-1}} \\
\cmidrule(r){2-3} \cmidrule(r){4-5} \cmidrule(r){6-7} \cmidrule(lr){9-10} \cmidrule(r){11-12} \cmidrule(r){13-14} \cmidrule(lr){16-17} \cmidrule(r){18-19} \cmidrule(r){20-21} \cmidrule(l){23-24} \cmidrule(r){25-26} \cmidrule(l){27-28}
Model & ap@1 & p\textasciicircum{}3 & ap@1 & p\textasciicircum{}3 & ap@1 & p\textasciicircum{}3 && ap@1 & p\textasciicircum{}3 & ap@1 & p\textasciicircum{}3 & ap@1 & p\textasciicircum{}3 && ap@1 & p\textasciicircum{}3 & ap@1 & p\textasciicircum{}3 & ap@1 & p\textasciicircum{}3 && ap@1 & p\textasciicircum{}3 & ap@1 & p\textasciicircum{}3 & ap@1 & p\textasciicircum{}3 \\
\midrule
\rowcolor{gray!10} Random Guess & 37.2 & 6.6 & 37.2 & 6.6 & 27.5 & 5.3 && 41.0 & 8.9 & 41.0 & 8.9 & 36.0 & 8.5 && 31.5 & 4.4 & 31.5 & 4.4 & 19.2 & 3.3 && 36.8 & 5.6 & 36.8 & 5.6 & 23.7 & 3.1 \\
\rowcolor{gray!10} Human Expert & \multicolumn{2}{c}{77.7} & \multicolumn{2}{c}{80.1} & \multicolumn{2}{c}{68.9} && \multicolumn{2}{c}{80.4} & \multicolumn{2}{c}{82.7} & \multicolumn{2}{c}{74.7} && \multicolumn{2}{c}{70.3} & \multicolumn{2}{c}{70.4} & \multicolumn{2}{c}{54.4} && \multicolumn{2}{c}{79.9} & \multicolumn{2}{c}{84.0} & \multicolumn{2}{c}{72.4} \\
\multicolumn{28}{c}{\textit{Closed-Source Models}} \\
\bottomrule
\end{tabular}%
}
\end{table}

\subsection{Comprehensive Reward-Model Results}
\label{app:rm_full_results}

Table~\ref{tab:app:rm_full_results} reports comprehensive reward-model results, including \texttt{Bot-1}, across overall and domain-level evaluations. In contrast to the condensed main-text presentation in Table~\ref{tab:rm:results}, this table shows that some reward models are materially stronger at rejecting poor images than at identifying the best one. In particular, the detailed \texttt{Bot-1} columns make the failure mode discussed in Section~\ref{subsec:results:rm} directly visible, especially for Q-Instruct in photography.

\begin{table}[t]
\centering
\footnotesize
\setlength{\tabcolsep}{2.2pt}
\renewcommand{\arraystretch}{1.03}
\caption{Comprehensive reward-model results on \bench{}. Each image is scored independently three times and the scores are averaged before within-task ranking. We report \texttt{Top-1}, \texttt{Bot-1}, and \texttt{TB-1} accuracy across overall and domain-level evaluations (\%). Best result in each non-baseline column is shown in \textbf{bold}.}
\label{tab:app:rm_full_results}
\begin{tabular}{@{}c *{3}{c} c *{3}{c} c *{3}{c} c *{3}{c}@{}}
\toprule
& \multicolumn{3}{c}{Overall} && \multicolumn{3}{c}{Fine Art} && \multicolumn{3}{c}{Illustration} && \multicolumn{3}{c}{Photography} \\
\cmidrule(r){2-4} \cmidrule(lr){6-8} \cmidrule(lr){10-12} \cmidrule(l){14-16}
Model & Top-1 & Bot-1 & TB-1 && Top-1 & Bot-1 & TB-1 && Top-1 & Bot-1 & TB-1 && Top-1 & Bot-1 & TB-1 \\
\midrule
\rowcolor{gray!10} Random Guess & 37.2 & 37.2 & 27.5 && 41.0 & 41.0 & 36.0 && 31.5 & 31.5 & 19.2 && 36.8 & 36.8 & 23.7 \\
\rowcolor{gray!10} Human Expert & 77.7 & 80.1 & 68.9 && 80.4 & 82.7 & 74.7 && 70.3 & 70.4 & 54.4 && 79.9 & 84.0 & 72.4 \\
\midrule
\textsc{ArtiMuse} & 44.2 & 51.2 & 34.8 && 50.9 & \textbf{54.7} & \textbf{46.0} && \textbf{37.0} & 33.0 & 20.0 && 41.7 & 60.4 & 32.4 \\
\textsc{PEAS-aes} & 48.5 & 49.5 & 37.8 && 49.7 & 51.6 & 44.1 && 34.0 & \textbf{36.0} & \textbf{22.0} && 57.6 & 56.8 & 41.7 \\
\textsc{PEAS-ava} & 49.0 & 48.2 & 37.0 && 50.9 & 49.1 & 44.1 && 25.0 & 31.0 & 13.0 && 64.0 & 59.7 & 46.0 \\
\textsc{Q-Align} & \textbf{52.2} & 47.8 & \textbf{38.2} && \textbf{52.2} & 49.7 & 44.7 && 31.0 & 28.0 & 16.0 && \textbf{67.6} & 59.7 & \textbf{46.8} \\
\textsc{Q-Instruct} & 38.5 & \textbf{51.5} & 31.2 && 37.9 & 39.1 & 34.2 && 32.0 & \textbf{36.0} & 21.0 && 43.9 & \textbf{77.0} & 35.2 \\
\textsc{RealQA-ava} & 44.2 & 42.2 & 30.2 && 40.4 & 43.5 & 36.6 && 33.0 & 27.0 & 15.0 && 56.8 & 51.8 & 33.8 \\
\bottomrule
\end{tabular}
\end{table}

\subsection{Topic-Level MLLM Results}
\label{app:subdomain_results}

To match the main-text MLLM table, we report topic-level results under all four metrics: \texttt{Top-1} \texttt{ap@1}, \texttt{Top-1} \texttt{pass\textasciicircum{}3}, \texttt{TB-1} \texttt{ap@1}, and \texttt{TB-1} \texttt{pass\textasciicircum{}3}. Together, these tables show that the domain-level trends in Section~\ref{subsec:results:mllm} are not driven by a single topic or a single evaluation metric. Illustration remains the hardest domain across all four views, while photography is generally more tractable but still highly uneven across topics.

Several topic effects recur across metrics. In fine art, Chinese painting and portrait-color tasks are often relatively tractable, whereas landscape-color and some sketch-based tasks become much less stable under the stricter criteria. In illustration, the remaining signal is concentrated mostly in anime-and-manga and comic tasks, while concept art, digital-and-AI art, and stylized 3D frequently collapse under \texttt{pass\textasciicircum{}3}. In photography, macro and sports tasks are often close to saturation for strong models, but night-and-astrophotography, architecture, and some street-and-city tasks remain much less stable, especially when consistency across repeated evaluations is required.

\medskip
\noindent\textbf{\texttt{Top-1} \texttt{ap@1}.}\par
Tables~\ref{tab:app:topic_top1_ap1_fine-art}, \ref{tab:app:topic_top1_ap1_illustration}, and \ref{tab:app:topic_top1_ap1_photography} report topic-level \texttt{Top-1} \texttt{ap@1} results for fine art, illustration, and photography, respectively.

\begin{table}[t]
\centering
\caption{Topic-level MLLM \texttt{Top-1} \texttt{ap@1} results on fine-art tasks. The final column reports the corresponding domain-level average (\%).}
\label{tab:app:topic_top1_ap1_fine-art}
\resizebox{\textwidth}{!}{%
\scriptsize
\setlength{\tabcolsep}{3pt}
\renewcommand{\arraystretch}{1.02}
\begin{tabular}{@{}c c c c c c c c c c c@{}}
\toprule
Model & Calligraphy & Chinese Painting & Ink \& Wash & Landscape Color & Portrait Color & Portrait Sketch & Quick Sketch & Still Life Color & Still Life Sketch & Avg. \\
\midrule
\rowcolor{gray!10} Random Guess & 50.0 & 50.0 & 50.0 & 25.2 & 50.0 & 50.0 & 50.0 & 47.7 & 50.0 & 41.0 \\
\rowcolor{gray!10} Human Expert & 89.1 & 94.4 & 90.0 & 66.6 & 85.0 & 90.3 & 88.8 & 81.7 & 86.5 & 80.4 \\
\multicolumn{11}{c}{\textit{Closed-Source Models}} \\
\midrule
\iconClaude\textsc{Claude-Haiku-4.5} & 33.3 & 81.5 & 66.7 & 29.8 & 50.0 & 53.5 & 54.2 & 52.8 & 42.9 & 44.7 \\
\iconClaude\textsc{Claude-Sonnet-4.5} & 42.4 & 85.2 & 66.7 & 27.4 & 50.0 & 54.5 & 62.5 & 48.6 & 40.5 & 44.5 \\
\iconClaude\textsc{Claude-Sonnet-4.6} & 54.5 & 74.1 & 50.0 & 39.9 & 83.3 & 48.5 & 62.5 & 48.6 & 47.6 & 48.4 \\
\iconClaude\textsc{Claude-Opus-4.5} & 42.4 & 81.5 & 50.0 & 32.7 & 83.3 & 46.5 & 58.3 & 51.4 & 47.6 & 45.3 \\
\bottomrule
\end{tabular}%
}
\end{table}

\begin{table}[t]
\centering
\caption{Topic-level MLLM \texttt{Top-1} \texttt{ap@1} results on illustration tasks. The final column reports the corresponding domain-level average (\%).}
\label{tab:app:topic_top1_ap1_illustration}
\resizebox{\textwidth}{!}{%
\scriptsize
\setlength{\tabcolsep}{3pt}
\renewcommand{\arraystretch}{1.02}
\begin{tabular}{@{}c c c c c c c c@{}}
\toprule
Model & Anime \& Manga & Comic & Concept Art & Digital \& AI Art & Pixel Art & Stylized 3D & Avg. \\
\midrule
\rowcolor{gray!10} Random Guess & 50.0 & 47.5 & 25.0 & 29.3 & 25.0 & 25.0 & 31.5 \\
\rowcolor{gray!10} Human Expert & 87.7 & 82.0 & 66.8 & 65.2 & 62.4 & 71.8 & 70.3 \\
\multicolumn{8}{c}{\textit{Closed-Source Models}} \\
\midrule
\iconClaude\textsc{Claude-Haiku-4.5} & 56.4 & 26.7 & 40.9 & 20.3 & 25.4 & 33.3 & 32.7 \\
\iconClaude\textsc{Claude-Sonnet-4.5} & 61.5 & 26.7 & 33.3 & 33.3 & 41.3 & 27.3 & 37.3 \\
\iconClaude\textsc{Claude-Sonnet-4.6} & 71.8 & 63.3 & 28.8 & 37.7 & 46.0 & 45.5 & 45.3 \\
\iconClaude\textsc{Claude-Opus-4.5} & 64.1 & 46.7 & 27.3 & 31.9 & 42.9 & 42.4 & 40.0 \\
\bottomrule
\end{tabular}%
}
\end{table}

\begin{table}[t]
\centering
\caption{Topic-level MLLM \texttt{Top-1} \texttt{ap@1} results on photography tasks. The final column reports the corresponding domain-level average (\%).}
\label{tab:app:topic_top1_ap1_photography}
\resizebox{\textwidth}{!}{%
\scriptsize
\setlength{\tabcolsep}{3pt}
\renewcommand{\arraystretch}{1.02}
\begin{tabular}{@{}c c c c c c c c c c c@{}}
\toprule
Model & Architecture & Food \& Product & Landscape & Macro & Night \& Astro & Portrait & Sports & Street \& City & Wildlife & Avg. \\
\midrule
\rowcolor{gray!10} Random Guess & 37.0 & 36.7 & 35.4 & 33.3 & 33.3 & 41.7 & 33.3 & 38.0 & 39.8 & 36.8 \\
\rowcolor{gray!10} Human Expert & 81.6 & 78.0 & 78.9 & 75.0 & 80.0 & 85.0 & 90.0 & 80.3 & 78.9 & 79.9 \\
\multicolumn{11}{c}{\textit{Closed-Source Models}} \\
\midrule
\iconClaude\textsc{Claude-Haiku-4.5} & 40.6 & 60.0 & 49.6 & 83.3 & 0.0 & 66.7 & 100.0 & 47.9 & 66.7 & 49.4 \\
\iconClaude\textsc{Claude-Sonnet-4.5} & 43.8 & 40.0 & 61.5 & 83.3 & 66.7 & 33.3 & 100.0 & 58.1 & 81.5 & 57.6 \\
\iconClaude\textsc{Claude-Sonnet-4.6} & 58.3 & 46.7 & 75.6 & 91.7 & 50.0 & 100.0 & 100.0 & 70.1 & 63.0 & 68.8 \\
\iconClaude\textsc{Claude-Opus-4.5} & 58.3 & 46.7 & 69.6 & 83.3 & 50.0 & 83.3 & 100.0 & 66.7 & 70.4 & 65.9 \\
\bottomrule
\end{tabular}%
}
\end{table}

\medskip
\noindent\textbf{\texttt{Top-1} \texttt{pass\textasciicircum{}3}.}\par
Tables~\ref{tab:app:topic_top1_p3_fine-art}, \ref{tab:app:topic_top1_p3_illustration}, and \ref{tab:app:topic_top1_p3_photography} report topic-level \texttt{Top-1} \texttt{pass\textasciicircum{}3} results for fine art, illustration, and photography, respectively.

\begin{table}[t]
\centering
\caption{Topic-level MLLM \texttt{Top-1} \texttt{pass\textasciicircum{}3} results on fine-art tasks. The final column reports the corresponding domain-level average (\%).}
\label{tab:app:topic_top1_p3_fine-art}
\resizebox{\textwidth}{!}{%
\scriptsize
\setlength{\tabcolsep}{3pt}
\renewcommand{\arraystretch}{1.02}
\begin{tabular}{@{}c c c c c c c c c c c@{}}
\toprule
Model & Calligraphy & Chinese Painting & Ink \& Wash & Landscape Color & Portrait Color & Portrait Sketch & Quick Sketch & Still Life Color & Still Life Sketch & Avg. \\
\midrule
\rowcolor{gray!10} Random Guess & 12.5 & 12.5 & 12.5 & 2.4 & 12.5 & 12.5 & 12.5 & 11.6 & 12.5 & 8.9 \\
\rowcolor{gray!10} Human Expert & 89.1 & 94.4 & 90.0 & 66.6 & 85.0 & 90.3 & 88.8 & 81.7 & 86.5 & 80.4 \\
\multicolumn{11}{c}{\textit{Closed-Source Models}} \\
\midrule
\iconClaude\textsc{Claude-Haiku-4.5} & 9.1 & 66.7 & 50.0 & 5.4 & 50.0 & 24.2 & 50.0 & 16.7 & 21.4 & 19.9 \\
\iconClaude\textsc{Claude-Sonnet-4.5} & 9.1 & 77.8 & 25.0 & 7.1 & 0.0 & 33.3 & 37.5 & 20.8 & 7.1 & 20.5 \\
\iconClaude\textsc{Claude-Sonnet-4.6} & 36.4 & 66.7 & 50.0 & 28.6 & 50.0 & 36.4 & 62.5 & 37.5 & 35.7 & 37.3 \\
\iconClaude\textsc{Claude-Opus-4.5} & 36.4 & 77.8 & 50.0 & 16.1 & 50.0 & 21.2 & 37.5 & 37.5 & 21.4 & 28.0 \\
\bottomrule
\end{tabular}%
}
\end{table}

\begin{table}[t]
\centering
\caption{Topic-level MLLM \texttt{Top-1} \texttt{pass\textasciicircum{}3} results on illustration tasks. The final column reports the corresponding domain-level average (\%).}
\label{tab:app:topic_top1_p3_illustration}
\resizebox{\textwidth}{!}{%
\scriptsize
\setlength{\tabcolsep}{3pt}
\renewcommand{\arraystretch}{1.02}
\begin{tabular}{@{}c c c c c c c c@{}}
\toprule
Model & Anime \& Manga & Comic & Concept Art & Digital \& AI Art & Pixel Art & Stylized 3D & Avg. \\
\midrule
\rowcolor{gray!10} Random Guess & 12.5 & 11.4 & 1.6 & 3.5 & 1.6 & 1.6 & 4.4 \\
\rowcolor{gray!10} Human Expert & 87.7 & 82.0 & 66.8 & 65.2 & 62.4 & 71.8 & 70.3 \\
\multicolumn{8}{c}{\textit{Closed-Source Models}} \\
\midrule
\iconClaude\textsc{Claude-Haiku-4.5} & 30.8 & 0.0 & 18.2 & 4.3 & 9.5 & 18.2 & 13.0 \\
\iconClaude\textsc{Claude-Sonnet-4.5} & 30.8 & 0.0 & 9.1 & 17.4 & 23.8 & 18.2 & 17.0 \\
\iconClaude\textsc{Claude-Sonnet-4.6} & 61.5 & 60.0 & 4.5 & 26.1 & 28.6 & 27.3 & 30.0 \\
\iconClaude\textsc{Claude-Opus-4.5} & 53.8 & 30.0 & 4.5 & 21.7 & 33.3 & 36.4 & 27.0 \\
\bottomrule
\end{tabular}%
}
\end{table}

\begin{table}[t]
\centering
\caption{Topic-level MLLM \texttt{Top-1} \texttt{pass\textasciicircum{}3} results on photography tasks. The final column reports the corresponding domain-level average (\%).}
\label{tab:app:topic_top1_p3_photography}
\resizebox{\textwidth}{!}{%
\scriptsize
\setlength{\tabcolsep}{3pt}
\renewcommand{\arraystretch}{1.02}
\begin{tabular}{@{}c c c c c c c c c c c@{}}
\toprule
Model & Architecture & Food \& Product & Landscape & Macro & Night \& Astro & Portrait & Sports & Street \& City & Wildlife & Avg. \\
\midrule
\rowcolor{gray!10} Random Guess & 5.6 & 5.5 & 4.8 & 3.7 & 3.7 & 8.1 & 3.7 & 6.2 & 7.4 & 5.6 \\
\rowcolor{gray!10} Human Expert & 81.6 & 78.0 & 78.9 & 75.0 & 80.0 & 85.0 & 90.0 & 80.3 & 78.9 & 79.9 \\
\multicolumn{11}{c}{\textit{Closed-Source Models}} \\
\midrule
\iconClaude\textsc{Claude-Haiku-4.5} & 15.6 & 20.0 & 24.4 & 75.0 & 0.0 & 0.0 & 100.0 & 15.4 & 33.3 & 21.6 \\
\iconClaude\textsc{Claude-Sonnet-4.5} & 15.6 & 0.0 & 37.8 & 75.0 & 50.0 & 0.0 & 100.0 & 35.9 & 66.7 & 33.8 \\
\iconClaude\textsc{Claude-Sonnet-4.6} & 28.1 & 20.0 & 60.0 & 75.0 & 50.0 & 100.0 & 100.0 & 56.4 & 55.6 & 51.1 \\
\iconClaude\textsc{Claude-Opus-4.5} & 31.2 & 0.0 & 53.3 & 75.0 & 50.0 & 50.0 & 100.0 & 48.7 & 66.7 & 46.8 \\
\bottomrule
\end{tabular}%
}
\end{table}

\medskip
\noindent\textbf{\texttt{TB-1} \texttt{ap@1}.}\par
Tables~\ref{tab:app:topic_tb1_ap1_fine-art}, \ref{tab:app:topic_tb1_ap1_illustration}, and \ref{tab:app:topic_tb1_ap1_photography} report topic-level \texttt{TB-1} \texttt{ap@1} results for fine art, illustration, and photography, respectively.

\begin{table}[t]
\centering
\caption{Topic-level MLLM \texttt{TB-1} \texttt{ap@1} results on fine-art tasks. The final column reports the corresponding domain-level average (\%).}
\label{tab:app:topic_tb1_ap1_fine-art}
\resizebox{\textwidth}{!}{%
\scriptsize
\setlength{\tabcolsep}{3pt}
\renewcommand{\arraystretch}{1.02}
\begin{tabular}{@{}c c c c c c c c c c c@{}}
\toprule
Model & Calligraphy & Chinese Painting & Ink \& Wash & Landscape Color & Portrait Color & Portrait Sketch & Quick Sketch & Still Life Color & Still Life Sketch & Avg. \\
\midrule
\rowcolor{gray!10} Random Guess & 50.0 & 50.0 & 50.0 & 11.3 & 50.0 & 50.0 & 50.0 & 46.4 & 50.0 & 36.0 \\
\rowcolor{gray!10} Human Expert & 89.1 & 94.4 & 90.0 & 50.5 & 85.0 & 90.3 & 88.8 & 80.8 & 86.5 & 74.7 \\
\multicolumn{11}{c}{\textit{Closed-Source Models}} \\
\midrule
\iconClaude\textsc{Claude-Haiku-4.5} & 33.3 & 81.5 & 66.7 & 17.9 & 50.0 & 53.5 & 54.2 & 51.4 & 42.9 & 40.4 \\
\iconClaude\textsc{Claude-Sonnet-4.5} & 42.4 & 85.2 & 66.7 & 15.5 & 50.0 & 54.5 & 62.5 & 47.2 & 40.5 & 40.2 \\
\iconClaude\textsc{Claude-Sonnet-4.6} & 54.5 & 74.1 & 50.0 & 28.6 & 83.3 & 48.5 & 62.5 & 50.0 & 47.6 & 44.7 \\
\iconClaude\textsc{Claude-Opus-4.5} & 42.4 & 81.5 & 50.0 & 19.0 & 83.3 & 46.5 & 58.3 & 50.0 & 47.6 & 40.4 \\
\bottomrule
\end{tabular}%
}
\end{table}

\begin{table}[t]
\centering
\caption{Topic-level MLLM \texttt{TB-1} \texttt{ap@1} results on illustration tasks. The final column reports the corresponding domain-level average (\%).}
\label{tab:app:topic_tb1_ap1_illustration}
\resizebox{\textwidth}{!}{%
\scriptsize
\setlength{\tabcolsep}{3pt}
\renewcommand{\arraystretch}{1.02}
\begin{tabular}{@{}c c c c c c c c@{}}
\toprule
Model & Anime \& Manga & Comic & Concept Art & Digital \& AI Art & Pixel Art & Stylized 3D & Avg. \\
\midrule
\rowcolor{gray!10} Random Guess & 50.0 & 45.8 & 8.3 & 15.6 & 8.3 & 8.3 & 19.2 \\
\rowcolor{gray!10} Human Expert & 87.7 & 79.0 & 42.3 & 49.1 & 43.3 & 49.1 & 54.4 \\
\multicolumn{8}{c}{\textit{Closed-Source Models}} \\
\midrule
\iconClaude\textsc{Claude-Haiku-4.5} & 56.4 & 26.7 & 13.6 & 15.9 & 9.5 & 0.0 & 18.7 \\
\iconClaude\textsc{Claude-Sonnet-4.5} & 61.5 & 23.3 & 10.6 & 17.4 & 12.7 & 12.1 & 20.7 \\
\iconClaude\textsc{Claude-Sonnet-4.6} & 71.8 & 66.7 & 16.7 & 26.1 & 15.9 & 3.0 & 29.3 \\
\iconClaude\textsc{Claude-Opus-4.5} & 64.1 & 43.3 & 13.6 & 20.3 & 17.5 & 12.1 & 25.3 \\
\bottomrule
\end{tabular}%
}
\end{table}

\begin{table}[t]
\centering
\caption{Topic-level MLLM \texttt{TB-1} \texttt{ap@1} results on photography tasks. The final column reports the corresponding domain-level average (\%).}
\label{tab:app:topic_tb1_ap1_photography}
\resizebox{\textwidth}{!}{%
\scriptsize
\setlength{\tabcolsep}{3pt}
\renewcommand{\arraystretch}{1.02}
\begin{tabular}{@{}c c c c c c c c c c c@{}}
\toprule
Model & Architecture & Food \& Product & Landscape & Macro & Night \& Astro & Portrait & Sports & Street \& City & Wildlife & Avg. \\
\midrule
\rowcolor{gray!10} Random Guess & 24.0 & 23.3 & 20.9 & 16.7 & 16.7 & 33.3 & 16.7 & 26.1 & 30.6 & 23.7 \\
\rowcolor{gray!10} Human Expert & 73.8 & 72.0 & 71.3 & 60.0 & 60.0 & 80.0 & 70.0 & 74.4 & 72.2 & 72.4 \\
\multicolumn{11}{c}{\textit{Closed-Source Models}} \\
\midrule
\iconClaude\textsc{Claude-Haiku-4.5} & 21.9 & 60.0 & 32.6 & 41.7 & 0.0 & 66.7 & 33.3 & 35.0 & 66.7 & 34.3 \\
\iconClaude\textsc{Claude-Sonnet-4.5} & 33.3 & 33.3 & 40.7 & 41.7 & 33.3 & 16.7 & 33.3 & 47.9 & 55.6 & 41.2 \\
\iconClaude\textsc{Claude-Sonnet-4.6} & 34.4 & 26.7 & 41.5 & 58.3 & 50.0 & 100.0 & 0.0 & 50.4 & 55.6 & 43.9 \\
\iconClaude\textsc{Claude-Opus-4.5} & 42.7 & 46.7 & 44.4 & 66.7 & 50.0 & 83.3 & 66.7 & 50.4 & 59.3 & 48.2 \\
\bottomrule
\end{tabular}%
}
\end{table}

\medskip
\noindent\textbf{\texttt{TB-1} \texttt{pass\textasciicircum{}3}.}\par
Tables~\ref{tab:app:topic_fine_art}, \ref{tab:app:topic_illustration}, and \ref{tab:app:topic_photography} report topic-level \texttt{TB-1} \texttt{pass\textasciicircum{}3} results for fine art, illustration, and photography, respectively.

\begin{table}[t]
\centering
\caption{Topic-level \texttt{TB-1} \texttt{pass\textasciicircum{}3} results on fine-art tasks. The final column reports the corresponding domain-level average (\%).}
\label{tab:app:topic_fine_art}
\resizebox{\textwidth}{!}{%
\scriptsize
\setlength{\tabcolsep}{3pt}
\renewcommand{\arraystretch}{1.02}
\begin{tabular}{@{}c c c c c c c c c c c@{}}
\toprule
Model & Calligraphy & Chinese Painting & Ink \& Wash & Landscape Color & Portrait Color & Portrait Sketch & Quick Sketch & Still Life Color & Still Life Sketch & Avg. \\
\midrule
\rowcolor{gray!10} Random Guess & 12.5 & 12.5 & 12.5 & 1.4 & 12.5 & 12.5 & 12.5 & 11.5 & 12.5 & 8.5 \\
\rowcolor{gray!10} Human Expert & 89.1 & 94.4 & 90.0 & 50.5 & 85.0 & 90.3 & 88.8 & 80.8 & 86.5 & 74.7 \\
\multicolumn{11}{c}{\textit{Closed-Source Models}} \\
\midrule
\iconClaude\textsc{Claude-Haiku-4.5} & 9.1 & 66.7 & 50.0 & 3.6 & 50.0 & 24.2 & 50.0 & 16.7 & 21.4 & 19.3 \\
\iconClaude\textsc{Claude-Sonnet-4.5} & 9.1 & 77.8 & 25.0 & 3.6 & 0.0 & 33.3 & 37.5 & 20.8 & 7.1 & 19.3 \\
\iconClaude\textsc{Claude-Sonnet-4.6} & 36.4 & 66.7 & 50.0 & 19.6 & 50.0 & 36.4 & 62.5 & 37.5 & 35.7 & 34.2 \\
\iconClaude\textsc{Claude-Opus-4.5} & 36.4 & 77.8 & 50.0 & 5.4 & 50.0 & 21.2 & 37.5 & 37.5 & 21.4 & 24.2 \\
\bottomrule
\end{tabular}%
}
\end{table}

\begin{table}[t]
\centering
\caption{Topic-level \texttt{TB-1} \texttt{pass\textasciicircum{}3} results on illustration tasks. The final column reports the corresponding domain-level average (\%).}
\label{tab:app:topic_illustration}
\resizebox{\textwidth}{!}{%
\scriptsize
\setlength{\tabcolsep}{3pt}
\renewcommand{\arraystretch}{1.02}
\begin{tabular}{@{}c c c c c c c c@{}}
\toprule
Model & Anime \& Manga & Comic & Concept Art & Digital \& AI Art & Pixel Art & Stylized 3D & Avg. \\
\midrule
\rowcolor{gray!10} Random Guess & 12.5 & 11.3 & 0.1 & 2.2 & 0.1 & 0.1 & 3.3 \\
\rowcolor{gray!10} Human Expert & 87.7 & 79.0 & 42.3 & 49.1 & 43.3 & 49.1 & 54.4 \\
\multicolumn{8}{c}{\textit{Closed-Source Models}} \\
\midrule
\iconClaude\textsc{Claude-Haiku-4.5} & 30.8 & 0.0 & 0.0 & 4.3 & 0.0 & 0.0 & 5.0 \\
\iconClaude\textsc{Claude-Sonnet-4.5} & 30.8 & 0.0 & 4.5 & 8.7 & 4.8 & 0.0 & 8.0 \\
\iconClaude\textsc{Claude-Sonnet-4.6} & 61.5 & 60.0 & 9.1 & 13.0 & 0.0 & 0.0 & 19.0 \\
\iconClaude\textsc{Claude-Opus-4.5} & 53.8 & 30.0 & 4.5 & 13.0 & 0.0 & 0.0 & 14.0 \\
\bottomrule
\end{tabular}%
}
\end{table}

\begin{table}[t]
\centering
\caption{Topic-level \texttt{TB-1} \texttt{pass\textasciicircum{}3} results on photography tasks. The final column reports the corresponding domain-level average (\%).}
\label{tab:app:topic_photography}
\resizebox{\textwidth}{!}{%
\scriptsize
\setlength{\tabcolsep}{3pt}
\renewcommand{\arraystretch}{1.02}
\begin{tabular}{@{}c c c c c c c c c c c@{}}
\toprule
Model & Architecture & Food \& Product & Landscape & Macro & Night \& Astro & Portrait & Sports & Street \& City & Wildlife & Avg. \\
\midrule
\rowcolor{gray!10} Random Guess & 3.1 & 2.9 & 2.1 & 0.5 & 0.5 & 6.5 & 0.5 & 3.9 & 5.8 & 3.1 \\
\rowcolor{gray!10} Human Expert & 73.8 & 72.0 & 71.3 & 60.0 & 60.0 & 80.0 & 70.0 & 74.4 & 72.2 & 72.4 \\
\multicolumn{11}{c}{\textit{Closed-Source Models}} \\
\midrule
\iconClaude\textsc{Claude-Haiku-4.5} & 3.1 & 40.0 & 4.4 & 25.0 & 0.0 & 0.0 & 0.0 & 12.8 & 33.3 & 10.1 \\
\iconClaude\textsc{Claude-Sonnet-4.5} & 9.4 & 0.0 & 13.3 & 25.0 & 0.0 & 0.0 & 0.0 & 15.4 & 33.3 & 13.7 \\
\iconClaude\textsc{Claude-Sonnet-4.6} & 12.5 & 0.0 & 17.8 & 50.0 & 50.0 & 100.0 & 0.0 & 28.2 & 44.4 & 23.0 \\
\iconClaude\textsc{Claude-Opus-4.5} & 18.8 & 20.0 & 15.6 & 25.0 & 50.0 & 50.0 & 0.0 & 17.9 & 44.4 & 20.1 \\
\bottomrule
\end{tabular}%
}
\end{table}

\subsection{Topic-Level Reward-Model Results}
\label{app:rm_subdomain_results}

To match the main-text reward-model table, we report topic-level results for both \texttt{Top-1} and \texttt{TB-1}. As in the MLLM case, the domain-level averages in the main text mask substantial variation across topics. The topic-level reward-model tables show that relatively stronger domain-level results in photography are driven by a subset of topics, while illustration remains difficult across most topics for all six reward models.

The topic breakdown also clarifies where reward models do and do not transfer. Fine-art performance is mixed but not uniformly weak, with several models reaching competitive results on Chinese painting and portrait-color tasks. Illustration remains systematically difficult: outside anime-and-manga and, in a few cases, comic, most models stay close to random on concept art, digital-and-AI art, pixel art, and stylized 3D, especially under \texttt{TB-1}. Photography remains the strongest domain, but the gains are concentrated in a subset of topics and do not transfer uniformly to the harder best-and-worst selection setting.

\medskip
\noindent\textbf{\texttt{Top-1}.}\par
Tables~\ref{tab:app:rm_topic_top1_fine-art}, \ref{tab:app:rm_topic_top1_illustration}, and \ref{tab:app:rm_topic_top1_photography} report topic-level \texttt{Top-1} results for fine art, illustration, and photography, respectively.

\medskip
\noindent\textbf{\texttt{TB-1}.}\par
Tables~\ref{tab:app:rm_topic_fine_art}, \ref{tab:app:rm_topic_illustration}, and \ref{tab:app:rm_topic_photography} report topic-level \texttt{TB-1} results for fine art, illustration, and photography, respectively.

\begin{table}[t]
\centering
\caption{Topic-level reward-model \texttt{Top-1} results on fine-art tasks. The final column reports the corresponding domain-level average (\%).}
\label{tab:app:rm_topic_top1_fine-art}
\resizebox{\textwidth}{!}{%
\scriptsize
\setlength{\tabcolsep}{3pt}
\renewcommand{\arraystretch}{1.02}
\begin{tabular}{@{}c c c c c c c c c c c@{}}
\toprule
Model & Calligraphy & Chinese Painting & Ink \& Wash & Landscape Color & Portrait Color & Portrait Sketch & Quick Sketch & Still Life Color & Still Life Sketch & Avg. \\
\midrule
\rowcolor{gray!10} Random Guess & 50.0 & 50.0 & 50.0 & 25.2 & 50.0 & 50.0 & 50.0 & 47.7 & 50.0 & 41.0 \\
\rowcolor{gray!10} Human Expert & 89.1 & 94.4 & 90.0 & 66.6 & 85.0 & 90.3 & 88.8 & 81.7 & 86.5 & 80.5 \\
\multicolumn{11}{c}{\textit{Reward Models}} \\
\midrule
\textsc{ArtiMuse} & 54.5 & 88.9 & 50.0 & 39.3 & 50.0 & 54.5 & 50.0 & 50.0 & 64.3 & 50.9 \\
\textsc{PEAS-aes} & 54.5 & 44.4 & 50.0 & 30.4 & 100.0 & 72.7 & 50.0 & 62.5 & 42.9 & 49.7 \\
\textsc{PEAS-ava} & 36.4 & 66.7 & 50.0 & 41.1 & 100.0 & 54.5 & 62.5 & 58.3 & 57.1 & 50.9 \\
\textsc{Q-Align} & 63.6 & 66.7 & 50.0 & 39.3 & 100.0 & 48.5 & 50.0 & 66.7 & 64.3 & 52.2 \\
\textsc{Q-Instruct} & 9.1 & 44.4 & 75.0 & 21.4 & 100.0 & 57.6 & 50.0 & 41.7 & 42.9 & 37.9 \\
\textsc{RealQA-ava} & 36.4 & 66.7 & 50.0 & 21.4 & 100.0 & 48.5 & 50.0 & 45.8 & 57.1 & 40.4 \\
\bottomrule
\end{tabular}%
}
\end{table}

\begin{table}[t]
\centering
\caption{Topic-level reward-model \texttt{Top-1} results on illustration tasks. The final column reports the corresponding domain-level average (\%).}
\label{tab:app:rm_topic_top1_illustration}
\resizebox{\textwidth}{!}{%
\scriptsize
\setlength{\tabcolsep}{3pt}
\renewcommand{\arraystretch}{1.02}
\begin{tabular}{@{}c c c c c c c c@{}}
\toprule
Model & Anime \& Manga & Comic & Concept Art & Digital \& AI Art & Pixel Art & Stylized 3D & Avg. \\
\midrule
\rowcolor{gray!10} Random Guess & 50.0 & 47.5 & 25.0 & 29.3 & 25.0 & 25.0 & 31.5 \\
\rowcolor{gray!10} Human Expert & 87.7 & 82.0 & 66.8 & 65.2 & 62.4 & 71.8 & 70.3 \\
\multicolumn{8}{c}{\textit{Reward Models}} \\
\midrule
\textsc{ArtiMuse} & 53.8 & 60.0 & 22.7 & 39.1 & 33.3 & 27.3 & 37.0 \\
\textsc{PEAS-aes} & 30.8 & 20.0 & 22.7 & 43.5 & 47.6 & 27.3 & 34.0 \\
\textsc{PEAS-ava} & 30.8 & 20.0 & 9.1 & 30.4 & 42.9 & 9.1 & 25.0 \\
\textsc{Q-Align} & 38.5 & 10.0 & 36.4 & 21.7 & 38.1 & 36.4 & 31.0 \\
\textsc{Q-Instruct} & 46.2 & 30.0 & 18.2 & 34.8 & 42.9 & 18.2 & 32.0 \\
\textsc{RealQA-ava} & 38.5 & 50.0 & 45.5 & 13.0 & 38.1 & 18.2 & 33.0 \\
\bottomrule
\end{tabular}%
}
\end{table}

\begin{table}[t]
\centering
\caption{Topic-level reward-model \texttt{Top-1} results on photography tasks. The final column reports the corresponding domain-level average (\%).}
\label{tab:app:rm_topic_top1_photography}
\resizebox{\textwidth}{!}{%
\scriptsize
\setlength{\tabcolsep}{3pt}
\renewcommand{\arraystretch}{1.02}
\begin{tabular}{@{}c c c c c c c c c c c@{}}
\toprule
Model & Architecture & Food \& Product & Landscape & Macro & Night \& Astro & Portrait & Sports & Street \& City & Wildlife & Avg. \\
\midrule
\rowcolor{gray!10} Random Guess & 37.0 & 36.7 & 35.4 & 33.3 & 33.3 & 41.7 & 33.3 & 38.0 & 39.8 & 36.8 \\
\rowcolor{gray!10} Human Expert & 81.6 & 78.0 & 78.9 & 75.0 & 80.0 & 85.0 & 90.0 & 80.3 & 78.9 & 79.9 \\
\multicolumn{11}{c}{\textit{Reward Models}} \\
\midrule
\textsc{ArtiMuse} & 25.0 & 60.0 & 46.7 & 50.0 & 100.0 & 50.0 & 0.0 & 38.5 & 66.7 & 41.7 \\
\textsc{PEAS-aes} & 50.0 & 80.0 & 57.8 & 100.0 & 50.0 & 0.0 & 0.0 & 61.5 & 55.6 & 57.6 \\
\textsc{PEAS-ava} & 56.2 & 80.0 & 68.9 & 100.0 & 50.0 & 0.0 & 0.0 & 66.7 & 55.6 & 64.0 \\
\textsc{Q-Align} & 62.5 & 80.0 & 77.8 & 75.0 & 50.0 & 100.0 & 100.0 & 59.0 & 55.6 & 67.6 \\
\textsc{Q-Instruct} & 37.5 & 40.0 & 42.2 & 25.0 & 50.0 & 0.0 & 0.0 & 51.3 & 66.7 & 43.9 \\
\textsc{RealQA-ava} & 46.9 & 60.0 & 57.8 & 100.0 & 50.0 & 50.0 & 100.0 & 59.0 & 55.6 & 56.8 \\
\bottomrule
\end{tabular}%
}
\end{table}

\begin{table}[t]
\centering
\caption{Topic-level reward-model \texttt{TB-1} results on fine-art tasks. The final column reports the corresponding domain-level average (\%).}
\label{tab:app:rm_topic_fine_art}
\resizebox{\textwidth}{!}{%
\scriptsize
\setlength{\tabcolsep}{3pt}
\renewcommand{\arraystretch}{1.02}
\begin{tabular}{@{}c c c c c c c c c c c@{}}
\toprule
Model & Calligraphy & Chinese Painting & Ink \& Wash & Landscape Color & Portrait Color & Portrait Sketch & Quick Sketch & Still Life Color & Still Life Sketch & Avg. \\
\midrule
\rowcolor{gray!10} Random Guess & 50.0 & 50.0 & 50.0 & 11.3 & 50.0 & 50.0 & 50.0 & 46.4 & 50.0 & 36.0 \\
\rowcolor{gray!10} Human Expert & 89.1 & 94.4 & 90.0 & 50.5 & 85.0 & 90.3 & 88.8 & 80.8 & 86.5 & 74.7 \\
\multicolumn{11}{c}{\textit{Reward Models}} \\
\midrule
\textsc{ArtiMuse} & 54.5 & 88.9 & 50.0 & 25.0 & 50.0 & 54.5 & 50.0 & 50.0 & 64.3 & 46.0 \\
\textsc{PEAS-aes} & 54.5 & 44.4 & 50.0 & 16.1 & 100.0 & 72.7 & 50.0 & 58.3 & 42.9 & 44.1 \\
\textsc{PEAS-ava} & 36.4 & 66.7 & 50.0 & 23.2 & 100.0 & 54.5 & 62.5 & 54.2 & 57.1 & 44.1 \\
\textsc{Q-Align} & 63.6 & 66.7 & 50.0 & 19.6 & 100.0 & 48.5 & 50.0 & 62.5 & 64.3 & 44.7 \\
\textsc{Q-Instruct} & 9.1 & 44.4 & 75.0 & 10.7 & 100.0 & 57.6 & 50.0 & 41.7 & 42.9 & 34.2 \\
\textsc{RealQA-ava} & 36.4 & 66.7 & 50.0 & 12.5 & 100.0 & 48.5 & 50.0 & 41.7 & 57.1 & 36.6 \\
\bottomrule
\end{tabular}%
}
\end{table}

\begin{table}[t]
\centering
\caption{Topic-level reward-model \texttt{TB-1} results on illustration tasks. The final column reports the corresponding domain-level average (\%).}
\label{tab:app:rm_topic_illustration}
\resizebox{\textwidth}{!}{%
\scriptsize
\setlength{\tabcolsep}{3pt}
\renewcommand{\arraystretch}{1.02}
\begin{tabular}{@{}c c c c c c c c@{}}
\toprule
Model & Anime \& Manga & Comic & Concept Art & Digital \& AI Art & Pixel Art & Stylized 3D & Avg. \\
\midrule
\rowcolor{gray!10} Random Guess & 50.0 & 45.8 & 8.3 & 15.6 & 8.3 & 8.3 & 19.2 \\
\rowcolor{gray!10} Human Expert & 87.7 & 79.0 & 42.3 & 49.1 & 43.3 & 49.1 & 54.4 \\
\multicolumn{8}{c}{\textit{Reward Models}} \\
\midrule
\textsc{ArtiMuse} & 53.8 & 60.0 & 4.5 & 13.0 & 4.8 & 18.2 & 20.0 \\
\textsc{PEAS-aes} & 30.8 & 20.0 & 9.1 & 30.4 & 28.6 & 9.1 & 22.0 \\
\textsc{PEAS-ava} & 30.8 & 20.0 & 4.5 & 17.4 & 9.5 & 0.0 & 13.0 \\
\textsc{Q-Align} & 38.5 & 10.0 & 13.6 & 13.0 & 19.1 & 0.0 & 16.0 \\
\textsc{Q-Instruct} & 46.2 & 30.0 & 9.1 & 21.7 & 19.1 & 9.1 & 21.0 \\
\textsc{RealQA-ava} & 38.5 & 50.0 & 13.6 & 4.3 & 4.8 & 0.0 & 15.0 \\
\bottomrule
\end{tabular}%
}
\end{table}

\subsection{Temperature Ablation}
\label{app:temp_ablation}

Table~\ref{tab:ablation:temp} reports the full model-by-model temperature ablation referenced in Section~\ref{subsec:ablation}. The overall trend is that greedy decoding modestly improves stability, but the effect size varies substantially across models and does not alter the main conclusion that the dominant limitation is comparative judgment rather than decoding noise alone.

The gains are not uniform. Gemini 3.1 Pro and Claude Sonnet 4.5 benefit the most from greedy decoding, with especially visible improvements on the stricter \texttt{TB-1} \texttt{pass\textasciicircum{}3} metric. A small set of models, including Claude Sonnet 4.6, Doubao Seed 2.0 Pro, GLM-4.6V, and Qwen3.5-397B-A17B, are flat or slightly worse on at least one metric, reinforcing the point that decoding choice matters but does not dominate the benchmark outcome.

\begin{table}[t]
\centering
\footnotesize
\setlength{\tabcolsep}{3pt}
\renewcommand{\arraystretch}{1.05}
\caption{Effect of decoding temperature on MLLM performance. We compare default decoding ($t{=}1$) with greedy decoding ($t{=}0$); $\Delta$ denotes the change from $t{=}1$ to $t{=}0$ (\%).}
\label{tab:ablation:temp}
\begin{tabular}{@{}c cc c cc c@{}}
\toprule
& \multicolumn{3}{c}{\texttt{Top-1} p\textasciicircum{}3} & \multicolumn{3}{c}{\texttt{TB-1} p\textasciicircum{}3} \\
\cmidrule(r){2-4} \cmidrule(l){5-7}
Model & $t{=}1$ & $t{=}0$ & $\Delta$ & $t{=}1$ & $t{=}0$ & $\Delta$ \\
\midrule
\multicolumn{7}{c}{\textit{Closed-Source Models}} \\
\midrule
\iconClaude\textsc{Claude-Haiku-4.5}     & 18.8 & 21.2 & \small+2.4 & 12.5 & 17.2 & \small+4.7 \\
\iconClaude\textsc{Claude-Sonnet-4.5}    & 24.2 & 27.3 & \small+3.1 & 14.5 & 20.5 & \small+6.0 \\
\iconClaude\textsc{Claude-Sonnet-4.6}    & 40.2 & 39.2 & \small$-$1.0 & 26.5 & 26.8 & \small+0.3 \\
\iconClaude\textsc{Claude-Opus-4.5}      & 34.2 & 35.2 & \small+1.0 & 20.2 & 22.5 & \small+2.3 \\
\iconClaude\textsc{Claude-Opus-4.6}      & 35.5 & 36.5 & \small+1.0 & 20.0 & 21.5 & \small+1.5 \\
\bottomrule
\end{tabular}
\end{table}

\begin{table}[t]
\centering
\caption{\texttt{TB-1} results by candidate set size $k$ for the evaluated MLLMs and reference baselines. For each set size, we report \texttt{pass\textasciicircum{}3} and \texttt{ap@1} (\%).}
\label{tab:imgcount:results}
\resizebox{\textwidth}{!}{%
\footnotesize
\setlength{\tabcolsep}{3pt}
\renewcommand{\arraystretch}{0.95}
\begin{tabular}{@{}c *{2}{c} c *{2}{c} c *{2}{c} c *{2}{c} c *{2}{c}@{}}
\toprule
& \multicolumn{2}{c}{$k=2$} && \multicolumn{2}{c}{$k=3$} && \multicolumn{2}{c}{$k=4$} && \multicolumn{2}{c}{$k=5$} && \multicolumn{2}{c}{$k=6$} \\
\cmidrule(r){2-3} \cmidrule(lr){5-6} \cmidrule(lr){8-9} \cmidrule(lr){11-12} \cmidrule(l){14-15}
Model & p\textasciicircum{}3 & ap@1 && p\textasciicircum{}3 & ap@1 && p\textasciicircum{}3 & ap@1 && p\textasciicircum{}3 & ap@1 && p\textasciicircum{}3 & ap@1 \\
\midrule
\rowcolor{gray!10} Random Guess & 12.5 & 50.0 && 0.5 & 16.7 && 0.1 & 8.3 && 0.0 & 5.0 && 0.0 & 3.3 \\
\rowcolor{gray!10} Human Expert & \multicolumn{2}{c}{87.1} && \multicolumn{2}{c}{69.3} && \multicolumn{2}{c}{43.6} && \multicolumn{2}{c}{45.9} && \multicolumn{2}{c}{40.0} \\
\midrule
\iconClaude Claude Haiku 4.5 & 24.8 & 49.7 && 8.1 & 32.1 && 0.0 & 12.7 && 0.0 & 6.9 && 0.0 & 0.0 \\
\iconClaude Claude Sonnet 4.5 & 27.3 & 53.1 && 9.9 & 37.5 && 2.2 & 12.0 && 0.0 & 7.8 && 0.0 & 0.0 \\
\iconClaude Claude Sonnet 4.6 & 47.3 & 58.2 && 17.1 & 39.0 && 6.7 & 17.2 && 8.8 & 22.5 && 0.0 & 0.0 \\
\iconClaude Claude Opus 4.5 & 37.0 & 55.2 && 15.3 & 45.0 && 3.4 & 13.5 && 0.0 & 12.7 && 0.0 & 0.0 \\
\bottomrule
\end{tabular}%
}
\end{table}

\clearpage
\subsection{Results by Candidate Set Size}
\label{app:imgcount}

Table~\ref{tab:imgcount:results} provides the full breakdown by candidate set size. Consistent with the summary in the main text, performance drops sharply as $k$ increases, and the decline is especially severe under the stricter \texttt{TB-1} \texttt{pass\textasciicircum{}3} criterion. The table also shows that the gap between models and human experts widens noticeably on larger candidate sets.

\begin{table}[t]
\centering
\caption{Topic-level reward-model \texttt{TB-1} results on photography tasks. The final column reports the corresponding domain-level average (\%).}
\label{tab:app:rm_topic_photography}
\resizebox{\textwidth}{!}{%
\scriptsize
\setlength{\tabcolsep}{3pt}
\renewcommand{\arraystretch}{1.02}
\begin{tabular}{@{}c c c c c c c c c c c@{}}
\toprule
Model & Architecture & Food \& Product & Landscape & Macro & Night \& Astro & Portrait & Sports & Street \& City & Wildlife & Avg. \\
\midrule
\rowcolor{gray!10} Random Guess & 24.0 & 23.3 & 20.9 & 16.7 & 16.7 & 33.3 & 16.7 & 26.1 & 30.6 & 23.7 \\
\rowcolor{gray!10} Human Expert & 73.8 & 72.0 & 71.3 & 60.0 & 60.0 & 80.0 & 70.0 & 74.4 & 72.2 & 72.5 \\
\multicolumn{11}{c}{\textit{Reward Models}} \\
\midrule
\textsc{ArtiMuse} & 15.6 & 40.0 & 28.9 & 50.0 & 100.0 & 50.0 & 0.0 & 35.9 & 66.7 & 32.4 \\
\textsc{PEAS-aes} & 37.5 & 60.0 & 42.2 & 75.0 & 0.0 & 0.0 & 0.0 & 46.2 & 33.3 & 41.7 \\
\textsc{PEAS-ava} & 40.6 & 40.0 & 46.7 & 100.0 & 0.0 & 0.0 & 0.0 & 51.3 & 44.4 & 46.0 \\
\textsc{Q-Align} & 46.9 & 40.0 & 44.4 & 75.0 & 50.0 & 100.0 & 0.0 & 46.2 & 44.4 & 46.8 \\
\textsc{Q-Instruct} & 31.2 & 20.0 & 31.1 & 0.0 & 50.0 & 0.0 & 0.0 & 43.6 & 66.7 & 35.2 \\
\textsc{RealQA-ava} & 25.0 & 60.0 & 24.4 & 50.0 & 50.0 & 50.0 & 0.0 & 41.0 & 55.6 & 33.8 \\
\bottomrule
\end{tabular}%
}
\end{table}
\section{Human Study Details}
\label{sec:hs:appendix}

\subsection{Detailed Setup and Metric Definitions}
\label{app:hs:setup}

This subsection provides the full metric definitions summarized in Section~\ref{sec:human_study}.

Eight expert annotators evaluate identical images under both protocols. Under scoring, each annotator independently assigns an absolute score $s_{i,a}\in[0,10]$ to each image. Under ranking, annotators provide a complete within-task ordering when presented with groups of size $n\in\{2,3,4,5\}$. We operationalize ``more scientifically sound'' along three dimensions: fidelity to human comparative preference, reproducibility across annotators, and sharpness of the resulting label signal.

The study comprises 119 \emph{homogeneous-content} tasks drawn directly from the curated benchmark, where all candidates share the same underlying content, and 107 \emph{heterogeneous-content} tasks formed by regrouping 370 images from distinct homogeneous task groups while preserving topic consistency. Each image appears in at most one regrouped task. Examples of both settings are given in Appendix~\ref{app:hs:examples}.

\paragraph{Fidelity.}
We ask whether the ranking induced by absolute scores matches the direct ranking produced by the same annotator. Let $\rho_a^{\text{same}}$ and $\rho_a^{\text{diff}}$ denote annotator $a$'s direct rankings for homogeneous-content and heterogeneous-content tasks, and let $\sigma_a^{\text{score}}$ denote the ranking induced by the absolute scores. We report Kendall's $\tau$ \citep{kendall1938new},
\begin{equation*}
\tau = \frac{C-D}{\binom{n}{2}},
\end{equation*}
where $C$ and $D$ are the numbers of concordant and discordant pairs. We compute $\tau$ for each annotator within each task, average across annotators to obtain a task-level value, and report the mean and standard deviation across tasks. We additionally report a top-1 self-consistency (SC) rate: the fraction of annotator--task pairs for which $\sigma_a^{\text{score}}$ and the direct ranking select the same top image.

\paragraph{Reproducibility.}
We define task-level accuracy $\mathrm{acc}$ as the fraction of tasks on which at least five of the eight annotators agree on the relevant label, and report $\mathrm{acc}_{\text{best}}$, $\mathrm{acc}_{\text{worst}}$, and $\mathrm{acc}_{\text{best\&worst}}$ for both protocols.

\paragraph{Distinguishability.}
Restricting to the 119 homogeneous-content tasks, where entropy reflects concentration of aesthetic preference rather than content mismatch, we measure the entropy of the eight annotators' top-1 selections \citep{shannon1948mathematical},
\begin{equation*}
H = -\sum_i p_i \log_2 p_i,
\end{equation*}
where $p_i$ is the fraction of annotators selecting image $i$ as the top image in a task. Because the maximum entropy depends on task size $n$, we report the unnormalized quantity so that lower values directly indicate stronger concentration of top-choice judgments.

\begin{figure}[t]
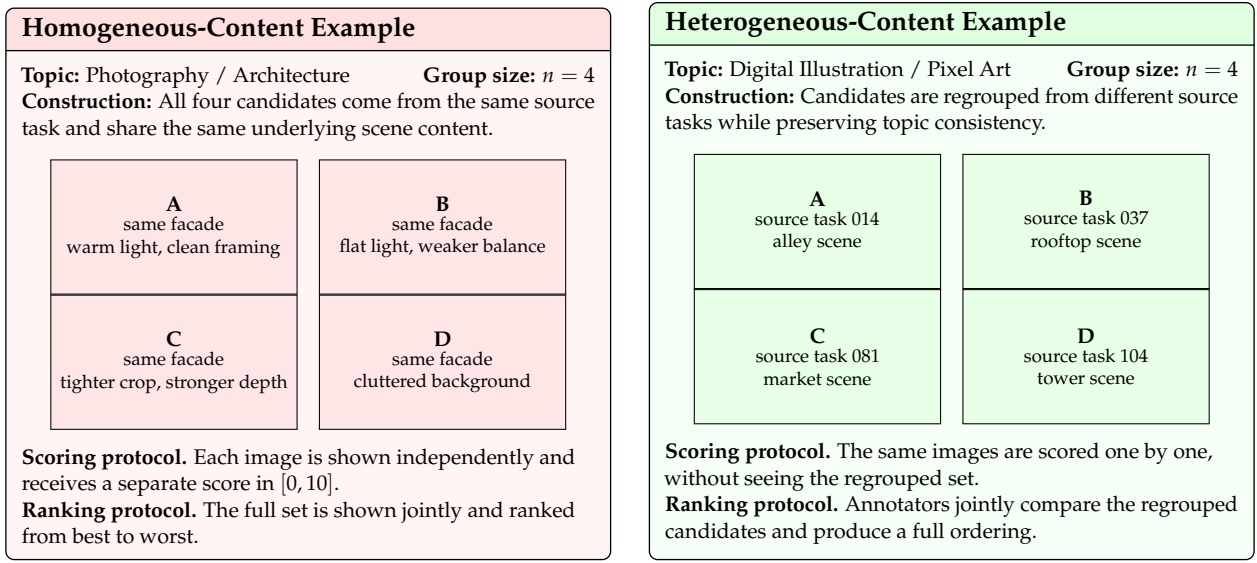

\centering
\setlength{\tabcolsep}{4pt}
\begin{minipage}[t]{0.485\linewidth}
\begin{tcolorbox}[
    enhanced,
    colback=red!4,
    colframe=black,
    colbacktitle=red!12,
    coltitle=black,
    boxrule=0.5pt,
    title=\textbf{Homogeneous-Content Example},
    sharp corners=north,
    rounded corners,
    left=1mm,right=1mm,top=1mm,bottom=1mm
]
\footnotesize
\textbf{Topic:} Photography / Architecture \hfill \textbf{Group size:} $n=4$\\
\textbf{Construction:} All four candidates come from the same source task and share the same underlying scene content.

\medskip
\centering
\begin{tabular}{@{}cc@{}}
\fcolorbox{black}{red!10}{\parbox[c][1.55cm][c]{0.40\linewidth}{\centering \textbf{A}\\[-0.15em]\scriptsize same facade\\ warm light, clean framing}} &
\fcolorbox{black}{red!10}{\parbox[c][1.55cm][c]{0.40\linewidth}{\centering \textbf{B}\\[-0.15em]\scriptsize same facade\\ flat light, weaker balance}} \\
\fcolorbox{black}{red!10}{\parbox[c][1.55cm][c]{0.40\linewidth}{\centering \textbf{C}\\[-0.15em]\scriptsize same facade\\ tighter crop, stronger depth}} &
\fcolorbox{black}{red!10}{\parbox[c][1.55cm][c]{0.40\linewidth}{\centering \textbf{D}\\[-0.15em]\scriptsize same facade\\ cluttered background}} \\
\end{tabular}

\medskip
\raggedright
\textbf{Scoring protocol.} Each image is shown independently and receives a separate score in $[0,10]$.\\
\textbf{Ranking protocol.} The full set is shown jointly and ranked from best to worst.\\
\end{tcolorbox}
\end{minipage}\hfill
\begin{minipage}[t]{0.485\linewidth}
\begin{tcolorbox}[
    enhanced,
    colback=green!4,
    colframe=black,
    colbacktitle=green!12,
    coltitle=black,
    boxrule=0.5pt,
    title=\textbf{Heterogeneous-Content Example},
    sharp corners=north,
    rounded corners,
    left=1mm,right=1mm,top=1mm,bottom=1mm
]
\footnotesize
\textbf{Topic:} Digital Illustration / Pixel Art \hfill \textbf{Group size:} $n=4$\\
\textbf{Construction:} Candidates are regrouped from different source tasks while preserving topic consistency.

\medskip
\centering
\begin{tabular}{@{}cc@{}}
\fcolorbox{black}{green!10}{\parbox[c][1.55cm][c]{0.40\linewidth}{\centering \textbf{A}\\[-0.15em]\scriptsize source task 014\\ alley scene}} &
\fcolorbox{black}{green!10}{\parbox[c][1.55cm][c]{0.40\linewidth}{\centering \textbf{B}\\[-0.15em]\scriptsize source task 037\\ rooftop scene}} \\
\fcolorbox{black}{green!10}{\parbox[c][1.55cm][c]{0.40\linewidth}{\centering \textbf{C}\\[-0.15em]\scriptsize source task 081\\ market scene}} &
\fcolorbox{black}{green!10}{\parbox[c][1.55cm][c]{0.40\linewidth}{\centering \textbf{D}\\[-0.15em]\scriptsize source task 104\\ tower scene}} \\
\end{tabular}

\medskip
\raggedright
\textbf{Scoring protocol.} The same images are scored one by one, without seeing the regrouped set.\\
\textbf{Ranking protocol.} Annotators jointly compare the regrouped candidates and produce a full ordering.\\
\end{tcolorbox}
\end{minipage}
\caption{Schematic examples of the two settings used in the human study. \textbf{Left}: a homogeneous-content task, where all candidates come from the same source task and share the same underlying content. \textbf{Right}: a heterogeneous-content task, where candidates are regrouped from different source tasks. In both settings, the scoring protocol rates images independently, whereas the ranking protocol orders the full set jointly.}
\label{fig:hs:examples}
\end{figure}

\subsection{Extended Interpretation}
\label{subsec:hs:interpretation:app}

\paragraph{Philosophical lens.}
Our results echo the long-standing debate between \textit{Aesthetic Absolutism}, which posits a context-free, intrinsic standard of beauty, and \textit{Aesthetic Relativism}, which holds that judgments arise from comparison and context. Modern philosophical aesthetics has largely moved toward the relational view: people can reliably tell which image they prefer when alternatives are juxtaposed.

\paragraph{Statistical lens.}
Scoring introduces an additional measurement layer. Annotators must first form an internal judgment and then translate it onto a numeric scale, after which pairwise preference is inferred from score differences. This pipeline adds calibration variance, scale-use heterogeneity, and extra measurement noise. Ranking measures the relative decision of interest more directly, which helps explain its higher consistency and sharper signal.

\paragraph{Cognitive lens.}
Comparative judgment is a natural human operation. People can often say which of two images feels better before they can specify by how much. Absolute scoring is cognitively heavier because it requires maintaining an internal reference scale, calibrating intervals, and converting a qualitative impression into a number. That added abstraction likely makes scoring less stable than direct comparison.

\subsection{Task Examples}
\label{app:hs:examples}

Figure~\ref{fig:hs:examples} provides schematic examples of the two task settings used in the human study.

\subsection{Annotation Protocol}
\label{app:hs:protocol}

\paragraph{Scoring task.}
Under the scoring protocol, each annotator independently assigned an absolute score $s_{i,a}\in[0,10]$ to each of the 400 images in the base image pool. Images were presented individually, without access to the other members of the same task group. The resulting scores were later reused to derive score-derived rankings for both the 119 homogeneous-content tasks and the 107 heterogeneous-content regroupings. Domain-specific single-image scoring rubrics are given below.

\begin{rubricbox}[title=Fine Art Single-Image Scoring Rubric]
For a single fine art painting, evaluate the degree of artistic completeness and maturity of aesthetic expression based solely on the image itself, and assign an overall score from 0 to 10, accurate to one decimal place.

\textbf{Reference Scoring Criteria}
\begin{enumerate}[leftmargin=1.5em]
    \item \textbf{Composition and Visual Order}
    \begin{itemize}[leftmargin=1.5em]
        \item Is the primary subject clearly defined and visually grounded?
        \item Does the composition feel stable or intentionally dynamic?
        \item Is the spatial arrangement natural, without unnecessary congestion or emptiness?
    \end{itemize}
    \item \textbf{Color and Tonal Relationships}
    \begin{itemize}[leftmargin=1.5em]
        \item Are color relationships harmonious and layered?
        \item Is there effective control of warmth/coolness and light/dark contrast?
        \item Does the work avoid muddiness, dullness, or scattered color impressions?
    \end{itemize}
    \item \textbf{Technical Control}
    \begin{itemize}[leftmargin=1.5em]
        \item Are the brushstrokes confident, fluid, and rhythmic?
        \item Are there signs of hesitation, redundancy, or ineffective marks?
        \item Does technique serve the image rather than draw attention to itself?
    \end{itemize}
    \item \textbf{Degree of Completion}
    \begin{itemize}[leftmargin=1.5em]
        \item Are there areas that appear unfinished or perfunctory?
        \item Are edges and transitions handled naturally?
        \item Does the work feel resolved, at a point where the artist could reasonably stop?
    \end{itemize}
    \item \textbf{Artistic Expression and Overall Character}
    \begin{itemize}[leftmargin=1.5em]
        \item Is the mood or atmosphere focused and coherent?
        \item Does the work convey a unified overall character or presence?
        \item Does it demonstrate strong artistic persuasiveness?
    \end{itemize}
    \item \textbf{Overall Score}
    \begin{itemize}[leftmargin=1.5em]
        \item Based on all criteria above, assign an overall score from 0 to 10, accurate to one decimal place.
    \end{itemize}
\end{enumerate}

\textbf{Should Not Be Considered}
\begin{itemize}[leftmargin=1.5em]
    \item Personal preference for subject matter.
    \item Artist background, intent, or any contextual information beyond what is visible in the image itself.
\end{itemize}
\end{rubricbox}

\begin{rubricbox}[title=Photography Single-Image Scoring Rubric]
For a single photograph, evaluate its overall quality in terms of photographic language and visual expression, based solely on the image itself, and assign an overall score from 0 to 10, accurate to one decimal place.

\textbf{Reference Scoring Criteria}
\begin{enumerate}[leftmargin=1.5em]
    \item \textbf{Composition and Image Structure}
    \begin{itemize}[leftmargin=1.5em]
        \item Is the subject clearly emphasized?
        \item Does the frame feel balanced or intentionally dynamic?
        \item Is the background clean and supportive of the subject?
    \end{itemize}
    \item \textbf{Quality of Light}
    \begin{itemize}[leftmargin=1.5em]
        \item Is the light natural or expressive?
        \item Are highlights and shadows well controlled?
        \item Does the image avoid crushed blacks or blown highlights?
    \end{itemize}
    \item \textbf{Sharpness and Focus}
    \begin{itemize}[leftmargin=1.5em]
        \item Is focus accurate where it matters most?
        \item Is the relationship between sharpness and blur appropriate?
        \item Are there any noticeable technical flaws?
    \end{itemize}
    \item \textbf{Color or Tonal Rendering}
    \begin{itemize}[leftmargin=1.5em]
        \item Is white balance or color temperature accurate or clearly intentional?
        \item Are colors or tonal values consistent and unified?
        \item Does the image avoid unwanted color casts or muddy tones?
    \end{itemize}
    \item \textbf{Moment and Emotional Expression}
    \begin{itemize}[leftmargin=1.5em]
        \item Does the image capture a meaningful moment?
        \item Is the emotional or narrative content clear and compelling?
        \item Does the image convey a strong sense of photographic presence?
    \end{itemize}
    \item \textbf{Overall Score}
    \begin{itemize}[leftmargin=1.5em]
        \item Based on all criteria above, assign an overall score from 0 to 10, accurate to one decimal place.
    \end{itemize}
\end{enumerate}

\textbf{Should Not Be Considered}
\begin{itemize}[leftmargin=1.5em]
    \item The personal appeal or attractiveness of the subject.
    \item Assumptions about shooting difficulty.
    \item Assumptions about post-processing effort.
\end{itemize}
\end{rubricbox}

\begin{rubricbox}[title=Digital Illustration Single-Image Scoring Rubric]
For a single digital illustration, evaluate its overall quality in terms of visual design and professional illustration standards, based solely on the image itself, and assign an overall score from 0 to 10, accurate to one decimal place.

\textbf{Reference Scoring Criteria}
\begin{enumerate}[leftmargin=1.5em]
    \item \textbf{Visual Focus and Information Clarity}
    \begin{itemize}[leftmargin=1.5em]
        \item Is the visual focal point clear and well defined?
        \item Does the visual flow guide the viewer smoothly?
        \item Is there minimal visual noise or competition for attention?
    \end{itemize}
    \item \textbf{Form and Proportion}
    \begin{itemize}[leftmargin=1.5em]
        \item Are the structures of figures or objects accurate?
        \item Are poses and silhouettes aesthetically refined?
        \item Does the image avoid stiffness, distortion, or unnatural forms?
    \end{itemize}
    \item \textbf{Color Design}
    \begin{itemize}[leftmargin=1.5em]
        \item Is the color palette unified and thematically coherent?
        \item Are value and lighting hierarchies clear?
        \item Is contrast and visual rhythm used effectively?
    \end{itemize}
    \item \textbf{Detail Control and Finish}
    \begin{itemize}[leftmargin=1.5em]
        \item Are details handled appropriately, neither excessive nor careless?
        \item Are materials and lighting consistent throughout the image?
        \item Are edges, masks, and gradients clean and well controlled?
    \end{itemize}
    \item \textbf{Stylistic Consistency and Purpose Fit}
    \begin{itemize}[leftmargin=1.5em]
        \item Does the illustration align with its intended stylistic direction?
        \item Is quality consistent within a unified style?
        \item Is the image suitable for its potential use case?
    \end{itemize}
    \item \textbf{Overall Score}
    \begin{itemize}[leftmargin=1.5em]
        \item Based on all criteria above, assign an overall score from 0 to 10, accurate to one decimal place.
    \end{itemize}
\end{enumerate}

\textbf{Should Not Be Considered}
\begin{itemize}[leftmargin=1.5em]
    \item Personal preference for illustration style.
    \item Software or tools used.
    \item Hypothetical or imagined client requirements.
\end{itemize}
\end{rubricbox}

When converting absolute scores into induced rankings, no exact within-task score ties were observed in the collected annotations, so no tie-breaking rule was required.

\paragraph{Ranking task.}
Under the ranking protocol, each annotator provided a complete best-to-worst ordering of all $n$ images presented simultaneously within a group. Annotators were instructed to rank the images by overall aesthetic quality using the following domain-specific comparative rubrics.

\begin{rubricbox}[title=Fine Art Ranking Rubric]
For a set of fine art paintings that share the same style, subject matter, and medium, and differ only subtly in execution, rank the images from best to worst according to artistic completeness and maturity of aesthetic expression.

\textbf{Reference Comparison Criteria}
\begin{enumerate}[leftmargin=1.5em]
    \item \textbf{Composition and Visual Order}
    \begin{itemize}[leftmargin=1.5em]
        \item Is the primary subject more clearly defined?
        \item Does the composition feel more stable or intentionally dynamic?
        \item Is the spatial arrangement natural, without unnecessary congestion or emptiness?
    \end{itemize}
    \item \textbf{Color and Tonal Relationships}
    \begin{itemize}[leftmargin=1.5em]
        \item Are the color relationships more harmonious or layered?
        \item Is there stronger control over warmth/coolness and light/dark contrast?
        \item Does the work avoid muddiness, dullness, or scattered color impressions?
    \end{itemize}
    \item \textbf{Technical Control}
    \begin{itemize}[leftmargin=1.5em]
        \item Are the brushstrokes more confident, fluid, and rhythmic?
        \item Are there signs of hesitation, redundancy, or ineffective marks?
        \item Does the technique serve the image rather than draw attention to itself?
    \end{itemize}
    \item \textbf{Degree of Completion}
    \begin{itemize}[leftmargin=1.5em]
        \item Are there any areas that appear unfinished or perfunctory?
        \item Are edges and transitions handled naturally?
        \item Does the work feel resolved, at a point where the artist could reasonably stop?
    \end{itemize}
    \item \textbf{Artistic Expression and Overall Character}
    \begin{itemize}[leftmargin=1.5em]
        \item Is the mood or atmosphere more focused and coherent?
        \item Does the work convey a stronger overall character or presence?
        \item Does it have greater artistic persuasiveness?
    \end{itemize}
    \item \textbf{Final Selection}
    \begin{itemize}[leftmargin=1.5em]
        \item Considering the set as a whole, determine the relative ordering of the images from strongest to weakest.
    \end{itemize}
\end{enumerate}

\textbf{Should Not Be Considered}
\begin{itemize}[leftmargin=1.5em]
    \item Personal preference for subject matter.
    \item Any contextual or background information beyond what is visible in the image itself.
\end{itemize}
\end{rubricbox}

\begin{rubricbox}[title=Photography Ranking Rubric]
For a set of photographs taken of similar subjects, in similar scenes, and under closely matched conditions, rank the images from best to worst in terms of photographic language and visual expression.

\textbf{Reference Comparison Criteria}
\begin{enumerate}[leftmargin=1.5em]
    \item \textbf{Composition and Image Structure}
    \begin{itemize}[leftmargin=1.5em]
        \item Is the subject more clearly emphasized?
        \item Is the frame more balanced or intentionally dynamic?
        \item Is the background cleaner and more supportive of the subject?
    \end{itemize}
    \item \textbf{Quality of Light}
    \begin{itemize}[leftmargin=1.5em]
        \item Is the light more natural or more expressive?
        \item Are highlights and shadows better controlled?
        \item Does the image avoid crushed blacks or blown highlights?
    \end{itemize}
    \item \textbf{Sharpness and Focus}
    \begin{itemize}[leftmargin=1.5em]
        \item Is focus more accurate where it matters?
        \item Is the relationship between sharpness and blur more appropriate?
        \item Are there any technical errors that detract from the image?
    \end{itemize}
    \item \textbf{Color or Tonal Rendering}
    \begin{itemize}[leftmargin=1.5em]
        \item Is white balance or color temperature more accurate or more intentional?
        \item Are colors or tonal values more consistent and unified?
        \item Does the image avoid unwanted color casts or muddy tones?
    \end{itemize}
    \item \textbf{Moment and Emotional Expression}
    \begin{itemize}[leftmargin=1.5em]
        \item Does the image capture a stronger or more meaningful moment?
        \item Is the emotional or narrative content more compelling?
        \item Does it carry a stronger sense of photographic presence?
    \end{itemize}
    \item \textbf{Final Selection}
    \begin{itemize}[leftmargin=1.5em]
        \item Considering the set as a whole, determine the relative ordering of the photographs from strongest to weakest.
    \end{itemize}
\end{enumerate}

\textbf{Should Not Be Considered}
\begin{itemize}[leftmargin=1.5em]
    \item The personal appeal or attractiveness of the subject.
    \item Assumptions about shooting difficulty.
    \item Assumptions about post-processing effort.
\end{itemize}
\end{rubricbox}

\begin{rubricbox}[title=Digital Illustration Ranking Rubric]
For a set of digital illustrations with similar purpose, style, and level of completion, rank the images from best to worst according to visual design and professional illustration quality.

\textbf{Reference Comparison Criteria}
\begin{enumerate}[leftmargin=1.5em]
    \item \textbf{Visual Focus and Information Clarity}
    \begin{itemize}[leftmargin=1.5em]
        \item Is the visual focal point clearer?
        \item Does the visual flow guide the viewer more effectively?
        \item Is there less visual noise or competition for attention?
    \end{itemize}
    \item \textbf{Form and Proportion}
    \begin{itemize}[leftmargin=1.5em]
        \item Are the structures of figures or objects more accurate?
        \item Are poses and silhouettes more aesthetically refined?
        \item Does the image avoid stiffness, distortion, or unnatural forms?
    \end{itemize}
    \item \textbf{Color Design}
    \begin{itemize}[leftmargin=1.5em]
        \item Is the color palette more unified and thematically coherent?
        \item Are value and lighting hierarchies clearer?
        \item Is there effective use of contrast and visual rhythm?
    \end{itemize}
    \item \textbf{Detail Control and Finish}
    \begin{itemize}[leftmargin=1.5em]
        \item Are details handled appropriately, neither excessive nor careless?
        \item Are materials and lighting consistent throughout?
        \item Are edges, masks, and gradients clean and well controlled?
    \end{itemize}
    \item \textbf{Stylistic Consistency and Purpose Fit}
    \begin{itemize}[leftmargin=1.5em]
        \item Does the illustration better align with its intended stylistic direction?
        \item Is quality more consistent within a unified style?
        \item Is the image more suitable for its potential use case?
    \end{itemize}
    \item \textbf{Final Selection}
    \begin{itemize}[leftmargin=1.5em]
        \item Considering the set as a whole, determine the relative ordering of the illustrations from strongest to weakest.
    \end{itemize}
\end{enumerate}

\textbf{Should Not Be Considered}
\begin{itemize}[leftmargin=1.5em]
    \item Personal preference for illustration style.
    \item Software or tools used.
    \item Hypothetical or imagined client requirements.
\end{itemize}
\end{rubricbox}

\subsection{Notation}
\label{app:hs:notation}

\begin{center}
\small
\renewcommand{\arraystretch}{1.3}
\begin{tabular}{@{}>{$}l<{$} >{\raggedright\arraybackslash}p{0.78\linewidth}@{}}
\toprule
\textnormal{\textbf{Symbol}} & \textbf{Meaning} \\
\midrule
A = 8 & number of annotators \\
n & number of images in a task, $n \in \{2, 3, 4, 5\}$ \\
s_{i,a} & absolute score assigned to image $i$ by annotator $a$, $s_{i,a} \in [0, 10]$ \\
\sigma^{\text{score}}_a & score-derived ranking for annotator $a$, obtained by sorting images in descending order of $s_{i,a}$ \\
\rho^{\text{same}}_a & direct ranking provided by annotator $a$ for a homogeneous-content task \\
\rho^{\text{diff}}_a & direct ranking provided by annotator $a$ for a heterogeneous-content task \\
y^{\text{best}}_a,\, y^{\text{worst}}_a & annotator $a$'s best and worst labels in a task \\
\mathrm{SC} & top-1 self-consistency rate between score-derived and direct rankings \\
\mathrm{acc} & fraction of tasks on which annotators reach majority agreement on the relevant label \\
\bottomrule
\end{tabular}
\end{center}

\paragraph{Task-level accuracy.}
For the majority-agreement comparison, task-level accuracy is defined as follows:
\begin{itemize}
\item $\mathrm{acc}_{\text{best}}$: fraction of tasks on which at least five of the eight annotators select the same best label.
\item $\mathrm{acc}_{\text{worst}}$: fraction of tasks on which at least five of the eight annotators select the same worst label.
\item $\mathrm{acc}_{\text{best\&worst}}$: fraction of tasks on which both majority conditions are satisfied simultaneously.
\end{itemize}

\subsection{Statistical Tests}
\label{app:hs:tests}

We report two-sided paired tests over tasks to assess whether the ranking and scoring protocols differ systematically. Unless otherwise noted, $p$-values are uncorrected for multiple comparisons.

\paragraph{McNemar's test for majority-agreement comparisons.}
For each task and each metric $\mathrm{acc}_{\text{best}}$, $\mathrm{acc}_{\text{worst}}$, and $\mathrm{acc}_{\text{best\&worst}}$, we construct paired binary outcomes indicating whether the protocol achieves a five-of-eight majority hit. We then apply an exact two-sided McNemar test \citep{mcnemar1947note,fagerland2013mcnemar} to the discordant pairs.
\begin{itemize}
\item Homogeneous-content ($N=119$ tasks): $p=1.0\times 10^{-11}$ for $\mathrm{acc}_{\text{best}}$, $p=2.4\times 10^{-9}$ for $\mathrm{acc}_{\text{worst}}$, and $p=1.2\times 10^{-14}$ for $\mathrm{acc}_{\text{best\&worst}}$.
\item Heterogeneous-content ($N=107$ tasks): $p=1.2\times 10^{-7}$ for $\mathrm{acc}_{\text{best}}$, $p=2.4\times 10^{-5}$ for $\mathrm{acc}_{\text{worst}}$, and $p=1.0\times 10^{-9}$ for $\mathrm{acc}_{\text{best\&worst}}$.
\end{itemize}

\paragraph{Wilcoxon signed-rank test for entropy comparison.}
We compare the paired per-task entropies $H_{\text{rank}}$ and $H_{\text{score}}$ over the 119 homogeneous-content tasks using a two-sided Wilcoxon signed-rank test \citep{wilcoxon1945individual}, which yields $p=1.7\times 10^{-15}$.

\paragraph{Multiple comparisons.}
After Bonferroni correction \citep{bonferroni1936teoria} over the three metrics in each setting, all six tests remain significant at $p<0.001$.

\subsection{Task Lists}
\label{app:hs:diff_tasks}

\subsubsection*{Homogeneous-Content Tasks (119 Tasks)}

Each homogeneous-content task groups images that share the same underlying content but differ in aesthetic treatment. The 119 tasks span four topics across three domains.

\begingroup
\footnotesize
\setlength{\LTpre}{0pt}
\setlength{\LTpost}{0pt}
\setlength{\tabcolsep}{4pt}
\renewcommand{\arraystretch}{1.02}
\begin{longtable}{@{}>{\ttfamily\raggedright\arraybackslash}p{0.78\linewidth} c@{}}
\caption{Homogeneous-content tasks (119 tasks).}\label{tab:hs:same_task_list}\\
\toprule
Task ID & $n$ \\
\midrule
\endfirsthead
\multicolumn{2}{@{}l}{\tablename~\thetable\ (continued)}\\
\toprule
Task ID & $n$ \\
\midrule
\endhead
\midrule
\multicolumn{2}{r@{}}{\footnotesize Continued on next page} \\
\endfoot
\bottomrule
\endlastfoot
\texttt{fine-art\_landscape-color\_61} & 4 \\
\texttt{fine-art\_landscape-color\_64} & 2 \\
\texttt{fine-art\_landscape-color\_65} & 4 \\
\texttt{fine-art\_landscape-color\_66} & 4 \\
\texttt{fine-art\_landscape-color\_68} & 5 \\
\texttt{fine-art\_landscape-color\_69} & 5 \\
\texttt{fine-art\_landscape-color\_70} & 5 \\
\texttt{fine-art\_landscape-color\_71} & 5 \\
\end{longtable}
\endgroup

\subsubsection*{Heterogeneous-Content Tasks (107 Tasks)}

Each heterogeneous-content task is formed by regrouping images drawn from distinct homogeneous-content tasks, typically within the same topic. Of the original 119 regrouped tasks, 12 are excluded because the score-based Kendall's $W$ \citep{kendall1939problem} is at least 0.825, indicating that the regrouped task already exhibits near-saturated consensus under the scoring protocol. The remaining 107 tasks are retained for analysis. The Source Task IDs column lists the originating homogeneous-content task for each image in the group, in label order A, B, C, \ldots.

\begingroup
\footnotesize
\setlength{\LTpre}{0pt}
\setlength{\LTpost}{0pt}
\setlength{\tabcolsep}{4pt}
\renewcommand{\arraystretch}{1.02}
\begin{longtable}{@{}c c >{\ttfamily\raggedright\arraybackslash}p{0.69\linewidth}@{}}
\caption{Heterogeneous-content task construction (107 tasks, after saturation filter).}\label{tab:hs:diff_task_list}\\
\toprule
ID & $n$ & Source Task IDs \\
\midrule
\endfirsthead
\multicolumn{3}{@{}l}{\tablename~\thetable\ (continued)}\\
\toprule
ID & $n$ & Source Task IDs \\
\midrule
\endhead
\midrule
\multicolumn{3}{r@{}}{\footnotesize Continued on next page} \\
\endfoot
\bottomrule
\endlastfoot
1 & 3 & \texttt{illustration\_pixel-art\_19}, \texttt{illustration\_pixel-art\_23}, \texttt{illustration\_pixel-art\_17} \\
2 & 3 & \texttt{photograph\_landscape\_27}, \texttt{photograph\_landscape\_2}, \texttt{photograph\_landscape\_3} \\
3 & 3 & \texttt{illustration\_pixel-art\_35}, \texttt{illustration\_pixel-art\_4}, \texttt{illustration\_pixel-art\_12} \\
4 & 4 & \texttt{fine-art\_landscape-color\_81}, \texttt{fine-art\_landscape-color\_82}, \texttt{fine-art\_landscape-color\_70}, \texttt{fine-art\_landscape-color\_69} \\
5 & 4 & \texttt{illustration\_pixel-art\_31}, \texttt{illustration\_pixel-art\_40}, \texttt{illustration\_pixel-art\_8}, \texttt{illustration\_pixel-art\_3} \\
6 & 2 & \texttt{photograph\_landscape\_86}, \texttt{photograph\_landscape\_81} \\
7 & 2 & \texttt{illustration\_pixel-art\_22}, \texttt{illustration\_pixel-art\_8} \\
8 & 3 & \texttt{illustration\_pixel-art\_20}, \texttt{illustration\_pixel-art\_36}, \texttt{illustration\_pixel-art\_19} \\
\end{longtable}
\endgroup

\section{Fine-Tuning Details}
\label{app:finetune}

\subsection{Hyperparameters}

Table~\ref{tab:app:finetune} lists the hyperparameters used to fine-tune \model{}.

\begin{table}[H]
\centering
\caption{Hyperparameters used for fine-tuning \model{}.}
\label{tab:app:finetune}
\scriptsize
\begin{tabular}{@{}cc@{}}
\toprule
Parameter & Value \\
\midrule
Base Model & \iconQwenThreeFive Qwen3.5-35B-A3B \\
Fine-tuning Method & LoRA \\
LoRA Rank & 64 \\
LoRA Alpha & 128 \\
LoRA Target & All layers \\
Vision Tower & Frozen \\
MLP Projector & Trainable \\
Learning Rate & $1 \times 10^{-4}$ \\
LR Scheduler & Cosine \\
Warmup Steps & 10 \\
Epochs & 3 \\
Batch Size & 4 per device \\
Precision & BF16 \\
Max Sequence Length & 65{,}536 \\
\bottomrule
\end{tabular}
\end{table}

\subsection{Detailed Results}
\label{app:finetune_results}

As in the main-text fine-tuning table, we report topic-level detailed results under all four metrics: \texttt{Top-1} \texttt{ap@1}, \texttt{Top-1} \texttt{pass\textasciicircum{}3}, \texttt{TB-1} \texttt{ap@1}, and \texttt{TB-1} \texttt{pass\textasciicircum{}3}. The full breakdown shows that the gains from fine-tuning are uneven across topics: \model{} improves most clearly on several fine-art topics, remains competitive with larger open-weight baselines on illustration, and does not uniformly dominate its base model in photography.

\medskip
\noindent\textbf{\texttt{Top-1} \texttt{ap@1}.}\par
Tables~\ref{tab:app:finetune_topic_top1_ap1_fine-art}, \ref{tab:app:finetune_topic_top1_ap1_illustration}, and \ref{tab:app:finetune_topic_top1_ap1_photography} report topic-level \texttt{Top-1} \texttt{ap@1} results for fine art, illustration, and photography, respectively.

\begin{table}[t]
\centering
\caption{Topic-level fine-tuning \texttt{Top-1} \texttt{ap@1} results on fine-art tasks. The final column reports the corresponding domain-level average (\%).}
\label{tab:app:finetune_topic_top1_ap1_fine-art}
\resizebox{\textwidth}{!}{%
\scriptsize
\setlength{\tabcolsep}{3pt}
\renewcommand{\arraystretch}{1.02}
\begin{tabular}{@{}c c c c c c c c c c c@{}}
\toprule
Model & Calligraphy & Chinese Painting & Ink \& Wash & Landscape Color & Portrait Color & Portrait Sketch & Quick Sketch & Still Life Color & Still Life Sketch & Avg. \\
\midrule
\multicolumn{11}{c}{\textit{Open-Weight Models}} \\
\midrule
\iconGLM\textsc{GLM-4.6V} & 21.2 & 63.0 & 33.3 & 26.8 & 66.7 & 52.5 & 54.2 & 58.3 & 52.4 & 42.7 \\
\iconKimi\textsc{Kimi-K2.5} & 30.3 & 85.2 & 50.0 & 27.4 & 100.0 & 46.5 & 62.5 & 50.0 & 52.4 & 43.5 \\
\iconQwenVL\textsc{Qwen3-VL-235B-A22B} & 15.2 & 85.2 & 50.0 & 26.8 & 66.7 & 48.5 & 66.7 & 47.2 & 57.1 & 42.4 \\
\iconQwenThreeFive\textsc{Qwen3.5-397B-A17B} & 24.2 & 96.3 & 66.7 & 33.9 & 100.0 & 37.4 & 58.3 & 43.1 & 45.2 & 42.7 \\
\midrule
\multicolumn{11}{c}{\textit{Ours}} \\
\midrule
\iconQwenThreeFive\textsc{Qwen3.5-35B-A3B} (base) & 21.2 & 85.2 & 75.0 & 31.0 & 83.3 & 41.4 & 75.0 & 37.5 & 47.6 & 41.8 \\
\iconKallisti\model{} & 24.2 & 81.5 & 58.3 & 36.3 & 66.7 & 52.5 & 75.0 & 47.2 & 50.0 & 47.0 \\
\bottomrule
\end{tabular}%
}
\end{table}

\begin{table}[t]
\centering
\caption{Topic-level fine-tuning \texttt{Top-1} \texttt{ap@1} results on illustration tasks. The final column reports the corresponding domain-level average (\%).}
\label{tab:app:finetune_topic_top1_ap1_illustration}
\resizebox{\textwidth}{!}{%
\scriptsize
\setlength{\tabcolsep}{3pt}
\renewcommand{\arraystretch}{1.02}
\begin{tabular}{@{}c c c c c c c c@{}}
\toprule
Model & Anime \& Manga & Comic & Concept Art & Digital \& AI Art & Pixel Art & Stylized 3D & Avg. \\
\midrule
\multicolumn{8}{c}{\textit{Open-Weight Models}} \\
\midrule
\iconGLM\textsc{GLM-4.6V} & 51.3 & 53.3 & 31.8 & 37.7 & 60.3 & 27.3 & 43.3 \\
\iconKimi\textsc{Kimi-K2.5} & 71.8 & 56.7 & 36.4 & 36.2 & 39.7 & 30.3 & 43.0 \\
\iconQwenVL\textsc{Qwen3-VL-235B-A22B} & 69.2 & 33.3 & 30.3 & 30.4 & 39.7 & 36.4 & 38.3 \\
\iconQwenThreeFive\textsc{Qwen3.5-397B-A17B} & 71.8 & 53.3 & 33.3 & 27.5 & 50.8 & 48.5 & 44.3 \\
\midrule
\multicolumn{8}{c}{\textit{Ours}} \\
\midrule
\iconQwenThreeFive\textsc{Qwen3.5-35B-A3B} (base) & 53.8 & 46.7 & 30.3 & 44.9 & 41.3 & 30.3 & 40.7 \\
\iconKallisti\model{} & 66.7 & 50.0 & 28.8 & 36.2 & 44.4 & 36.4 & 41.7 \\
\bottomrule
\end{tabular}%
}
\end{table}

\begin{table}[t]
\centering
\caption{Topic-level fine-tuning \texttt{Top-1} \texttt{ap@1} results on photography tasks. The final column reports the corresponding domain-level average (\%).}
\label{tab:app:finetune_topic_top1_ap1_photography}
\resizebox{\textwidth}{!}{%
\scriptsize
\setlength{\tabcolsep}{3pt}
\renewcommand{\arraystretch}{1.02}
\begin{tabular}{@{}c c c c c c c c c c c@{}}
\toprule
Model & Architecture & Food \& Product & Landscape & Macro & Night \& Astro & Portrait & Sports & Street \& City & Wildlife & Avg. \\
\midrule
\multicolumn{11}{c}{\textit{Open-Weight Models}} \\
\midrule
\iconGLM\textsc{GLM-4.6V} & 35.4 & 53.3 & 39.3 & 58.3 & 16.7 & 0.0 & 100.0 & 48.7 & 63.0 & 43.2 \\
\iconKimi\textsc{Kimi-K2.5} & 49.0 & 80.0 & 63.7 & 75.0 & 50.0 & 50.0 & 100.0 & 61.5 & 70.4 & 60.9 \\
\iconQwenVL\textsc{Qwen3-VL-235B-A22B} & 37.5 & 66.7 & 54.1 & 58.3 & 50.0 & 50.0 & 66.7 & 57.3 & 66.7 & 52.5 \\
\iconQwenThreeFive\textsc{Qwen3.5-397B-A17B} & 44.8 & 66.7 & 64.4 & 83.3 & 50.0 & 50.0 & 100.0 & 72.6 & 66.7 & 62.8 \\
\midrule
\multicolumn{11}{c}{\textit{Ours}} \\
\midrule
\iconQwenThreeFive\textsc{Qwen3.5-35B-A3B} (base) & 51.0 & 46.7 & 57.0 & 83.3 & 50.0 & 33.3 & 100.0 & 65.0 & 70.4 & 59.0 \\
\iconKallisti\model{} & 50.0 & 73.3 & 67.4 & 75.0 & 50.0 & 33.3 & 100.0 & 66.7 & 70.4 & 63.3 \\
\bottomrule
\end{tabular}%
}
\end{table}

\medskip
\noindent\textbf{\texttt{Top-1} \texttt{pass\textasciicircum{}3}.}\par
Tables~\ref{tab:app:finetune_topic_top1_p3_fine-art}, \ref{tab:app:finetune_topic_top1_p3_illustration}, and \ref{tab:app:finetune_topic_top1_p3_photography} report topic-level \texttt{Top-1} \texttt{pass\textasciicircum{}3} results for fine art, illustration, and photography, respectively.

\begin{table}[t]
\centering
\caption{Topic-level fine-tuning \texttt{Top-1} \texttt{pass\textasciicircum{}3} results on fine-art tasks. The final column reports the corresponding domain-level average (\%).}
\label{tab:app:finetune_topic_top1_p3_fine-art}
\resizebox{\textwidth}{!}{%
\scriptsize
\setlength{\tabcolsep}{3pt}
\renewcommand{\arraystretch}{1.02}
\begin{tabular}{@{}c c c c c c c c c c c@{}}
\toprule
Model & Calligraphy & Chinese Painting & Ink \& Wash & Landscape Color & Portrait Color & Portrait Sketch & Quick Sketch & Still Life Color & Still Life Sketch & Avg. \\
\midrule
\multicolumn{11}{c}{\textit{Open-Weight Models}} \\
\midrule
\iconGLM\textsc{GLM-4.6V} & 9.1 & 55.6 & 0.0 & 5.4 & 50.0 & 21.2 & 25.0 & 33.3 & 14.3 & 18.0 \\
\iconKimi\textsc{Kimi-K2.5} & 9.1 & 77.8 & 25.0 & 5.4 & 100.0 & 18.2 & 50.0 & 16.7 & 28.6 & 19.9 \\
\iconQwenVL\textsc{Qwen3-VL-235B-A22B} & 9.1 & 77.8 & 50.0 & 8.9 & 50.0 & 24.2 & 50.0 & 20.8 & 21.4 & 22.4 \\
\iconQwenThreeFive\textsc{Qwen3.5-397B-A17B} & 18.2 & 88.9 & 50.0 & 16.1 & 100.0 & 6.1 & 37.5 & 16.7 & 14.3 & 21.1 \\
\midrule
\multicolumn{11}{c}{\textit{Ours}} \\
\midrule
\iconQwenThreeFive\textsc{Qwen3.5-35B-A3B} (base) & 0.0 & 66.7 & 75.0 & 14.3 & 50.0 & 15.2 & 50.0 & 16.7 & 28.6 & 21.7 \\
\iconKallisti\model{} & 0.0 & 77.8 & 50.0 & 19.6 & 50.0 & 27.3 & 62.5 & 33.3 & 14.3 & 28.0 \\
\bottomrule
\end{tabular}%
}
\end{table}

\begin{table}[t]
\centering
\caption{Topic-level fine-tuning \texttt{Top-1} \texttt{pass\textasciicircum{}3} results on illustration tasks. The final column reports the corresponding domain-level average (\%).}
\label{tab:app:finetune_topic_top1_p3_illustration}
\resizebox{\textwidth}{!}{%
\scriptsize
\setlength{\tabcolsep}{3pt}
\renewcommand{\arraystretch}{1.02}
\begin{tabular}{@{}c c c c c c c c@{}}
\toprule
Model & Anime \& Manga & Comic & Concept Art & Digital \& AI Art & Pixel Art & Stylized 3D & Avg. \\
\midrule
\multicolumn{8}{c}{\textit{Open-Weight Models}} \\
\midrule
\iconGLM\textsc{GLM-4.6V} & 23.1 & 20.0 & 9.1 & 17.4 & 38.1 & 0.0 & 19.0 \\
\iconKimi\textsc{Kimi-K2.5} & 53.8 & 30.0 & 13.6 & 21.7 & 9.5 & 9.1 & 21.0 \\
\iconQwenVL\textsc{Qwen3-VL-235B-A22B} & 46.2 & 0.0 & 4.5 & 4.3 & 9.5 & 18.2 & 12.0 \\
\iconQwenThreeFive\textsc{Qwen3.5-397B-A17B} & 53.8 & 10.0 & 13.6 & 4.3 & 33.3 & 45.5 & 24.0 \\
\midrule
\multicolumn{8}{c}{\textit{Ours}} \\
\midrule
\iconQwenThreeFive\textsc{Qwen3.5-35B-A3B} (base) & 23.1 & 20.0 & 9.1 & 34.8 & 14.3 & 18.2 & 20.0 \\
\iconKallisti\model{} & 53.8 & 10.0 & 9.1 & 21.7 & 19.0 & 27.3 & 22.0 \\
\bottomrule
\end{tabular}%
}
\end{table}

\begin{table}[t]
\centering
\caption{Topic-level fine-tuning \texttt{Top-1} \texttt{pass\textasciicircum{}3} results on photography tasks. The final column reports the corresponding domain-level average (\%).}
\label{tab:app:finetune_topic_top1_p3_photography}
\resizebox{\textwidth}{!}{%
\scriptsize
\setlength{\tabcolsep}{3pt}
\renewcommand{\arraystretch}{1.02}
\begin{tabular}{@{}c c c c c c c c c c c@{}}
\toprule
Model & Architecture & Food \& Product & Landscape & Macro & Night \& Astro & Portrait & Sports & Street \& City & Wildlife & Avg. \\
\midrule
\multicolumn{11}{c}{\textit{Open-Weight Models}} \\
\midrule
\iconGLM\textsc{GLM-4.6V} & 6.2 & 20.0 & 15.6 & 50.0 & 0.0 & 0.0 & 100.0 & 15.4 & 33.3 & 15.8 \\
\iconKimi\textsc{Kimi-K2.5} & 18.8 & 60.0 & 44.4 & 75.0 & 50.0 & 50.0 & 100.0 & 35.9 & 55.6 & 38.8 \\
\iconQwenVL\textsc{Qwen3-VL-235B-A22B} & 6.2 & 60.0 & 26.7 & 50.0 & 50.0 & 0.0 & 0.0 & 28.2 & 33.3 & 24.5 \\
\iconQwenThreeFive\textsc{Qwen3.5-397B-A17B} & 31.2 & 40.0 & 42.2 & 75.0 & 50.0 & 0.0 & 100.0 & 51.3 & 55.6 & 43.9 \\
\midrule
\multicolumn{11}{c}{\textit{Ours}} \\
\midrule
\iconQwenThreeFive\textsc{Qwen3.5-35B-A3B} (base) & 21.9 & 20.0 & 35.6 & 75.0 & 50.0 & 0.0 & 100.0 & 35.9 & 44.4 & 33.8 \\
\iconKallisti\model{} & 21.9 & 40.0 & 46.7 & 50.0 & 50.0 & 0.0 & 100.0 & 33.3 & 44.4 & 36.7 \\
\bottomrule
\end{tabular}%
}
\end{table}

\medskip
\noindent\textbf{\texttt{TB-1} \texttt{ap@1}.}\par
Tables~\ref{tab:app:finetune_topic_tb1_ap1_fine-art}, \ref{tab:app:finetune_topic_tb1_ap1_illustration}, and \ref{tab:app:finetune_topic_tb1_ap1_photography} report topic-level \texttt{TB-1} \texttt{ap@1} results for fine art, illustration, and photography, respectively.

\begin{table}[t]
\centering
\caption{Topic-level fine-tuning \texttt{TB-1} \texttt{ap@1} results on fine-art tasks. The final column reports the corresponding domain-level average (\%).}
\label{tab:app:finetune_topic_tb1_ap1_fine-art}
\resizebox{\textwidth}{!}{%
\scriptsize
\setlength{\tabcolsep}{3pt}
\renewcommand{\arraystretch}{1.02}
\begin{tabular}{@{}c c c c c c c c c c c@{}}
\toprule
Model & Calligraphy & Chinese Painting & Ink \& Wash & Landscape Color & Portrait Color & Portrait Sketch & Quick Sketch & Still Life Color & Still Life Sketch & Avg. \\
\midrule
\multicolumn{11}{c}{\textit{Open-Weight Models}} \\
\midrule
\iconGLM\textsc{GLM-4.6V} & 21.2 & 63.0 & 33.3 & 17.3 & 66.7 & 52.5 & 54.2 & 58.3 & 52.4 & 39.3 \\
\iconKimi\textsc{Kimi-K2.5} & 30.3 & 85.2 & 50.0 & 13.1 & 100.0 & 46.5 & 62.5 & 50.0 & 52.4 & 38.5 \\
\iconQwenVL\textsc{Qwen3-VL-235B-A22B} & 15.2 & 85.2 & 50.0 & 10.7 & 66.7 & 48.5 & 66.7 & 47.2 & 57.1 & 36.9 \\
\iconQwenThreeFive\textsc{Qwen3.5-397B-A17B} & 24.2 & 96.3 & 66.7 & 18.5 & 100.0 & 37.4 & 58.3 & 43.1 & 45.2 & 37.3 \\
\midrule
\multicolumn{11}{c}{\textit{Ours}} \\
\midrule
\iconQwenThreeFive\textsc{Qwen3.5-35B-A3B} (base) & 21.2 & 85.2 & 75.0 & 16.7 & 83.3 & 41.4 & 75.0 & 37.5 & 47.6 & 36.9 \\
\iconKallisti\model{} & 24.2 & 81.5 & 58.3 & 17.9 & 66.7 & 52.5 & 75.0 & 47.2 & 50.0 & 40.6 \\
\bottomrule
\end{tabular}%
}
\end{table}

\input{appendix_tables/tab_app_finetune_topic_tb1_ap1_illustration}

\input{appendix_tables/tab_app_finetune_topic_tb1_ap1_photography}

\medskip
\noindent\textbf{\texttt{TB-1} \texttt{pass\textasciicircum{}3}.}\par
Tables~\ref{tab:app:finetune_topic_fine_art}, \ref{tab:app:finetune_topic_illustration}, and \ref{tab:app:finetune_topic_photography} report topic-level \texttt{TB-1} \texttt{pass\textasciicircum{}3} results for fine art, illustration, and photography, respectively.

\input{appendix_tables/tab_app_finetune_topic_fine_art}

\input{appendix_tables/tab_app_finetune_topic_illustration}

\input{appendix_tables/tab_app_finetune_topic_photography}

\clearpage

\section{Prompt Templates and Scoring Interfaces}
\label{app:prompts}

This appendix records the prompt templates and scoring interfaces used in data construction and evaluation. We group them by usage: prompts for the photography pipeline, prompts for the illustration pipeline, prompts for MLLM evaluation, and scoring interfaces for reward-model evaluation.

\subsection{Photography Pipeline Prompts}
\label{app:prompts:photo}

This subsection records the prompts used in the photography agent-edit pipeline described in Section~\ref{sec:benchmark}. The pipeline consists of three stages: aspect-wise critique generation with ArtiMuse, edit planning with GPT through LiteLLM, and image editing with Gemini-3-Pro-Image (Nano Banana Pro). In Step~1, ArtiMuse is queried eight times independently with the same image, once per aesthetic aspect. The resulting responses are then concatenated into a single review string that is passed to the planning stage.

\subsubsection{Step 1: ArtiMuse Aspect-Wise Critique}

\begin{promptbox}[title=ArtiMuse Aspect-Wise Critique Calls]
\small
The same image is attached in every call. The following eight prompts are sent independently:
\begin{enumerate}[leftmargin=*,label=\arabic*.]
\item Please evaluate the aesthetic quality of this image from the aspect of Composition \& Design.
\item Please evaluate the aesthetic quality of this image from the aspect of Visual Elements \& Structure.
\item Please evaluate the aesthetic quality of this image from the aspect of Technical Execution.
\item Please evaluate the aesthetic quality of this image from the aspect of Originality \& Creativity.
\item Please evaluate the aesthetic quality of this image from the aspect of Theme \& Communication.
\item Please evaluate the aesthetic quality of this image from the aspect of Emotion \& Viewer Response.
\item Please evaluate the aesthetic quality of this image from the aspect of Overall Gestalt.
\item Please evaluate the aesthetic quality of this image from the aspect of Comprehensive Evaluation.
\end{enumerate}
\end{promptbox}

\begin{promptbox}[title=Critique Aggregation Format]
\small
The eight responses are concatenated into a single \texttt{review} string in the following format:

\medskip
{\ttfamily\raggedright
[Composition \& Design] <response>\par\smallskip
[Visual Elements \& Structure] <response>\par\smallskip
[Technical Execution] <response>\par\smallskip
[Originality \& Creativity] <response>\par\smallskip
[Theme \& Communication] <response>\par\smallskip
[Emotion \& Viewer Response] <response>\par\smallskip
[Overall Gestalt] <response>\par\smallskip
[Comprehensive Evaluation] <response>\par
}
\end{promptbox}

\subsubsection{Step 2: Edit Planning}

The planning stage is executed with GPT through LiteLLM. Below we reproduce the literal runtime system and user messages. The placeholder \texttt{\{review\}} denotes the full ArtiMuse output obtained by concatenating the eight aspect-wise responses from Step~1. We retain the original runtime field names, including \texttt{substyle}.

\begin{promptbox}[title=Planning System Message]
\small
\ttfamily
\raggedright
You are an expert aesthetic analyst and image editing planner for creating aesthetic preference benchmark datasets.\par
\end{promptbox}

\begin{promptbox}[title=Planning User Message]
\small
\ttfamily
\raggedright
Analyze this photograph and plan a synthetic edit for an aesthetic preference dataset.\par
\par
AESTHETIC ASSESSMENT:\par
\{review\}\par
\par
AVAILABLE CLASSIFICATIONS:\par
\par
Domains: Photo\par
\par
Substyles:\par
\{\par
\hspace*{1em}"Photo": [\par
\hspace*{2em}"Portrait",\par
\hspace*{2em}"Landscape",\par
\hspace*{2em}"StreetCity",\par
\hspace*{2em}"Architecture",\par
\hspace*{2em}"Wildlife",\par
\hspace*{2em}"Macro",\par
\hspace*{2em}"FoodProduct",\par
\hspace*{2em}"NightAstro",\par
\hspace*{2em}"Sports"\par
\hspace*{1em}]\par
\}\par
\par
TASK:\par
1. Classify the image: choose the most appropriate domain and substyle based on image itself\par
2. Freely decide how to edit this image to create better and worse versions\par
3. Write a "better" prompt: describe how to edit THIS image to make it aesthetically better\par
4. Write a "worse" prompt: describe how to edit THIS image to make it aesthetically worse\par
\par
- "Better" means: take the current image and improve its quality\par
- "Worse" means: take the current image and degrade its quality.\par
\par
EDIT Types - Do NOT always default to color/tone adjustments. Actively consider all edit types:\par
\par
1. **Color \& Tonal Adjustments**:\par
\hspace*{1em}- White balance, color cast correction or introduction\par
\hspace*{1em}- Saturation, vibrance, contrast adjustments\par
\hspace*{1em}- Color grading and mood shifts\par
\par
2. **Perspective \& Viewpoint Changes**:\par
\hspace*{1em}- Simulate a slightly different camera angle or position\par
\hspace*{1em}- Adjust the apparent depth or distance from the subject\par
\hspace*{1em}- Change the sense of scale or spatial relationships\par
\par
3. **Element Removal**:\par
\hspace*{1em}- Remove contextual elements that enhance (or clutter) the scene\par
\hspace*{1em}- Modify background elements to improve or degrade composition\par
\par
4. **Compositional Restructuring**:\par
\hspace*{1em}- Adjust cropping or framing emphasis\par
\hspace*{1em}- Rebalance visual weight distribution\par
\hspace*{1em}- Modify leading lines or focal points\par
\par
5. **Light \& Shadow Changes**:\par
\hspace*{1em}- Modify shadow depth, highlight recovery, or exposure\par
\hspace*{1em}- Change lighting direction or intensity\par
\hspace*{1em}- Adjust local brightness to guide visual attention\par
\par
Choose multiple edit types that best address the specific strengths and weaknesses identified in the aesthetic assessment.\par
\par
WORSE PROMPT RESTRICTIONS:\par
When creating the "worse" prompt, DO NOT use the following editing techniques:\par
- DO NOT add lens flare or light flares\par
- DO NOT add vignetting (darkened corners)\par
- DO NOT add object outlines, glows, or halos around subjects\par
These techniques are overused and create unrealistic degradation.\par
\par
PROMPT WRITING GUIDELINES:\par
- Describe the scene narratively, don't just list keywords, a narrative, descriptive paragraph will almost always produce a better, more coherent image than a list of disconnected words.\par
- Use photographic and cinematic language terms. Mention camera angles, lens types, lighting, and fine details to guide the model toward a photorealistic result.\par
- Explain how changes should integrate into the scene (lighting, style, composition)\par
- For adding/removing elements: describe what to add/remove and how it should blend in\par
- For style changes: describe both the target style and what to preserve\par
\par
Return JSON:\par
\{\par
\hspace*{1em}"content": \{"domain": "Photo", "substyle": "..."\},\par
\hspace*{1em}"planned\_prompts": \{\par
\hspace*{2em}"better": "...",\par
\hspace*{2em}"worse": "..."\par
\hspace*{1em}\},\par
\hspace*{1em}"reasoning": "Explain your substyle classification choice and why you chose these specific photographic edits for better/worse versions"\par
\}\par
\end{promptbox}

\subsubsection{Step 3: Image Editing}

The final stage uses Gemini-3-Pro-Image (Nano Banana Pro) to synthesize the edited variants. Round~1 generates the better image from the original image alone. Round~2 generates the worse image from the original image while conditioning on the better image as additional reference context. For Round~2, we report the file-backed with-context template actually used at runtime. This is the operative version loaded by the pipeline and therefore excludes the fallback sentence beginning with ``Important: Apply this edit to the ORIGINAL image'' that appears only in the default code path.

\begin{promptbox}[title=Round 1: Better-Image Editing Prompt]
\small
You are an expert image editor. Edit the provided image according to these instructions:

\medskip
\texttt{\{edit\_instruction\}}
\end{promptbox}

\begin{promptbox}[title=Round 2: Worse-Image Editing Prompt with Context]
\small
You are an expert image editor.

\medskip
Context: Previously, the following improvement was made to the original image:\\
\texttt{"\{context\_prompt\}"}\\
(The improved result is shown in the second image for reference)

\medskip
Now, edit the \textbf{ORIGINAL} image (first image) according to these instructions:\\
\texttt{"\{edit\_instruction\}"}
\end{promptbox}

\subsection{Illustration Pipeline Prompts}
\label{app:prompts:illustration}

This subsection records the prompt templates used in the prompt-based illustration generation pipeline. We use one template for digital and AI art, concept art, and pixel art, and a separate storyboard-style template for anime and manga and comic generation. The stylized 3D subset is constructed through asset rendering rather than text-to-image prompting.

\subsubsection{Digital and AI Art / Concept Art / Pixel Art Generation}

For digital and AI art, concept art, and pixel art generation, we use a structured template to control semantic content while allowing variation in visual execution. Each prompt is composed in a fixed order: subject or character, motion or action, environment or background, lighting condition, atmosphere or mood, camera or composition, and art style.

\begin{promptbox}[title=Illustration Structured Prompt Template]
Generate 25 concise prompts ready for text-to-image artwork generation on the topic of \texttt{[ART CATEGORY]}. Each prompt should describe a clear visual scene using the following order: subject or character, motion, environment/background, lighting condition, atmosphere or mood, camera or composition (view, framing, depth), and art style.
\end{promptbox}

\subsubsection{Manga / Comic Storyboard Generation}

For anime and manga and comic generation, we adopt a storyboard-style template with stronger structural constraints. Each prompt describes a short multi-panel narrative with consistent characters and environment, while each panel specifies shot type, character action, key props, and scene details.

\begin{promptbox}[title=Illustration Storyboard Prompt Template]
Generate 25 concise storyboard prompts for \texttt{[ART CATEGORY]} artwork generation. Each prompt should describe a micro-story using panels. Anchors such as character identity and environment must remain consistent. Each panel must specify shot type (wide, medium, close-up), character actions, key props, and environment details. Keep the description visual and concise.
\end{promptbox}

\subsection{MLLM Evaluation Prompts}
\label{app:prompts:eval}

This subsection reports the generic prompts used to evaluate MLLMs on \bench{}. Depending on the task setting, the model is asked to identify the best image, the worst image, or both the best and the worst images in a candidate set.

\subsubsection{Best-Image Selection Prompt}

\begin{promptbox}[title=Pick-Best Prompt]
\small
You will be shown \verb|{n}| images labeled \verb|{labels}|. Evaluate the overall aesthetic quality of each image, considering composition, color, technique, and artistic expression.

\medskip
Which image has the best overall aesthetic quality?\\
\verb|{options}|

\medskip
You may reason step by step. Then, on the very last line of your response, write your final answer in exactly this format:

\medskip
\texttt{BEST: X}

\medskip
where \texttt{X} is the letter label of the best image.
\end{promptbox}

\subsubsection{Worst-Image Selection Prompt}

\begin{promptbox}[title=Pick-Worst Prompt]
\small
You will be shown \verb|{n}| images labeled \verb|{labels}|. Evaluate the overall aesthetic quality of each image, considering composition, color, technique, and artistic expression.

\medskip
Which image has the worst overall aesthetic quality?\\
\verb|{options}|

\medskip
You may reason step by step. Then, on the very last line of your response, write your final answer in exactly this format:

\medskip
\texttt{WORST: X}

\medskip
where \texttt{X} is the letter label of the worst image.
\end{promptbox}

\subsubsection{Joint Best-and-Worst Selection Prompt}

\begin{promptbox}[title=Pick-Best-and-Worst Prompt]
\small
You will be shown \verb|{n}| images labeled \verb|{labels}|. Evaluate the overall aesthetic quality of each image, considering composition, color, technique, and artistic expression.

\medskip
Which image has the best overall aesthetic quality, and which has the worst?\\
\verb|{options}|

\medskip
You may reason step by step. Then, on the very last lines of your response, write your final answers in exactly this format:

\medskip
\texttt{BEST: X}\\
\texttt{WORST: Y}

\medskip
where \texttt{X} and \texttt{Y} are the letter labels.
\end{promptbox}

\subsection{Reward-Model Scoring Interfaces}
\label{app:prompts:rm}

This subsection records the model-specific scoring interfaces used to evaluate reward models on \bench{}. Unlike MLLMs, reward models do not share a single comparative prompt. Instead, each model scores images independently following its released evaluation method, and we convert the resulting per-image scores into within-task rankings. Across all reward models, each image is scored \texttt{num\_trials} times, the scores are averaged, and the averaged scores are used to determine the predicted best and worst images. We use the following terminology: \emph{explicit prompt} denotes prompt text exposed at the call site, \emph{implicit prompt} denotes prompt text embedded inside the model library, and \emph{no prompt} denotes a pure vision scorer with no language instruction.

\subsubsection{ArtiMuse}

ArtiMuse uses an \emph{implicit prompt}: the instruction is hardcoded inside the released library rather than passed from our evaluation script.

\begin{promptbox}[title=ArtiMuse Internal Scoring Prompt]
\small
Rate the aesthetics score of the image in 0-100.

\medskip
In the output format, numbers are replaced by 2 corresponding letters, and the mapping relationship is:

\medskip
score 0 to 25: 0-aa, 1-ab, 2-ac, 3-ad, \ldots, 25-az,\\
score 26 to 50: 26-ca, 27-cb, 28-cc, 29-cd, \ldots, 50-cy,\\
score 51 to 75: 51-da, 52-db, 53-dc, 54-dd, \ldots, 75-dy,\\
score 76 to 100: 76-ea, 77-eb, 78-ec, 79-ed, \ldots, 100-ey.

\medskip
The answer only outputs 2 corresponding letters.
\end{promptbox}

At runtime, the library automatically prepends an image token before the instruction. The model outputs a two-letter code, which is then decoded back to a numerical score in $[0, 100]$.

\subsubsection{Q-Align}

Q-Align also uses an \emph{implicit prompt}. In our experiments, we use the aesthetic scorer by default; the library additionally exposes a quality scorer.

\begin{promptbox}[title=Q-Align Internal Prompts]
\small
\textbf{Aesthetic scorer (used by default):}

\medskip
\texttt{USER: How would you rate the aesthetics of this image?}\\
\texttt{<|image|>}\\
\texttt{ASSISTANT: The aesthetics of the image is}

\medskip
\textbf{Quality scorer (alternative interface):}

\medskip
\texttt{USER: How would you rate the quality of this image?}\\
\texttt{<|image|>}\\
\texttt{ASSISTANT: The quality of the image is}
\end{promptbox}

The scorer reads the last-token logits over the five words \texttt{excellent}, \texttt{good}, \texttt{fair}, \texttt{poor}, and \texttt{bad}, applies a softmax, and computes a weighted sum with weights $[1, 0.75, 0.5, 0.25, 0]$.

\subsubsection{Q-Instruct}

Q-Instruct uses an \emph{explicit prompt} passed through the command-line interface.

\begin{promptbox}[title=Q-Instruct Evaluation Prompt]
\small
Rate the quality of the image.
\end{promptbox}

The instruction is wrapped in the \texttt{mplug\_owl2} conversation format with a prepended \texttt{DEFAULT\_IMAGE\_TOKEN}. The default score is the logit difference between \texttt{good} and \texttt{poor}.

\subsubsection{RealQA}

RealQA uses an \emph{explicit prompt}. Under the AVA adapter used in our experiments, the default instruction is:

\begin{promptbox}[title=RealQA Evaluation Prompt]
\small
Rate the aesthetics and quality of this image briefly.
\end{promptbox}

The model generates free text, from which a numerical score in $[0.0, 10.0]$ is parsed. If parsing fails, the library issues a second constrained prompt:

\begin{promptbox}[title=RealQA Retry Prompt]
\small
Output only one numeric score. Range: 0.0 to 10.0. No words, no explanation.
\end{promptbox}

\subsubsection{PEAS}

PEAS-aes and PEAS-ava use \emph{no prompt}. They are pure vision models with no language instruction. The scorer outputs a 10-bin distribution over scores from 1 to 10, and the final score is the expected value of that distribution.

\begin{promptbox}[title=PEAS Scoring Interface]
\small
No textual prompt is used. The model outputs a 10-bin score distribution over the range 1 to 10, and the final scalar score is computed as the expected value under that distribution.
\end{promptbox}

\section{Topic-Level Annotation Examples}
\label{app:annotation_examples}

This appendix provides one representative annotation example for each of the 24 topics in \bench{}, grouped by domain.

\subsection{Fine Art}
\label{app:annotation_examples:fineart}

\par\medskip
\noindent\begin{minipage}{\textwidth}
  \raggedright\textbf{Landscape Color}\par
  \medskip
  \centering
  \includegraphics[width=0.85\textwidth]{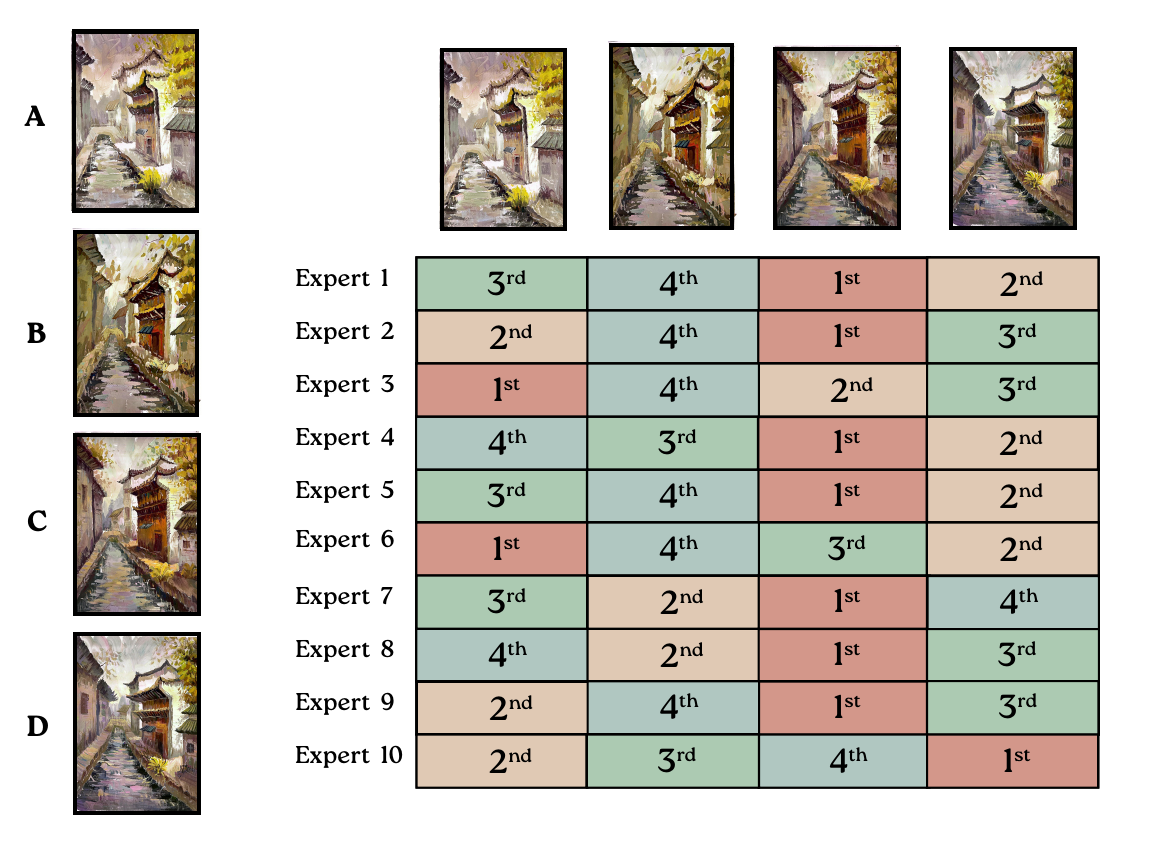}
\end{minipage}
\medskip
\par\medskip
\noindent\begin{minipage}{\textwidth}
  \raggedright\textbf{Still Life Color}\par
  \medskip
  \centering
  \includegraphics[width=0.85\textwidth]{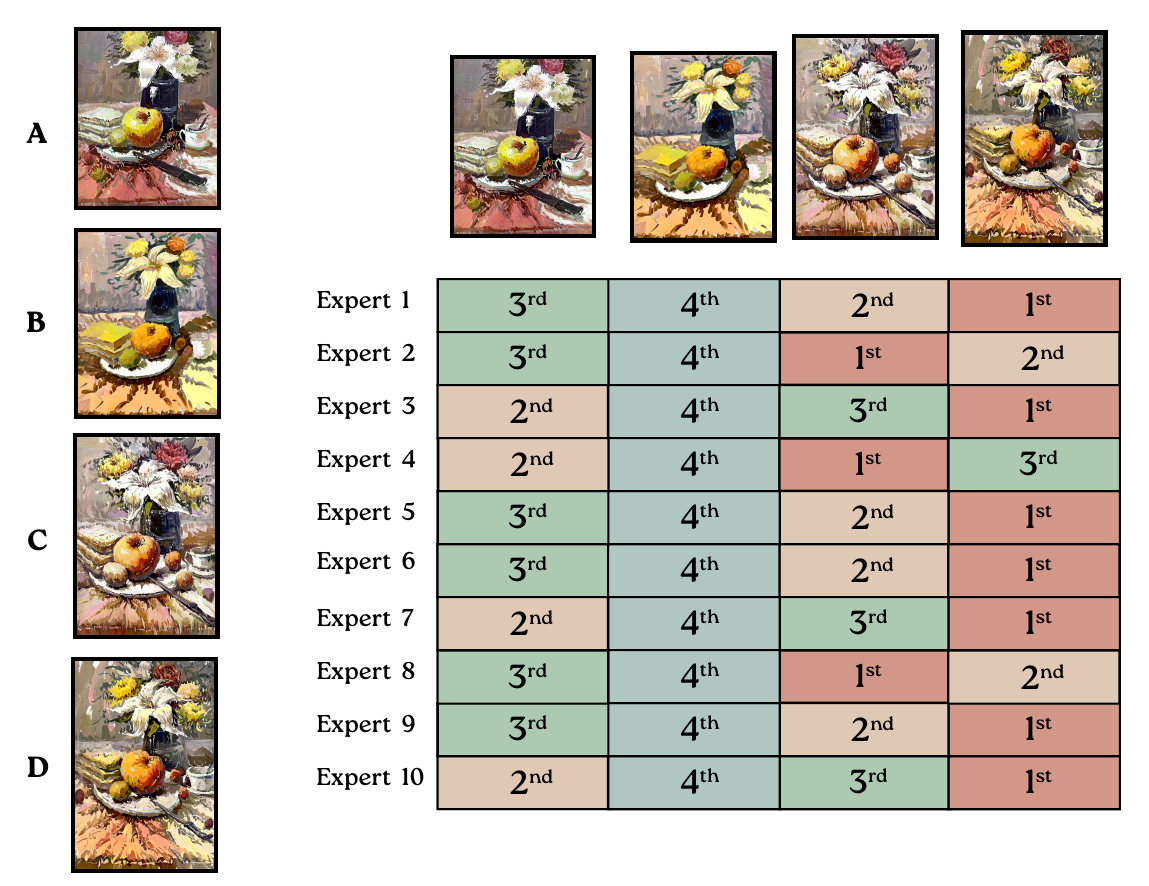}
\end{minipage}
\medskip

\par\medskip
\noindent\begin{minipage}{\textwidth}
  \raggedright\textbf{Calligraphy}\par
  \medskip
  \centering
  \includegraphics[width=0.85\textwidth]{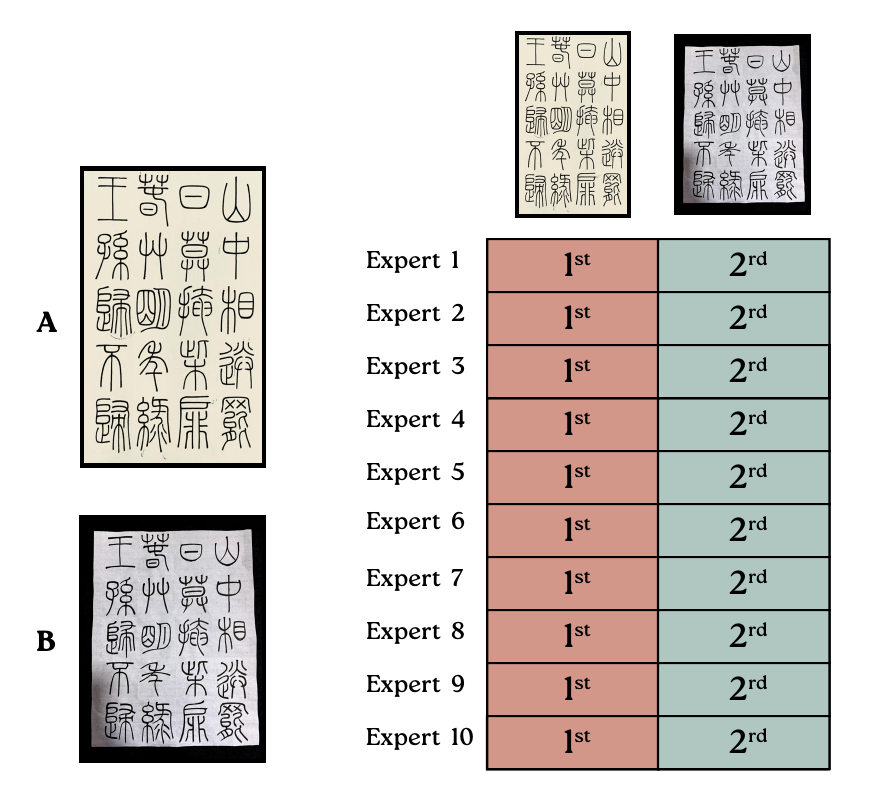}
\end{minipage}
\medskip
\par\medskip
\noindent\begin{minipage}{\textwidth}
  \raggedright\textbf{Chinese Painting}\par
  \medskip
  \centering
  \includegraphics[width=0.86\textwidth]{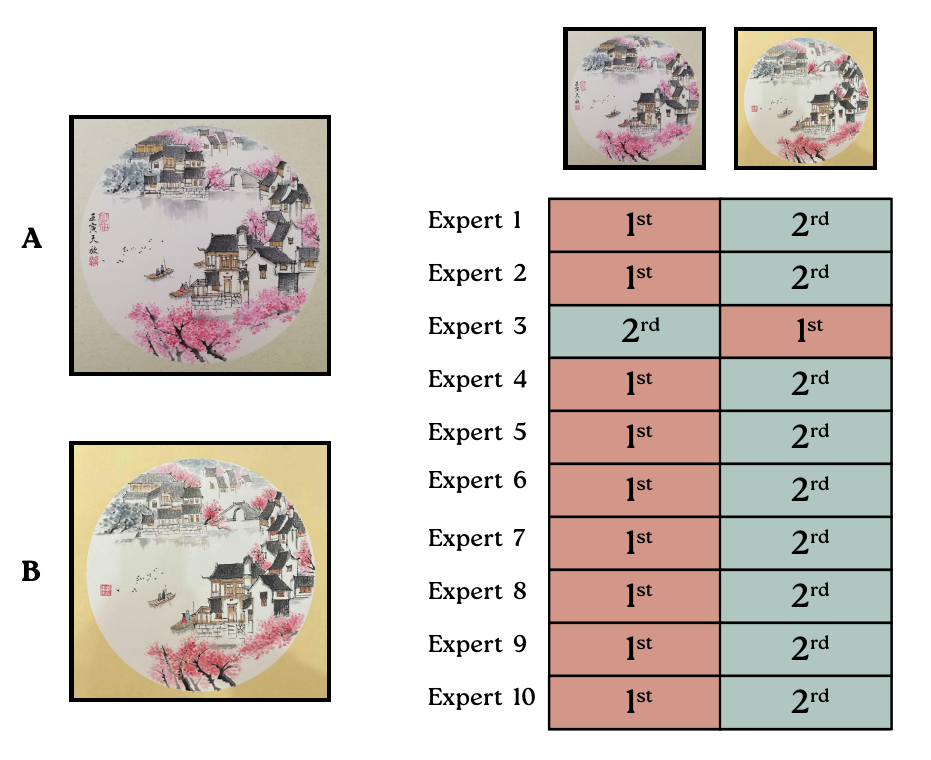}
\end{minipage}
\medskip
\par\medskip
\noindent\begin{minipage}{\textwidth}
  \raggedright\textbf{Ink and Wash}\par
  \medskip
  \centering
  \includegraphics[width=0.86\textwidth]{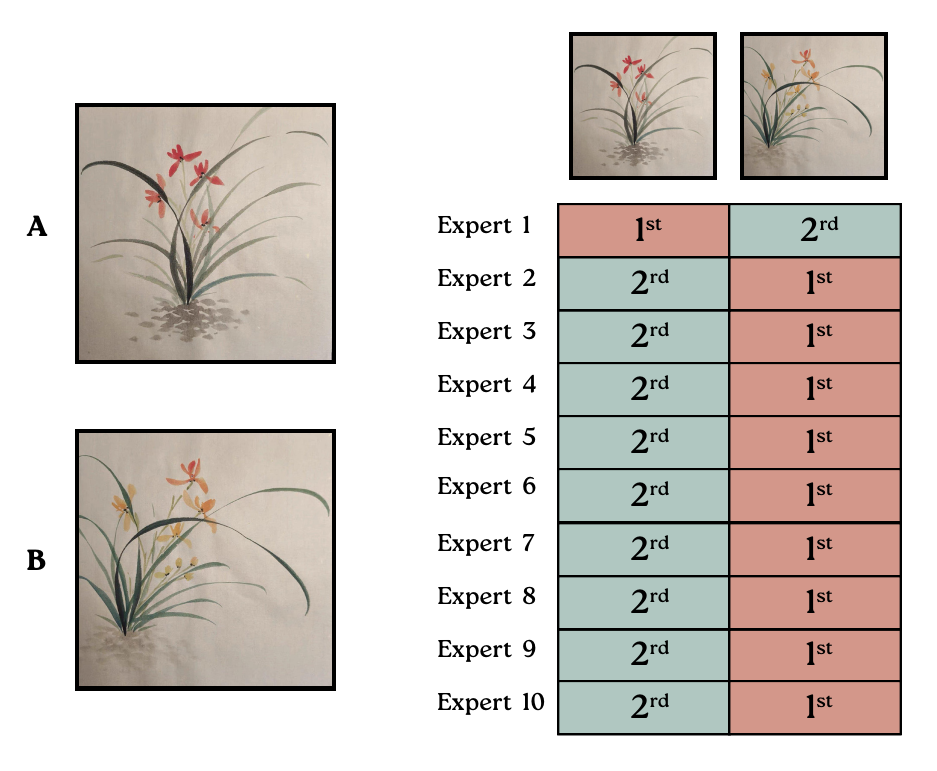}
\end{minipage}
\medskip
\par\medskip
\noindent\begin{minipage}{\textwidth}
  \raggedright\textbf{Portrait Color}\par
  \medskip
  \centering
  \includegraphics[width=0.8\textwidth]{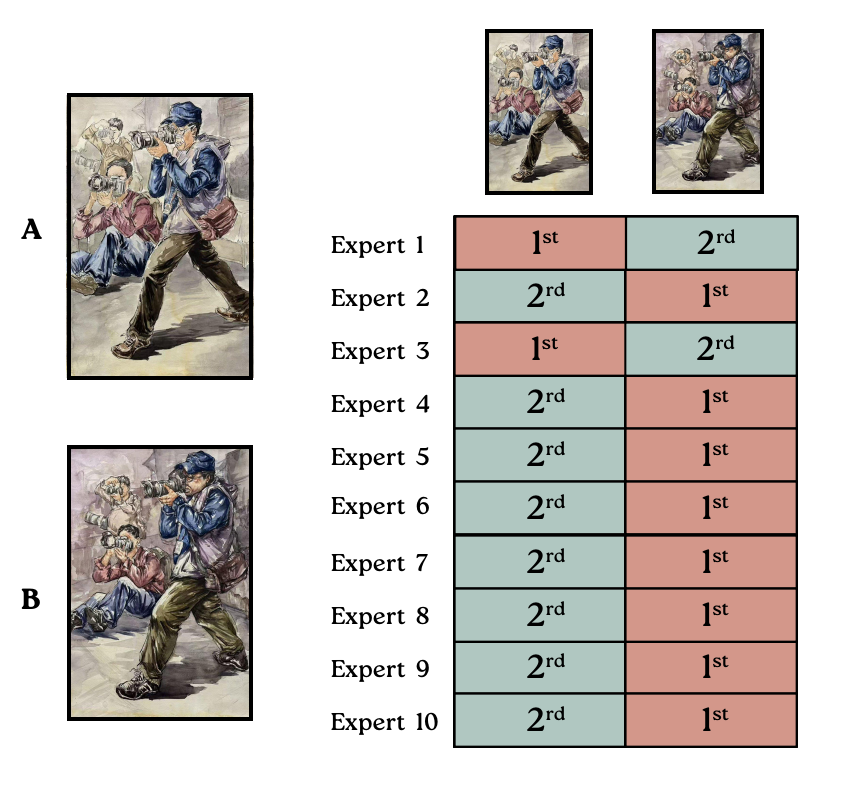}
\end{minipage}
\medskip

\par\medskip
\noindent\begin{minipage}{\textwidth}
  \raggedright\textbf{Portrait Sketch}\par
  \medskip
  \centering
  \includegraphics[width=0.80\textwidth]{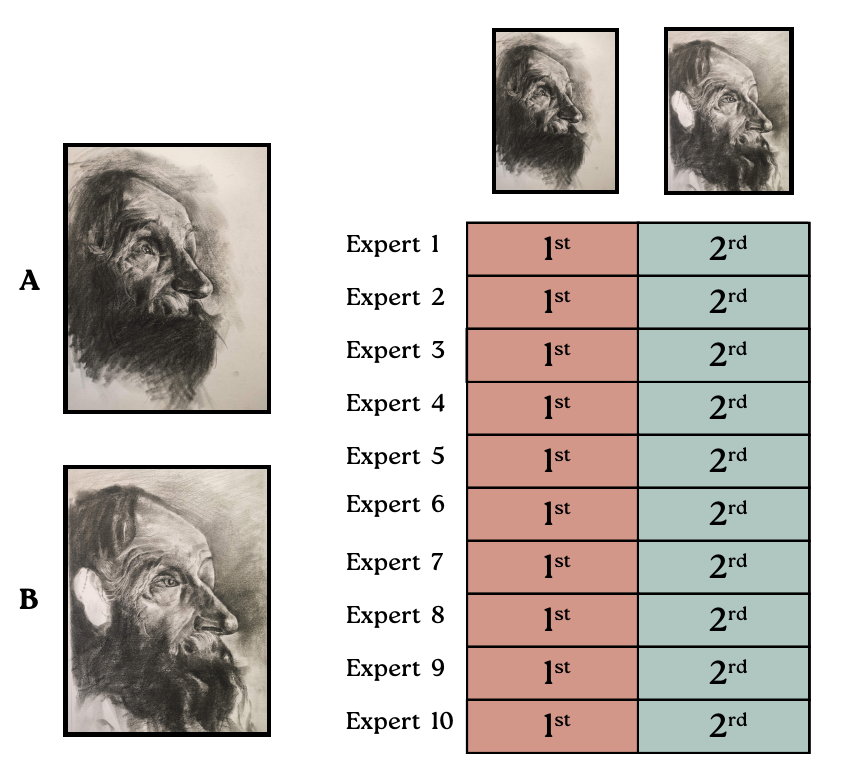}
\end{minipage}
\medskip
\par\medskip
\noindent\begin{minipage}{\textwidth}
  \raggedright\textbf{Quick Sketch}\par
  \medskip
  \centering
  \includegraphics[width=0.86\textwidth]{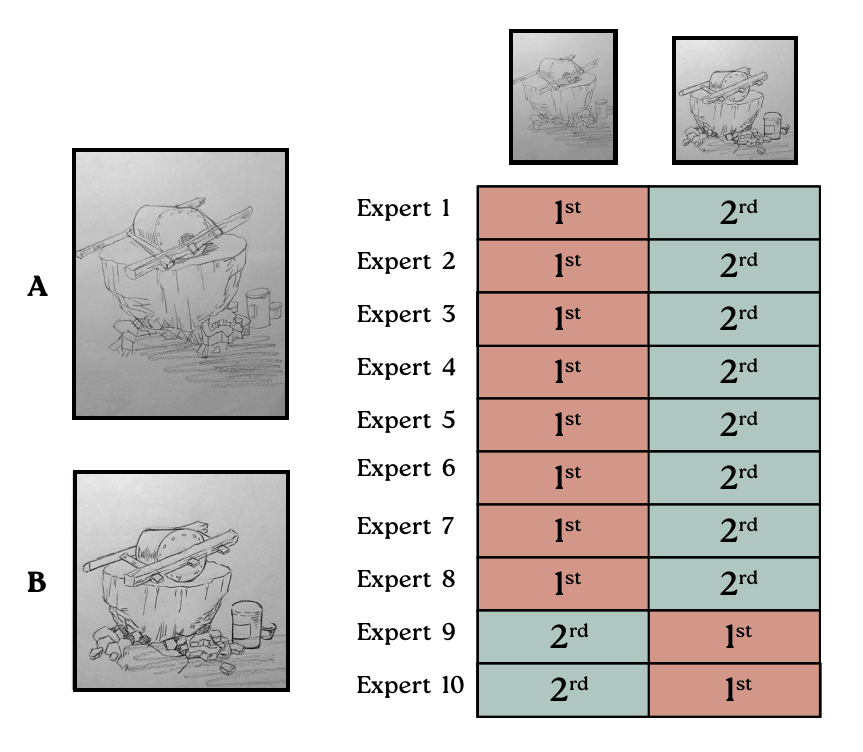}
\end{minipage}
\medskip
\par\medskip
\noindent\begin{minipage}{\textwidth}
  \raggedright\textbf{Still Life Sketch}\par
  \medskip
  \centering
  \includegraphics[width=0.78\textwidth]{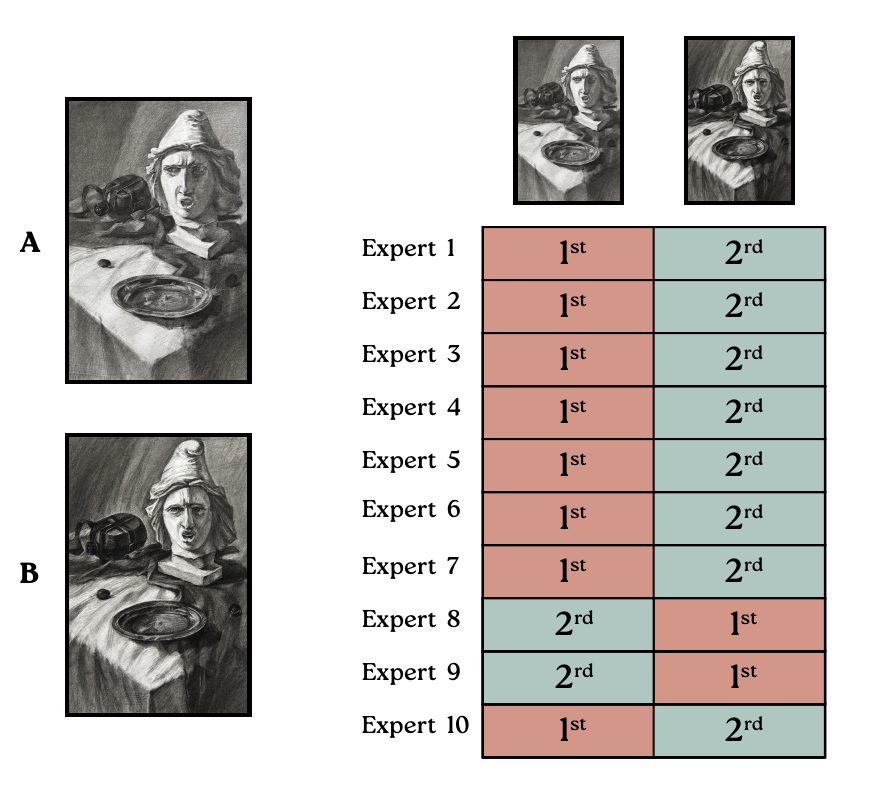}
\end{minipage}
\medskip

\subsection{Photography}
\label{app:annotation_examples:photo}

\par\medskip
\noindent\begin{minipage}{\textwidth}
  \raggedright\textbf{Architecture}\par
  \medskip
  \centering
  \includegraphics[width=0.78\textwidth]{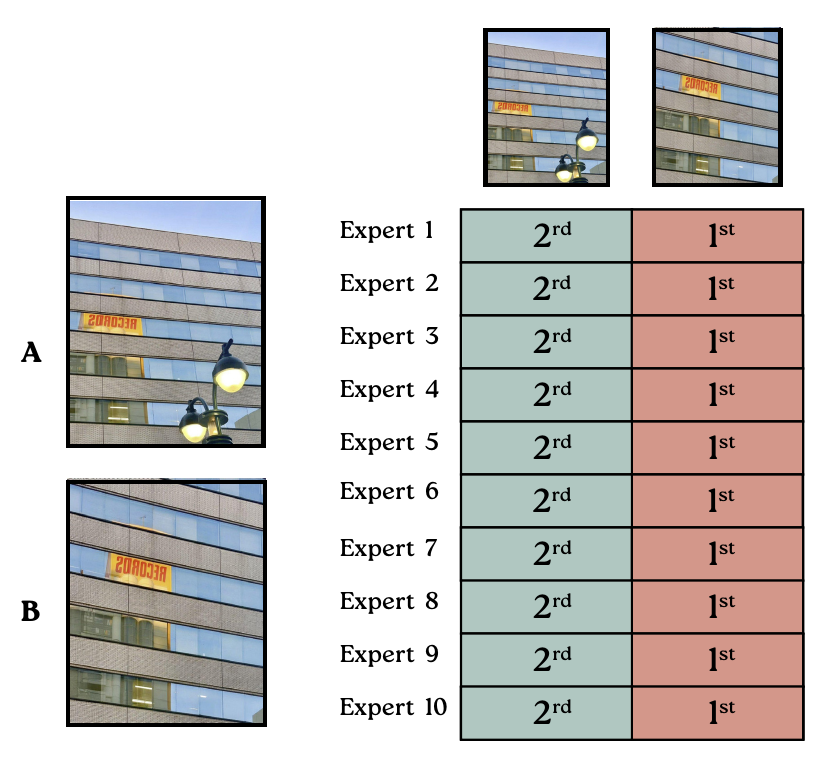}
\end{minipage}
\medskip
\par\medskip
\noindent\begin{minipage}{\textwidth}
  \raggedright\textbf{Food and Product}\par
  \medskip
  \centering
  \includegraphics[width=0.80\textwidth]{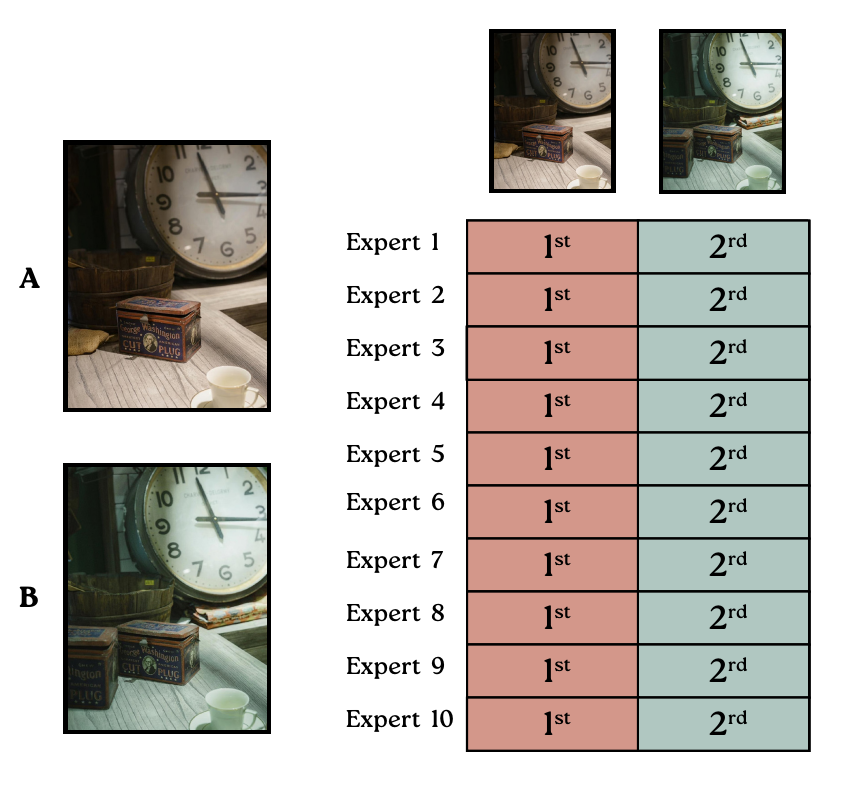}
\end{minipage}
\medskip
\par\medskip
\noindent\begin{minipage}{\textwidth}
  \raggedright\textbf{Landscape}\par
  \medskip
  \centering
  \includegraphics[width=0.9\textwidth]{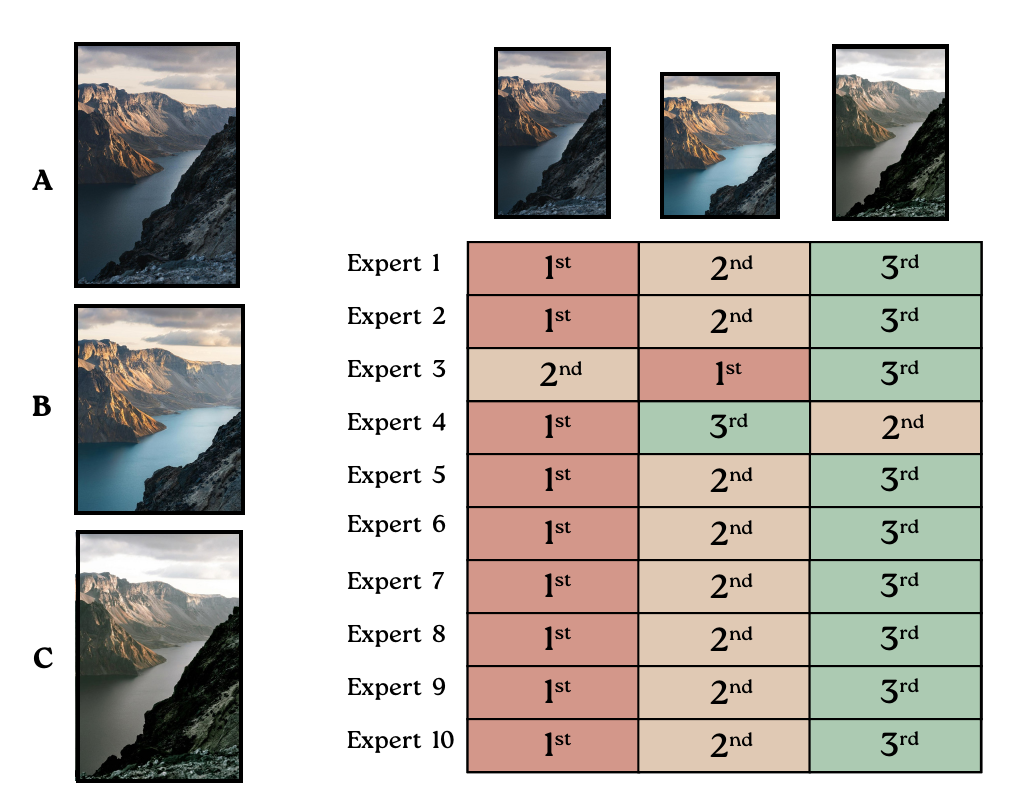}
\end{minipage}
\medskip
\par\medskip
\noindent\begin{minipage}{\textwidth}
  \raggedright\textbf{Macro}\par
  \medskip
  \centering
  \includegraphics[width=0.9\textwidth]{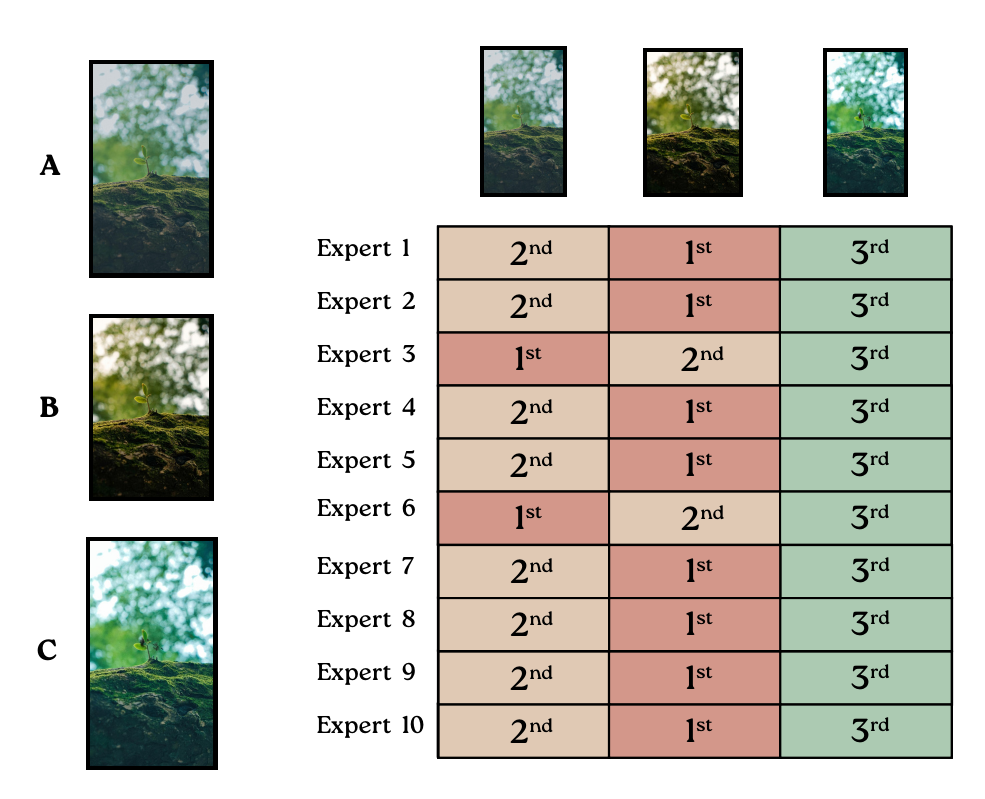}
\end{minipage}
\medskip
\par\medskip
\noindent\begin{minipage}{\textwidth}
  \raggedright\textbf{Night and Astrophotography}\par
  \medskip
  \centering
  \includegraphics[width=0.85\textwidth]{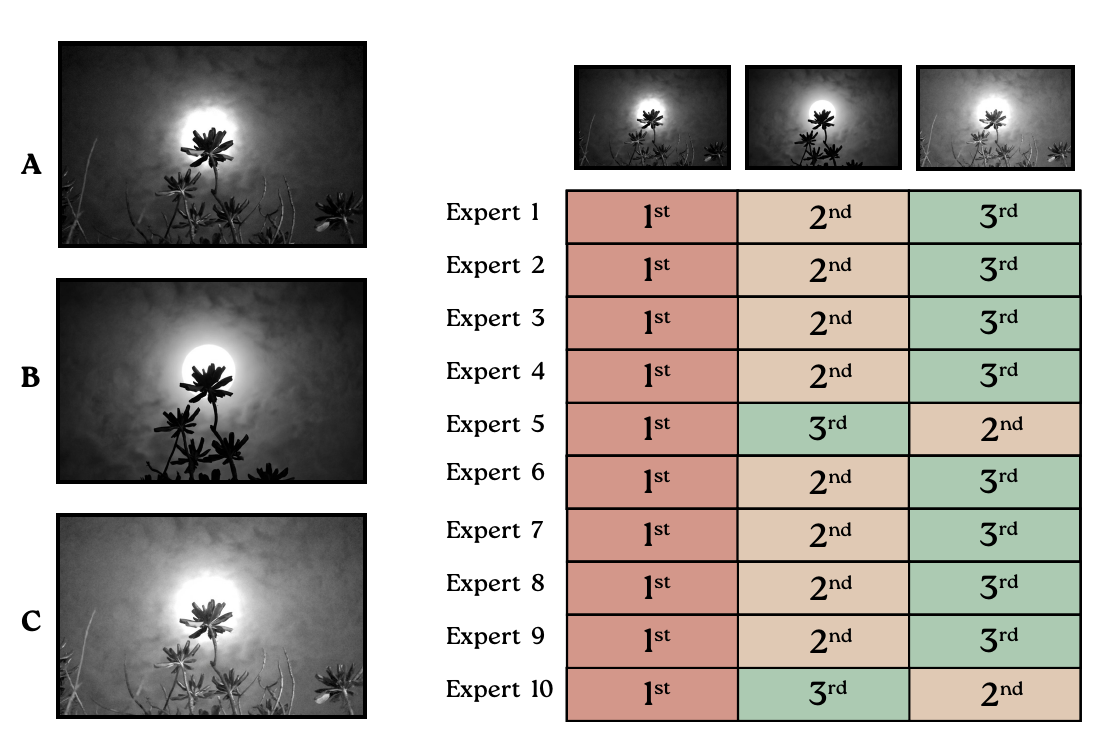}
\end{minipage}
\medskip
\par\medskip
\noindent\begin{minipage}{\textwidth}
  \raggedright\textbf{Portrait}\par
  \medskip
  \centering
  \includegraphics[width=0.85\textwidth]{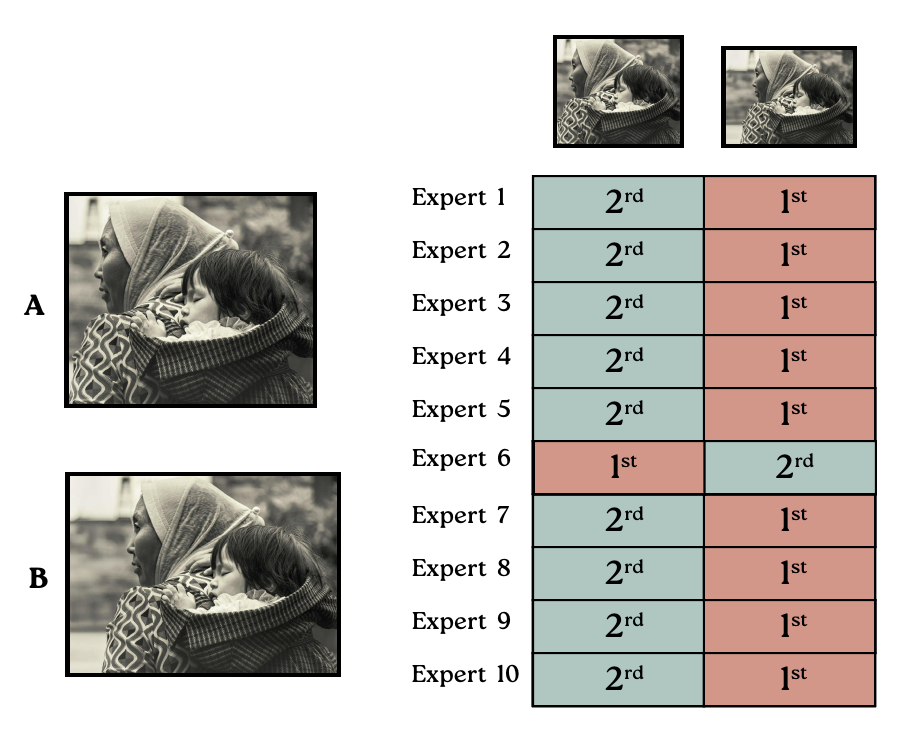}
\end{minipage}
\medskip
\par\medskip
\noindent\begin{minipage}{\textwidth}
  \raggedright\textbf{Sports}\par
  \medskip
  \centering
  \includegraphics[width=0.86\textwidth]{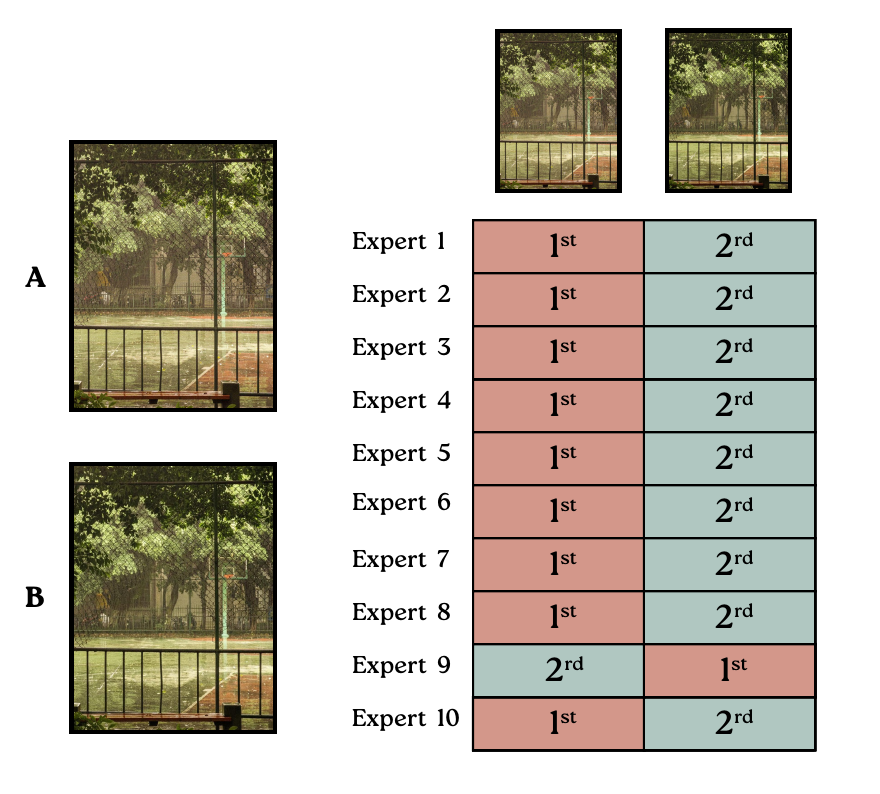}
\end{minipage}
\medskip
\par\medskip
\noindent\begin{minipage}{\textwidth}
  \raggedright\textbf{Street and City}\par
  \medskip
  \centering
  \includegraphics[width=0.9\textwidth]{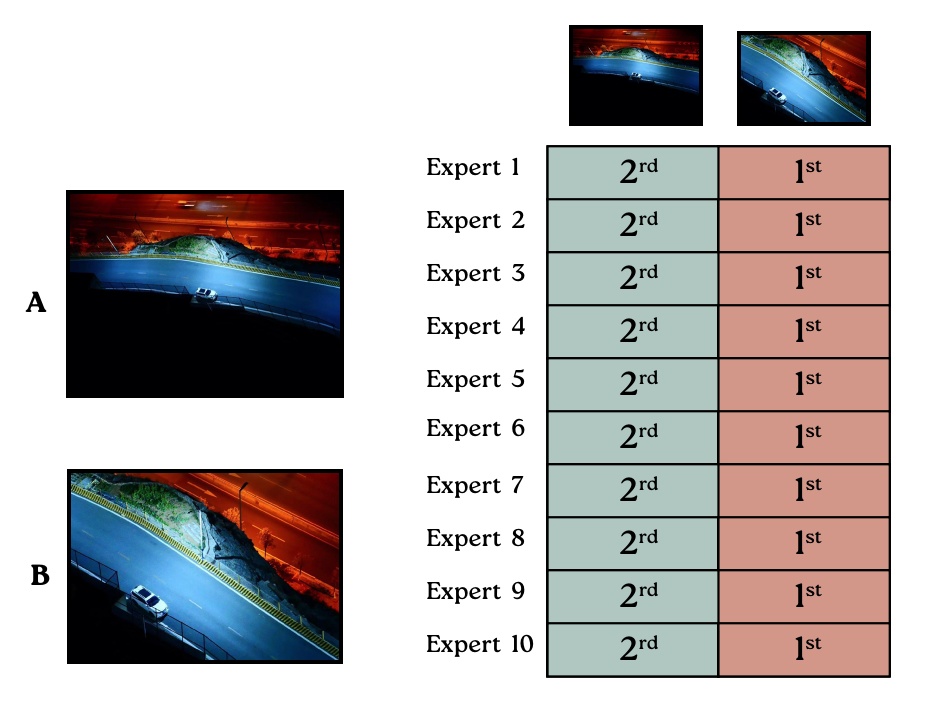}
\end{minipage}
\medskip
\par\medskip
\noindent\begin{minipage}{\textwidth}
  \raggedright\textbf{Wildlife}\par
  \medskip
  \centering
  \includegraphics[width=0.9\textwidth]{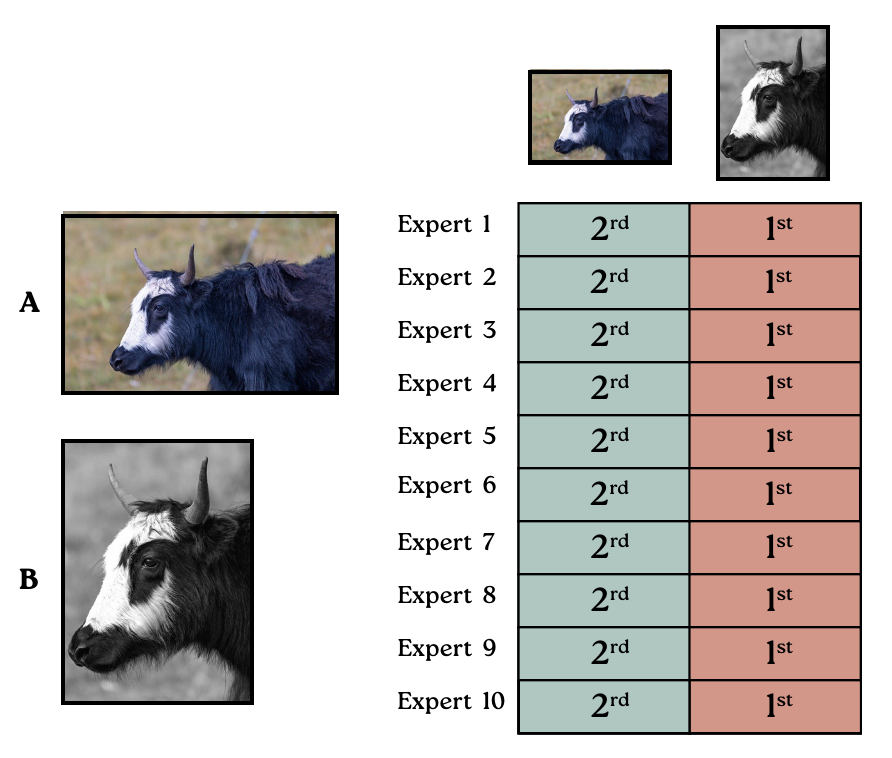}
\end{minipage}
\medskip

\subsection{Illustration}
\label{app:annotation_examples:illustration}

\par\medskip
\noindent\begin{minipage}{\textwidth}
  \raggedright\textbf{Anime and Manga}\par
  \medskip
  \centering
  \includegraphics[width=0.87\textwidth]{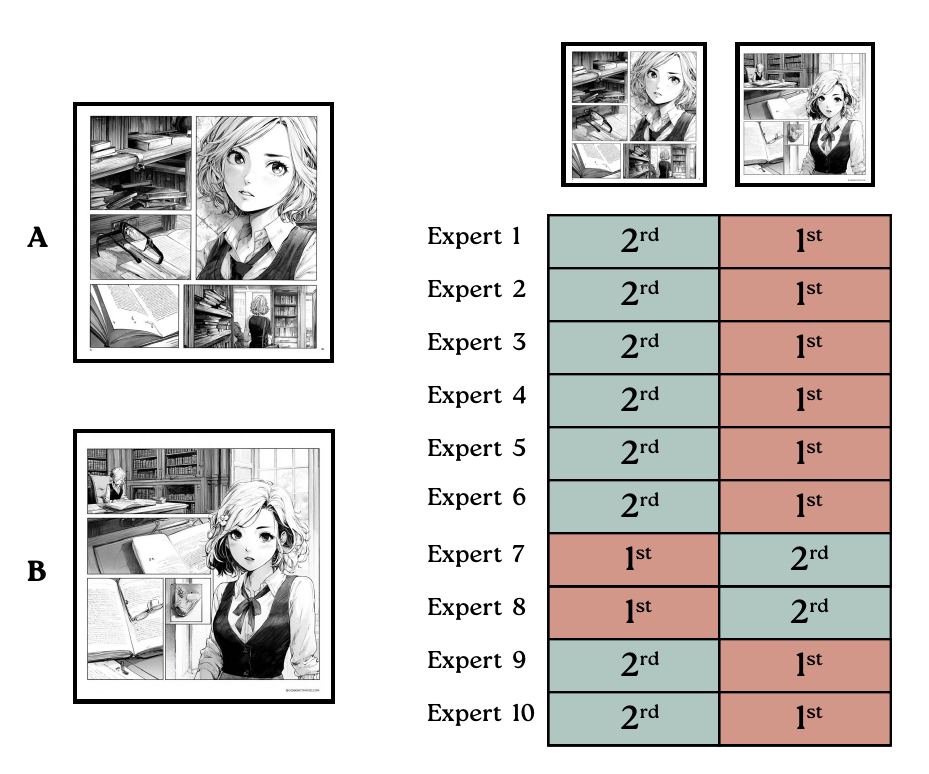}
\end{minipage}
\medskip
\par\medskip
\noindent\begin{minipage}{\textwidth}
  \raggedright\textbf{Comic}\par
  \medskip
  \centering
  \includegraphics[width=0.87\textwidth]{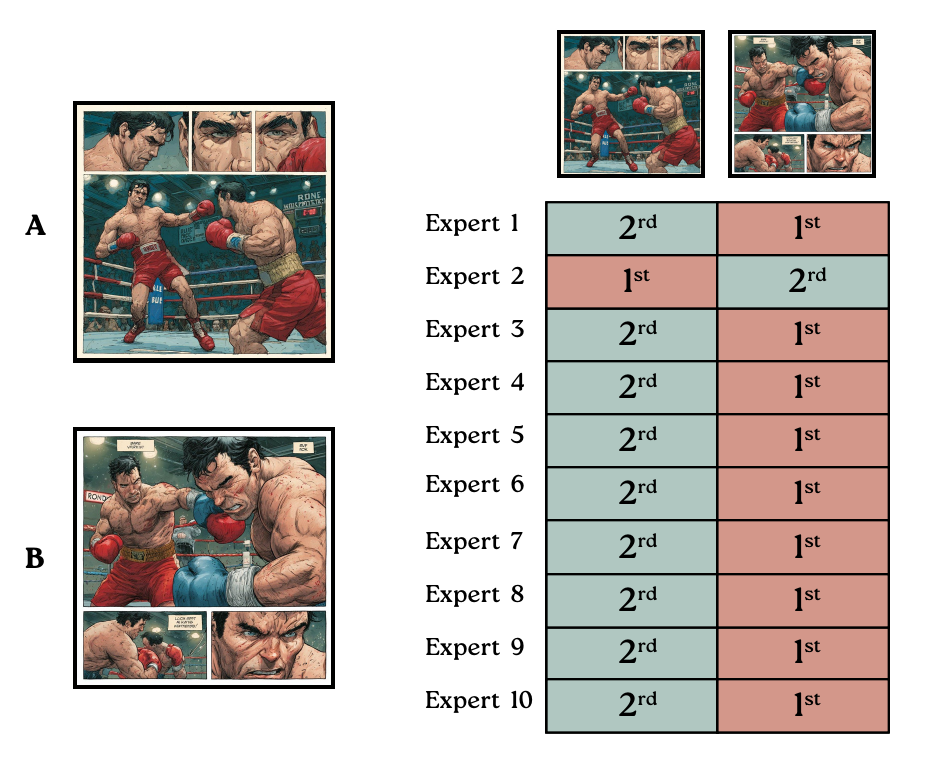}
\end{minipage}
\medskip
\par\medskip
\noindent\begin{minipage}{\textwidth}
  \raggedright\textbf{Concept Art}\par
  \medskip
  \centering
  \includegraphics[width=0.9\textwidth]{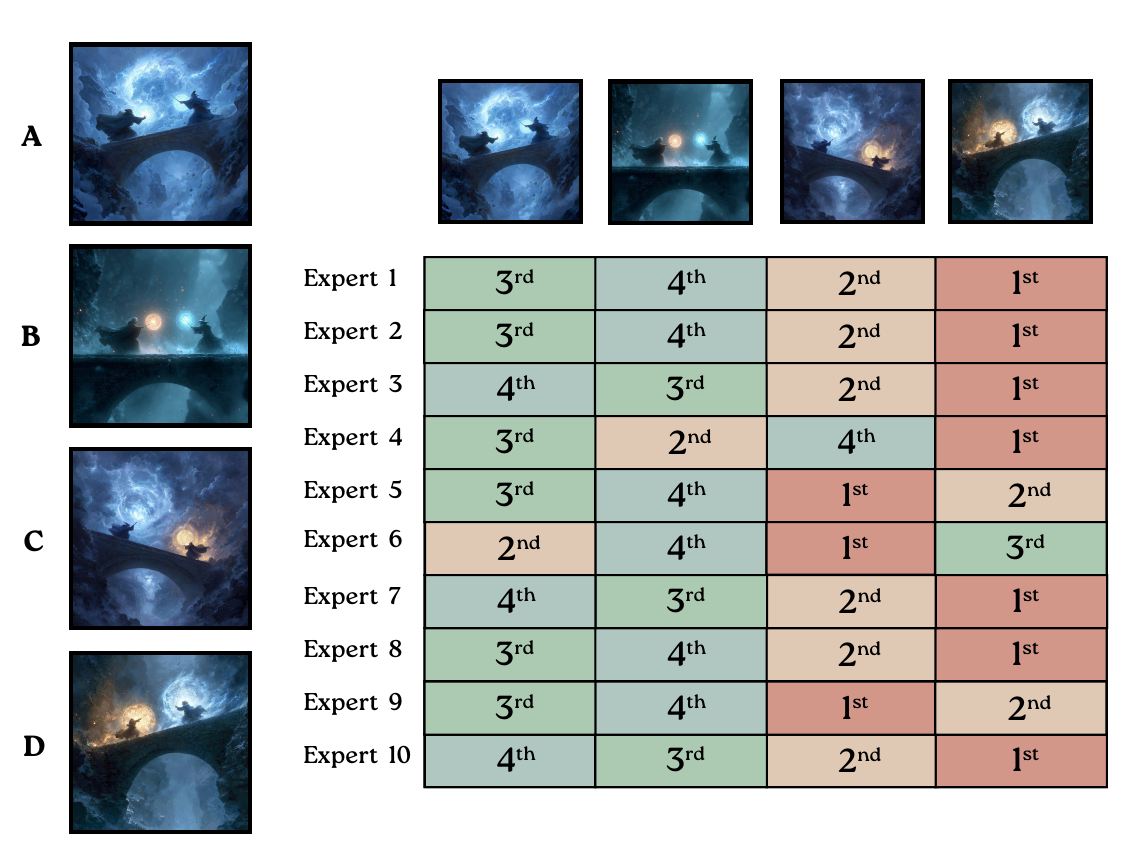}
\end{minipage}
\medskip
\par\medskip
\noindent\begin{minipage}{\textwidth}
  \raggedright\textbf{Digital and AI Art}\par
  \medskip
  \centering
  \includegraphics[width=1\textwidth]{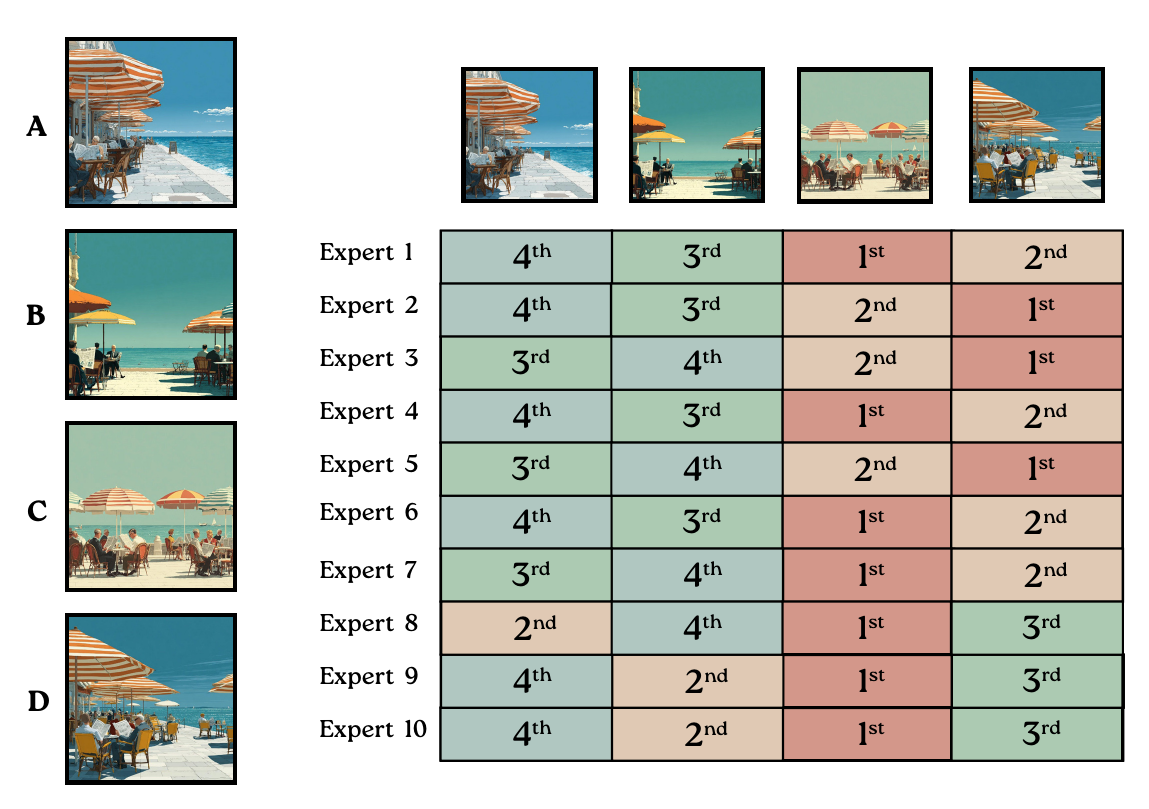}
\end{minipage}
\medskip
\par\medskip
\noindent\begin{minipage}{\textwidth}
  \raggedright\textbf{Pixel Art}\par
  \medskip
  \centering
  \includegraphics[width=1\textwidth]{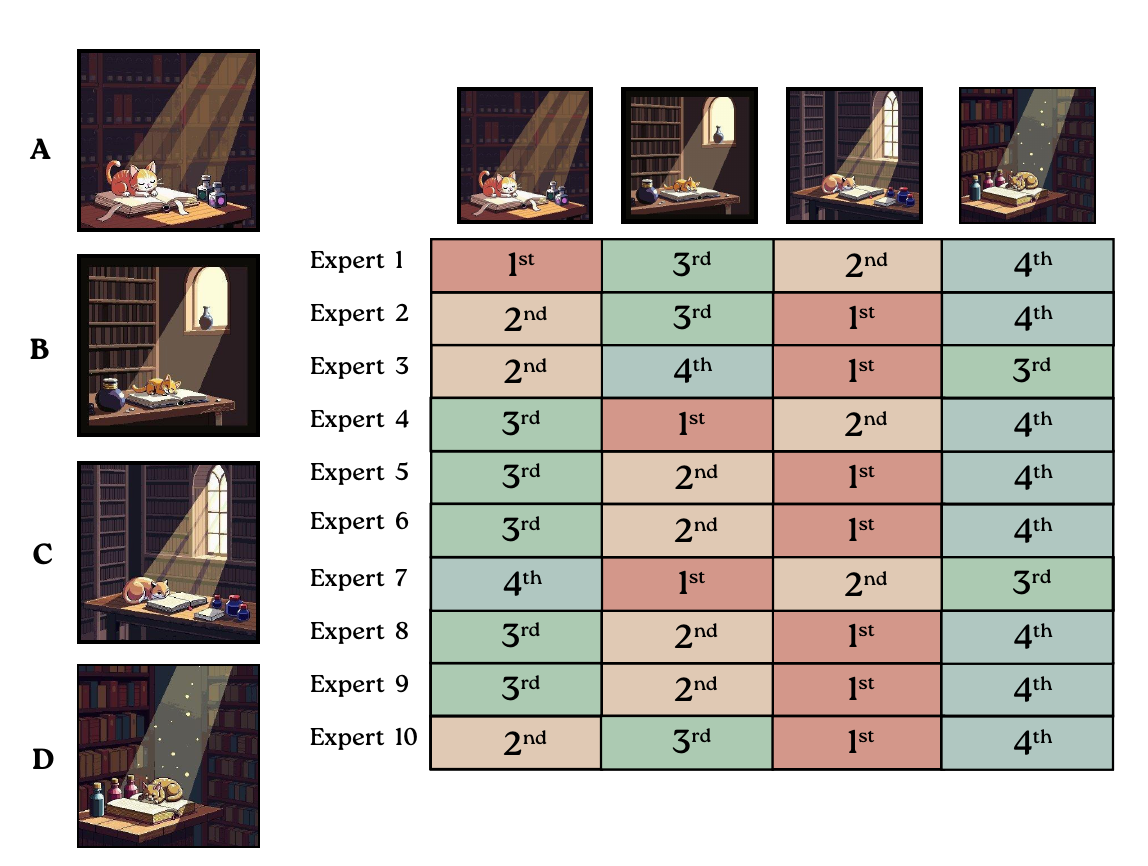}
\end{minipage}
\medskip
\par\medskip
\noindent\begin{minipage}{\textwidth}
  \raggedright\textbf{Stylized 3D}\par
  \medskip
  \centering
  \includegraphics[width=1\textwidth]{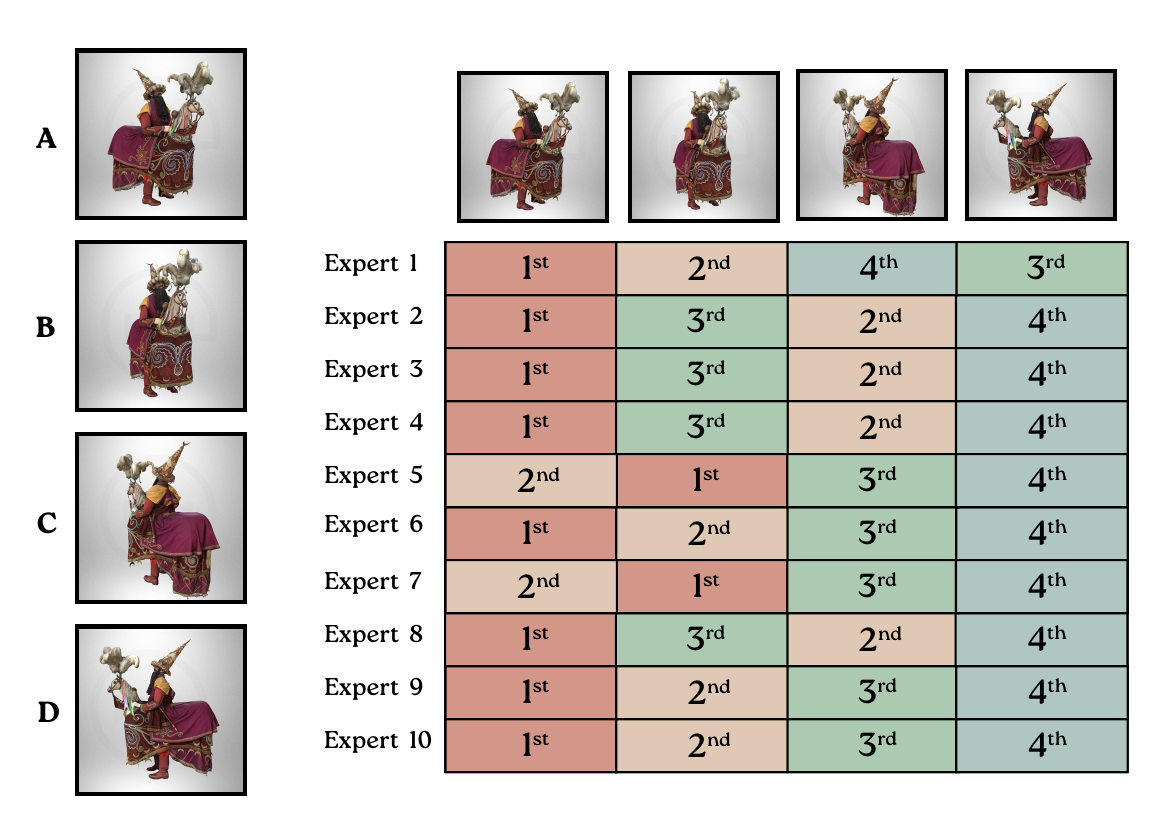}
\end{minipage}
\medskip

\end{document}